%% file: main.tex
\documentclass[10pt,twocolumn,letterpaper]{article}

\usepackage[pagenumbers]{cvpr} 

\definecolor{cvprblue}{rgb}{0.21,0.49,0.74}
\usepackage[breaklinks,colorlinks,allcolors=cvprblue]{hyperref}

\usepackage[most]{tcolorbox}
\usepackage{booktabs}
\usepackage{wrapfig}
\usepackage{amsmath}
\usepackage{algorithm}
\usepackage{algpseudocode} 
\usepackage{bm}
\usepackage{graphicx}
\usepackage{caption}
\usepackage[bottom]{footmisc}
\usepackage[capitalize]{cleveref}
\crefname{figure}{Fig.}{Figs.}
\crefname{table}{Tab.}{Tabs.}

\tcbset{
  mypromptbox/.style={
    colback=white,
    colframe=black!70,
    colbacktitle=black!80,
    coltitle=white,
    fonttitle=\bfseries,
    sharp corners,
    boxrule=0.5pt,
    top=4pt, bottom=4pt, left=2.5pt, right=2.5pt,
    before skip=10pt, after skip=10pt,
    breakable
  }
}

\newcommand{\todo}[1]{}
\newcommand{\note}[1]{}
\newcommand{\dothis}[1]{}
\newcommand{\andrew}[1]{}
\newcommand{\kate}[1]{}
\newcommand{\huzheng}[1]{}

\algrenewcommand\algorithmicrequire{\textbf{Input:}}
\algrenewcommand\algorithmicensure{\textbf{Output:}}

\def\paperID{*****} 
\def\confName{CVPR}
\def\confYear{2026}

\title{Vibe Spaces for Creatively Connecting and Expressing Visual Concepts}

\author{
Huzheng Yang\textsuperscript{1} \quad
Katherine Xu\textsuperscript{1} \quad
Andrew Lu\textsuperscript{1} \quad
Michael D. Grossberg\textsuperscript{2} \quad
Yutong Bai\textsuperscript{3} \quad
Jianbo Shi\textsuperscript{1} \\
\textsuperscript{1}UPenn \quad
\textsuperscript{2}CUNY \quad
\textsuperscript{3}UC Berkeley \\
\url{https://huzeyann.github.io/VibeSpace-webpage/}
}

\begin{document}

\input{figures/teaser}

\input{sec/0_abstract}

\input{sec/1_intro}
\input{sec/2_related_work}

\input{sec/3_path_finding}

\input{sec/4_vibe_space}

\input{sec/6_experiments}
\input{sec/7_conclusion}
\input{sec/X_appendix}

\clearpage
\newpage

{
    \small
    \bibliographystyle{ieeenat_fullname}
    \bibliography{main}
}

\end{document}

%% file: figures/teaser.tex
\twocolumn[{
\maketitle
\begin{center}
    \vspace{-14pt}
    \centering
    \includegraphics[trim=0in 0in 0in 0in, clip,width=\textwidth]{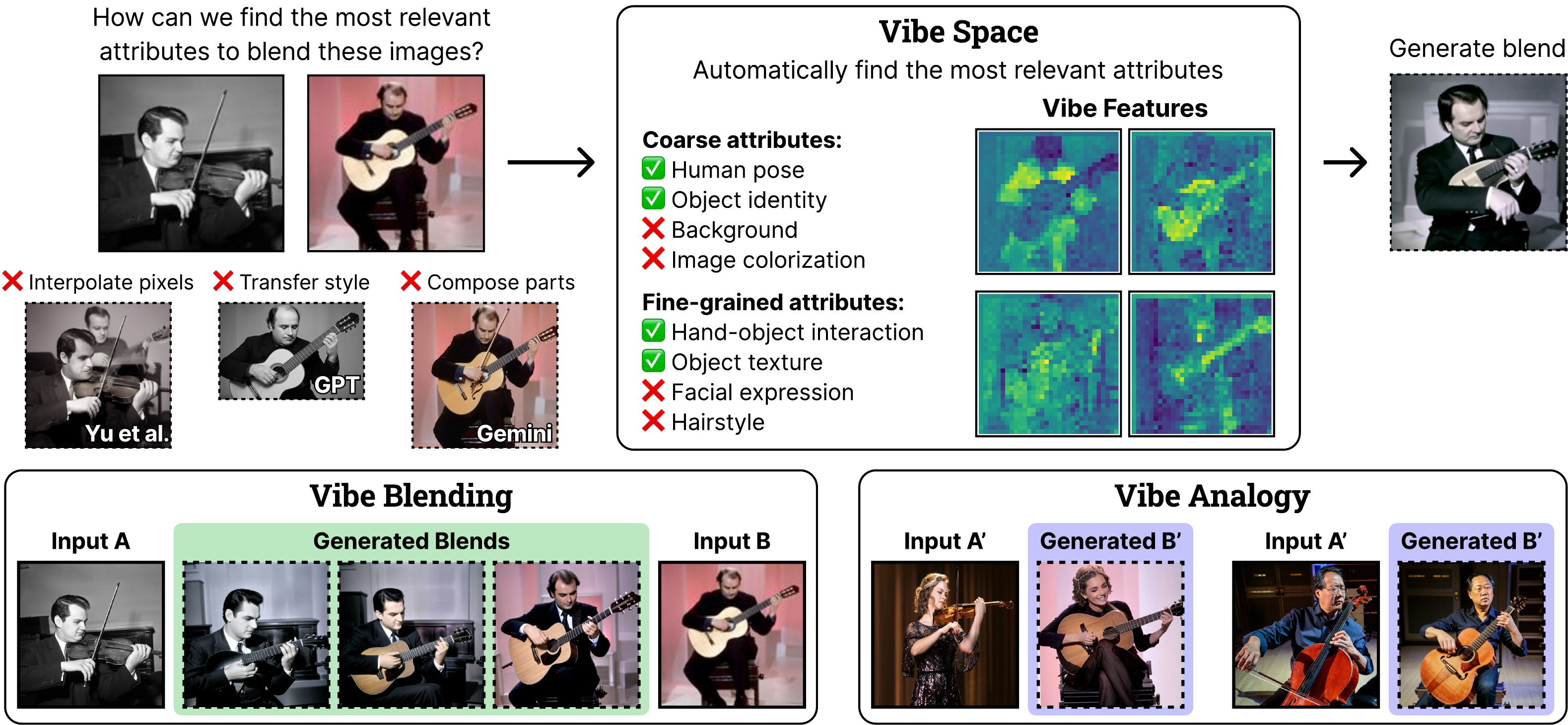}
    \vspace{-18pt}
    \captionof{figure}[Teaser]{What are the most relevant attributes for blending the violin player and guitar player? It is the instrument and how it is played, not the color or background. We call these attributes the ``vibe''\footnotemark[1]. Recent diffusion-based morphing methods such as Yu et al. \cite{yu2025probability} and LLMs like GPT \cite{gpt-image-1} and Gemini \cite{comanici2025gemini} struggle to blend the vibe, instead interpolating pixels, transferring style, or composing parts. We propose \textbf{Vibe Space} for identifying the vibe between input images and generating coherent, continuous blends that merge the vibe (\emph{Vibe Blending}). With the discovered vibe, we can extrapolate to nontrivial but related concepts, such as Hilary Hahn playing guitar (\emph{Vibe Analogy}).}
    \label{fig:teaser}
\end{center}
}]

%% file: sec/0_abstract.tex
\makeatletter
\renewenvironment{abstract}{
  \centerline{\large\bf Abstract}
  \vspace{8pt}   
  \noindent\it
}{\par}
\makeatother

\begin{abstract}
Creating new visual concepts often requires connecting distinct ideas through their most relevant shared attributes—their vibe. We introduce Vibe Blending, a novel task for generating coherent and meaningful hybrids that reveals these shared attributes between images. Achieving such blends is challenging for current methods, which struggle to identify and traverse nonlinear paths linking distant concepts in latent space. We propose Vibe Space, a hierarchical graph manifold that learns low-dimensional geodesics in feature spaces like CLIP, enabling smooth and semantically consistent transitions between concepts. To evaluate creative quality, we design a cognitively inspired framework combining human judgments, LLM reasoning, and a geometric path-based difficulty score. We find that Vibe Space produces blends that humans consistently rate as more creative and coherent than current methods.
\end{abstract}

%% file: sec/1_intro.tex
\vspace{-30pt}

\section{Introduction}
\label{sec:intro}

In \Cref{fig:teaser}, imagine blending a musician playing a violin with one playing a guitar. What are the most relevant attributes for blending? Large language models (LLMs) like Gemini or GPT might focus on object parts or style transfer, whereas a musician would attend to the instrument and how it is played. The intuitive process of identifying and fusing meaningful attributes---the ``vibe''\footnotemark[1]---reveals creative connections between distinct concepts. We call this process Vibe Blending: creating coherent hybrids that merge the relevant shared attributes between images. In Fig. \ref{fig:teaser}, blending the vibe of a violin and guitar yields a lute: an instrument played like a guitar but similar in size to a violin.

\footnotetext[1]{The term ``vibe,'' short for ``vibration,'' originated in 1960s jazz slang to describe the mood or feeling conveyed by music, a person, or space.}

Psychologist Sarnoff A. Mednick proposed that creativity arises from linking distant concepts in new and meaningful ways \cite{mednick1962associative}, such as a poet fusing a ``lion's ferocious chrysanthemum head'' or a scientist uniting disparate theories. Memory, attention, and control are central to creativity \cite{beaty2023associative}. LLMs have expansive memory through training data but limited attention and control, often struggling to recognize which attributes matter (e.g., the instrument and body pose in the violin--guitar example) and how visual features should be merged to express those attributes coherently.

The path of creatively blending the ``vibe'' is rarely linear. When two concepts lie far apart in latent space, naive linear interpolation produces incoherent images that lack semantic integrity. We propose \textbf{Vibe Space} to discover and traverse non-linear paths connecting distant visual concepts in pretrained feature spaces like CLIP. Our key insight is to reframe the problem: instead of navigating the full ambient feature space, we learn a compact, low-dimensional manifold---a ``small world''---from a few context images that captures the relevant geometry. Steps in this small world correspond to meaningful transitions in the original feature space, enabling coherent creative blending.

Evaluating creativity requires new metrics and datasets. We take clues from cognitive psychology and study whether we can quantify human judgments of creative potential and blend difficulty using foundation models like CLIP and LLMs. We curate evaluation datasets by selecting conceptually interesting image pairs from Totally Looks Like \cite{rosenfeld2018totally}, as well as from architectural design images. Our results show that LLMs offer a useful starting point for quantifying creativity. We also introduce a path-based metric for blend difficulty, enabling the curation of more challenging and engaging image pairs for future blending tasks.

Our key contributions are the following:
\begin{enumerate}[noitemsep, nolistsep]
    \item \textbf{Vibe Blending}, a task for generating creative, continuous morphs between visual concepts by \emph{discovering and merging the most relevant shared attributes}. 
    
    \item We propose \textbf{Vibe Space}, a method to learn a hierarchical graph manifold that guides geodesics in ambient feature spaces (e.g., CLIP or DINO) for Vibe Blending with minimal on-the-fly training (under 1M parameters).

    \item Drawing inspiration from cognitive psychology, we curate datasets and conduct human-subject experiments to measure creativity in Vibe Blending through creative potential and blend difficulty. 

    \item Our method generates more creative blends than strong baselines, such as Gemini and GPT. 
\end{enumerate}

%% file: sec/2_related_work.tex
\section{Related Work}
\label{sec:related_work}
\vspace{-5pt}

Creativity is often regarded abstract, but can be quantified through semantic memory networks, where concepts are nodes connected by associations \cite{beaty2023associative}. Cognitive psychology suggests that more creative individuals traverse the network further, connecting distant or weakly linked concepts \cite{he2020relation, kenett2016examining, mednick1962associative}. Rather than jumping directly between remote ideas (e.g., \emph{apple} → \emph{house}), they move through intermediate associations (e.g., \emph{apple} → \emph{tree} → \emph{wood} → \emph{house}), following nonlinear paths across clusters of related concepts. This perspective motivates Vibe Blending, which fuses the relevant attributes of different concepts.

Prior methods have mainly focused on pixel-level image blending in the latent space of generative models, such as GANs \cite{pan2021exploiting} and diffusion models \cite{he2024aid, wang2023interpolating, yang2023impus, zhang2024diffmorpher, yu2025probability, guo2024smooth, cao2025freemorph}. Additional work \cite{Kulikov_2025_ICCV, kwon2022diffusion, park2023understanding, wu2023latent} explored finding semantic directions in the latent space of diffusion models, such as for the noise space \cite{samuel2023norm, brack2023sega}, weight space \cite{dravid2024interpreting, gandikota2024concept}, and text embedding space \cite{baumann2024continuous, matsuhiraBlend2024}.
Closest to our setting is AID \cite{he2024aid}, which performs \emph{attention interpolation} inside diffusion models by fusing the attention. AID treats all attributes uniformly and relies on attention correspondences rather than identifying \emph{most relevant} attributes for the blend.

Recent work on creative image blending \cite{sun2025creative, peng2025probing} has relied on language, which often fails to capture precise visual attributes. In contrast, our method operates directly in image feature space and outperforms custom-trained diffusion models for morphing and recent multimodal LLMs.

\input{figures/figure_compare}

%% file: figures/figure_compare.tex
\begin{figure}[!th]
    \vspace{-5pt}
    \centering
    \includegraphics[trim=0in 0in 0in 0in, clip,width=\columnwidth]{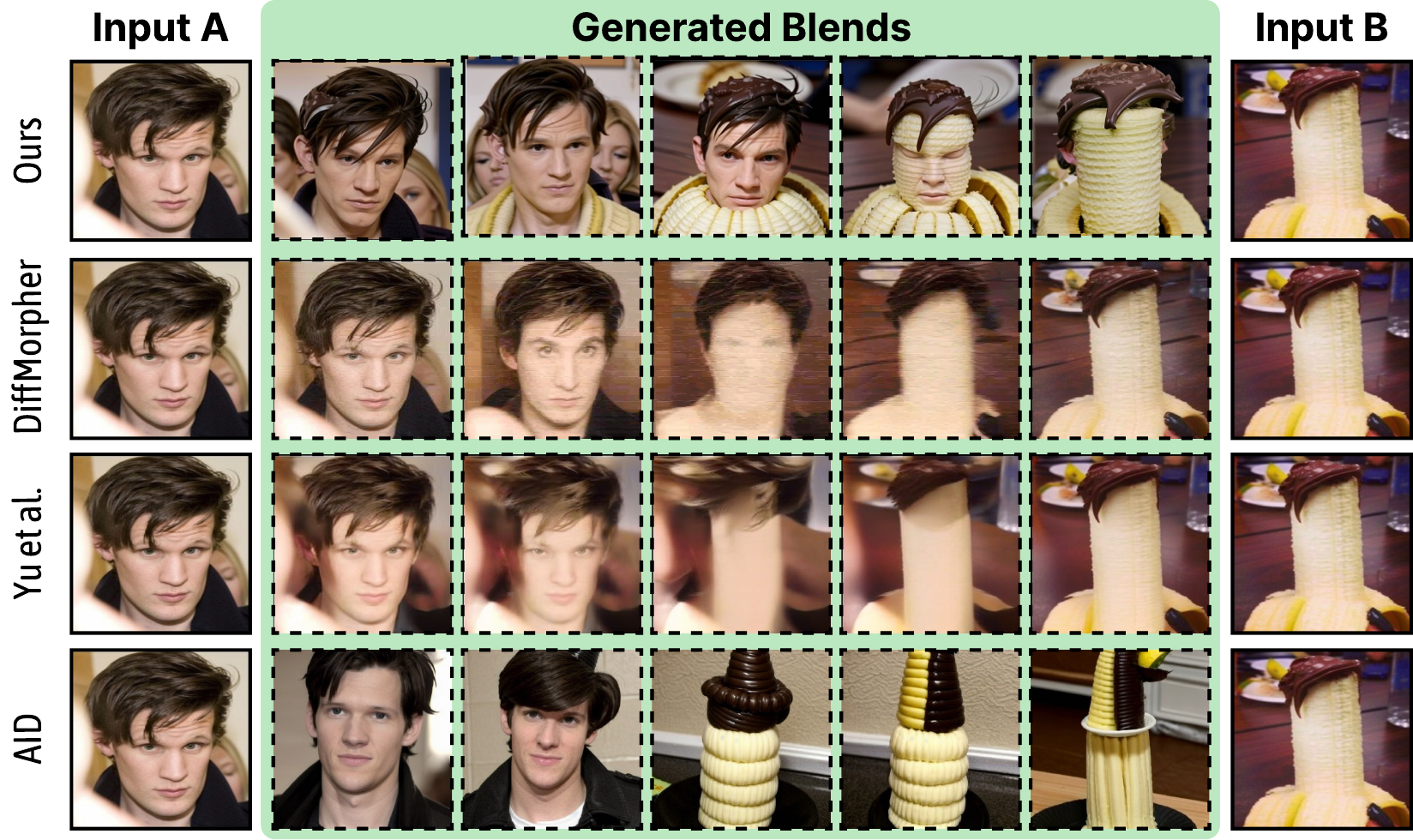}
    \vspace{-18pt}
    \caption{Our method generates coherent blends that focus on the most relevant attributes shared by the input images---the hairstyle. Diffusion-based morphing methods like DiffMorpher \cite{zhang2024diffmorpher} and Yu et al. \cite{yu2025probability} struggle to produce realistic blends of distant concepts, and AID \cite{he2024aid} fails to capture hairstyle as the relevant attribute.}
    \vspace{-15pt}
    \label{fig:compare_slerp_diffmorpher}
\end{figure}

%% file: sec/3_path_finding.tex
\section{Creative Path Finding}
\label{sec:method_path_finding}

\input{figures/figure2}

\subsection{Finding Paths to Connect Concepts Creatively}

Creativity often emerges from finding connections between different concepts through key attributes, or a ``vibe''. There are many possible attributes, and each attribute can link concepts through multiple \emph{paths} within the conceptual space. We aim to discover a simple path that captures the ``vibe'' without unnecessary complexity. This idea parallels how human creativity involves navigating rich semantic networks while using attention and cognitive control to focus on meaningful connections \cite{beaty2023associative, benedek2019toward}. We model the creative process using co-saliency, which reveals the relevant attributes that are jointly prominent across concepts.

However, finding meaningful paths within the conceptual space is difficult. High-dimensional feature spaces are highly nonlinear and contain numerous ``holes''---regions corresponding to implausible, distorted, or low-quality images. Naive linear interpolation in latent space often produces inconsistent intermediate images or ghosting artifacts. We hypothesize that these problematic ``holes'' arise from a fundamental mismatch between the intrinsic dimension of the data manifold and the higher-dimensional latent space. Recent diffusion-based morphing methods like DiffMorpher \cite{zhang2024diffmorpher}, Yu et al. \cite{yu2025probability}, and AID \cite{he2024aid} also struggle to connect distant or spatially misaligned concepts (\Cref{fig:compare_slerp_diffmorpher}). 

\vspace{-10pt}
\paragraph{Graph Diffusion and Unfolded Manifold Coordinates.} 
We derive a latent graph diffusion space that captures the intrinsic geometry of the data manifold, where connecting concepts yields geodesics\footnotemark[2] that stay close to the manifold.

\footnotetext[2]{A geodesic is a shortest path (curve) between points on a manifold.}

Finding a path along a manifold whose structure is approximated by a feature similarity graph \(\mathbf{W}\) relates to the generalized eigenvectors of the graph’s Laplacian matrix \(\mathbf{L}\) (Fig. \ref{fig:claim3}a). The process of unfolding the manifold aims to reveal these paths. The graph Laplacian is defined as \(\mathbf{L} = \mathbf{D} - \mathbf{W}\), where \(\mathbf{D}\) is the diagonal degree matrix. 
The eigenvectors of \(\mathbf{L}\) (or its normalized variants) -- in particular the \(m\) eigenvectors associated with the smallest nonzero eigenvalues, \(\mathbf{\psi}_2,\dots,\mathbf{\psi}_{m+1}\) -- serve as the new coordinates for the data points (Fig.~\ref{fig:claim3}b, \ref{fig:claim3}e).
These graph eigenvectors capture the intrinsic geometric structure of the manifold. Note that the \emph{graph diffusion map} \cite{coifman2005geometric} -- denoted by $\mathbf{\Psi}$ -- is a classical manifold-learning construction based on graph eigenvectors (\textbf{not related to diffusion-based image generation}). Projecting the original manifold into the graph diffusion map makes the originally curved manifold paths approximately linear and easy to compute.

\vspace{-10pt}
\paragraph{Graph Diffusion for Manifold Distance and Geodesics.} 
Let $D_t(\mathbf{x}_A, \mathbf{x}_B)$ between two points $\mathbf{x}_A$ and $\mathbf{x}_B$ represent the probability that a random walk will connect them in $t$ time steps. 
A key property of diffusion maps \cite{coifman2005geometric} is that this diffusion distance
equals the Euclidean distance between points in the embedding space $\mathbf{\Psi}_t$ (Fig. \ref{fig:claim3}b right):
\small
\vspace{-2pt}
\begin{equation}
D_t(\mathbf{x}_i, \mathbf{x}_j) = \| \mathbf{\Psi}_t(\mathbf{x}_i) - \mathbf{\Psi}_t(\mathbf{x}_j) \|_2 ,
\label{eq:diffusion_distance}
\end{equation}
\vspace{-2pt}
\normalsize
where the diffusion map $\mathbf{\Psi}_t$ projects each data point $\mathbf{x}_i$ into the first $m$ eigenvectors of the graph Laplacian
$\mathbf{\Psi}_t(\mathbf{x}_i) = (\lambda_1^t \mathbf{\psi}_1(i), \lambda_2^t \mathbf{\psi}_2(i), \ldots, \lambda_{m}^t \mathbf{\psi}_{m}(i))$.
This embedding enables constructing geodesics by iteratively querying ``midpoints'' that have a smoothly varying diffusion distance with respect to the endpoints. 

\vspace{-5pt}
\paragraph{Geodesics from Inverse Diffusion Mapping.}
Having embedded the data into graph diffusion space, we can now recover geodesics in the original manifold by inverse mapping. Given two points $\mathbf{x}_A$ and $\mathbf{x}_B$ with diffusion coordinates $\mathbf{\Psi}_t(\mathbf{x}_A)$ and $\mathbf{\Psi}_t(\mathbf{x}_B)$, we seek a path $\gamma(\alpha), \alpha \in [0, 1]$ with $\gamma(0) = \mathbf{x}_A$ and $\gamma(1) = \mathbf{x}_B$. We first compute the interpolated diffusion coordinate
$\mathbf{\Psi}_t(\mathbf{x}_\alpha) = (1 - \alpha)\,\mathbf{\Psi}_t(\mathbf{x}_A) + \alpha\,\mathbf{\Psi}_t(\mathbf{x}_B)$,
then recover the corresponding path $\gamma(\alpha)$ on the original manifold via inverse mapping (Fig. \ref{fig:claim3}c):
\small
\vspace{0pt}
\begin{equation}
\gamma(\alpha) = \arg\min_{\mathbf{x^*}} \big\| \mathbf{\Psi}_t(\mathbf{x^*}) - \mathbf{\Psi}_t(\mathbf{x}_\alpha) \big\|_2^2.
\label{eq:inverse_mapping}
\end{equation}
\vspace{-10pt}
\normalsize

Inverse map optimization is tractable because (1) the Jacobian \(\nabla_{\mathbf{x}^*}\mathbf{\Psi}_t(\mathbf{x}^*)\) admits a closed-form expression via eigenvalue perturbation theory \cite{rellich1969perturbation}, and (2) the Nystr\"om approximation \cite{fowlkes_nystrom} enables efficient updates of the \(\mathbf{\Psi}_t(\mathbf{x}^*)\) under small perturbations. Together, these yield a practical solver for Eq.~\eqref{eq:inverse_mapping}: we iteratively adjust \(\mathbf{x}^*\) to align its diffusion coordinates with the target \(\mathbf{\Psi}_t(\mathbf{x}_\alpha)\), producing a path \(\gamma(\alpha)\) that stays close to the data manifold.

\input{figures/figure3}

\subsection{Flag Space for Multiscale Paths}

\paragraph{Why Flag Space.}
Graph Laplacian eigenvectors produce manifold coordinates to guide a geodesic path, and they capture geometry at different scales: leading eigenvectors describe global structure, while higher-order eigenvectors encode local variations.

How many eigenvectors should we keep?  Truncating to a fixed $\mathbf{\Psi}^{1:m}$ selects one scale and discards the rest, leading to a path that focuses on too many or too few attributes. 

We propose to use a \emph{flag space}\footnotemark[3] \cite{Brion2004FlagVarieties}, a hierarchy of nested embeddings
$\mathbf{\Psi}^{1:m_1} \subset \mathbf{\Psi}^{1:m_2} \dots \subset \mathbf{\Psi}^{1:m_M},$ where $m_1 < m_2 < \dots < m_M$,
to encapsulate both the coarse and fine manifold structures and to alleviate the impact of picking the wrong number of eigenvectors (Fig. \ref{fig:claim3}b, \ref{fig:claim3}e).

\vspace{-10pt}
\paragraph{Multi-scale Path.}

We extend inverse diffusion mapping in \Cref{eq:inverse_mapping} to operate over \emph{flag space}. Using the same linear interpolation $\mathbf{\Psi}_t(\mathbf{x}_\alpha)$ in diffusion space, the manifold path $\gamma(\alpha), \alpha \in [0, 1]$ is recovered by minimizing the average reconstruction error across a set of scales $\mathcal{M}$:
\vspace{-2pt}
\begin{equation}
\small{
\hspace{-7pt}\gamma(\alpha) = \arg\min_{\mathbf{x^*}} \underbrace{\frac{1}{|\mathcal{M}|} \sum_{m_k \in \mathcal{M}}}_{\text{flag space}} \underbrace{\big\| \mathbf{\Psi}_t^{1:m_k}(\mathbf{x^*}) - \mathbf{\Psi}_t^{1:m_k}(\mathbf{x}_\alpha) \big\|_2^2}_{\text{inverse mapping}}\footnotesize{.}\hspace{-2pt}
\vspace{-2pt}
}
\label{eq:flag_space_inverse_mapping}
\end{equation}
This enforces consistency across global and local geometry when finding the path $\gamma(\alpha)$ between $\mathbf{x}_A$ and $\mathbf{x}_B$.

%% file: figures/figure2.tex
\begin{figure*}[!ht]
    \vspace{-20 pt}
    \centering
    \includegraphics[trim=0in 0in 0in 0in, clip,width=\textwidth]{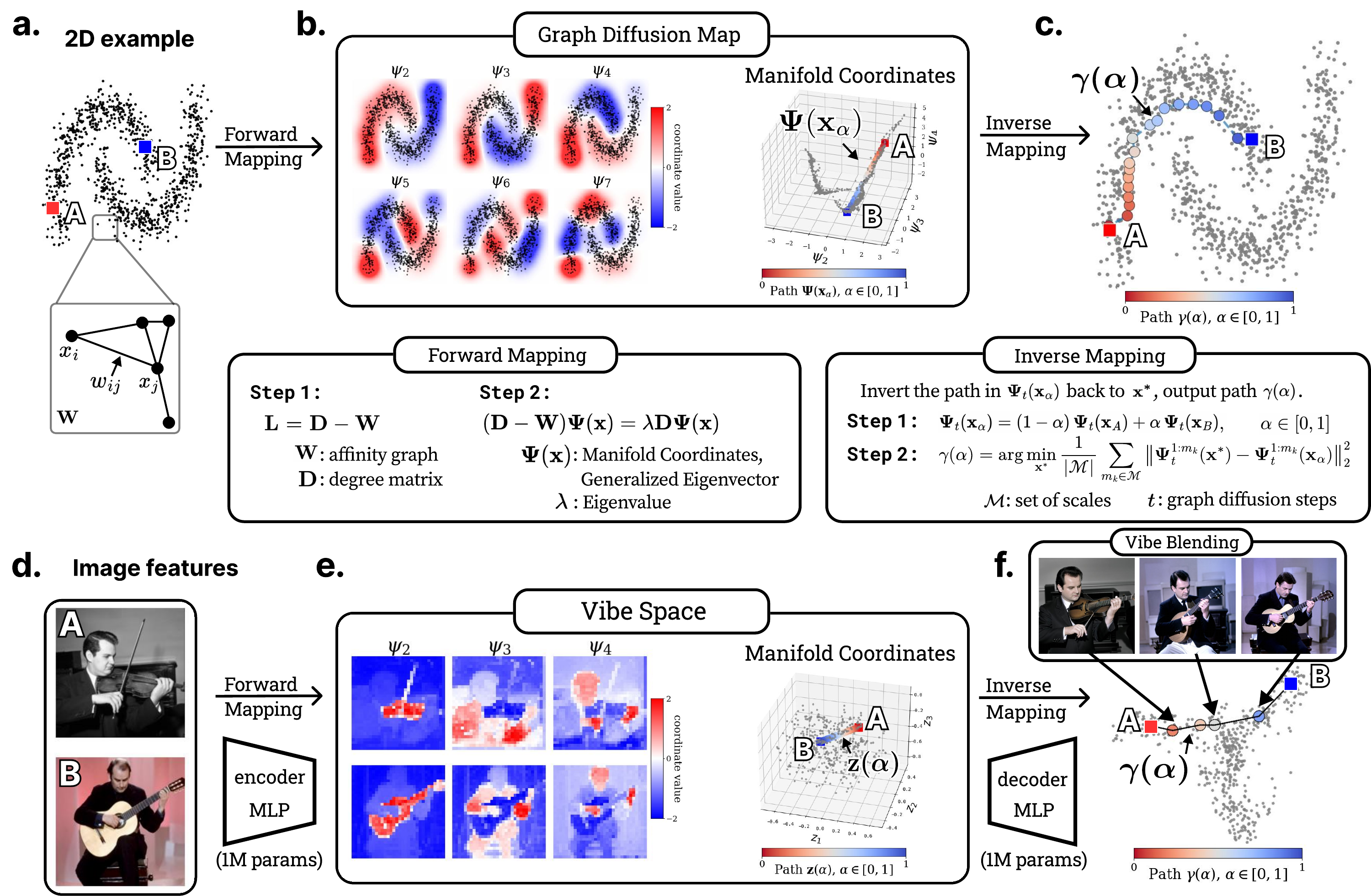}
    \vspace{-18 pt}
    \caption{\textbf{Top:} Using a 2D point cloud example, the forward mapping involves computing the \textbf{(a.)} affinity graph $\mathbf{W}$ of the points $\mathbf{x}$ on the manifold and \textbf{(b.)} generalized eigenvectors $\mathbf{\Psi}(\mathbf{x})$ of the graph Laplacian $\mathbf{L}$ as manifold coordinates of the point cloud $\mathbf{x}$. \textbf{(c.)} The inverse mapping performs linear interpolation in the manifold space and uses graph diffusion inversion to obtain the corresponding path in the original point cloud space. \textbf{Bottom:} \textbf{(d.)} On real images, we extract patch tokens from DINO features as graph nodes and compute token-wise affinity $\mathbf{W}$. \textbf{(e.)} The top $m$ graph eigenvectors produce co-salient segments across two images for blending, and manifold coordinates for expressing ``vibe'' features for blending. \textbf{(f.)} Similar to the point cloud example, we apply graph diffusion inversion to obtain a path in CLIP space. We ``render'' pixel images from CLIP features using a frozen IP-Adapter \cite{ye2023ip}.  We train two lightweight MLP networks in under 30 seconds: an encoder to simulate and compress the forward mapping, and a decoder to mimic the inverse mapping.}
    \vspace{-13 pt}
    \label{fig:claim3}
\end{figure*}

%% file: figures/figure3.tex
\begin{figure*}[!th]
    \vspace{-18 pt}
    \centering
    \includegraphics[trim=0in 0in 0in 0in, clip,width=\textwidth]{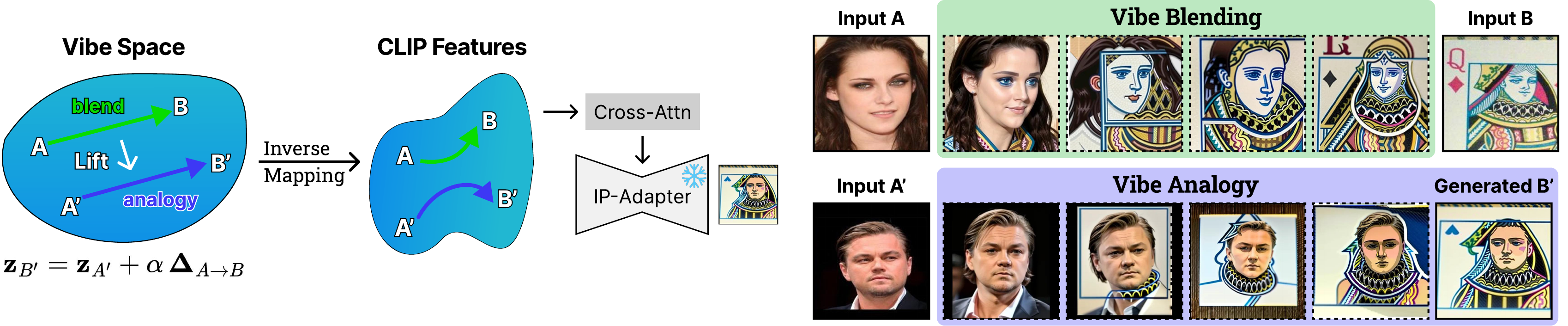}
    \vspace{-18 pt}
    \caption{From Vibe Blending to Vibe Analogy. Vibe Space enables creative connections between input images $A$ and $B$. A path approximately linear in Vibe Space results in a continuous manifold-following path in the ambient feature space, such as CLIP. We can lift the ``vibe'' $\mathbf{\Delta}_{A\to B}$ to a non-trivial but related image $A'$ to extrapolate an analogous path in the ambient space, resulting in image $B'$ that reflects the same vibe. For example, we can morph Leonardo DiCaprio's face into a playing card.}
    \vspace{-10 pt}
    \label{fig:method}
\end{figure*}

%% file: sec/4_vibe_space.tex
\section{Vibe Space and Image Generation} \label{sec:method_vibe_space_apply}

To speed up the inverse mapping, we learn two small MLPs on the fly: an encoder $f$ mimics the flag-space diffusion map, and a decoder $g$ learns the inverse mapping. Feature extraction and graph eigenvector computation\footnotemark[4] run in milliseconds. Encoder–decoder training runs in under 30s.

\footnotetext[3]{Flag space: \href{https://en.wikipedia.org/wiki/Flag_(linear_algebra)}{https://en.wikipedia.org/wiki/Flag\_(linear\_algebra)}}

\footnotetext[4]{Graph eigenvectors are computed efficiently in milliseconds \cite{ncut-pytorch}.}

\subsection{Learning the Vibe Space}

Given dense DINO \cite{caron2021emerging} features $\mathbf{x}\in\mathbb{R}^{(HW)\times D}$, the encoder
$f:\text{DINO} \to \text{Vibe}$ maps each token to a latent representation $\mathbf{z}=f(\mathbf{x})\in\mathbb{R}^{(HW)\times d}$, where $d\ll D$ (typically $d \approx 6$) (Fig. \ref{fig:claim3}d).
The decoder $g:\text{Vibe} \to \text{CLIP}$ maps this latent representation back to CLIP \cite{radford2021learning} space.
We refer to the latent embedding produced by $f$ as the \textbf{Vibe Space}. Its geometry is trained to align with the multiscale structure of flag-space diffusion maps, so that Euclidean distances and linear paths in Vibe Space correspond to geodesic distances and semantic paths on the underlying manifold (Fig.~\ref{fig:claim3}e, \ref{fig:claim3}f).

We enforce geometric alignment by matching the Gram matrix $\mathbf{z}\mathbf{z}^\top$ to the flag-space kernel matrix $\mathbf{S}(\mathbf{\Psi}(\mathbf{x}))$:
\small
\begin{equation}
\begin{aligned}
\mathcal{L}_{\text{flag\_enc}}(f)
&= \big\|\mathbf{z}\mathbf{z}^\top - \mathbf{S}(\mathbf{\Psi}(\mathbf{x}))\big\|_2^2, \\
\mathcal{L}_{\text{flag\_dec}}(f, g)
&= \big\|\mathbf{z}\mathbf{z}^\top - \mathbf{S}(\mathbf{\Psi}(g(\mathbf{z})))\big\|_2^2,
\end{aligned}
\qquad
\raisebox{0.5ex}{%
\(\begin{aligned}
\mathbf{z} &= f(\mathbf{x}),
\end{aligned}\)
}
\label{eq:flag_loss}
\end{equation}
\normalsize
where $\mathbf{S}(\mathbf{\Psi}(\mathbf{x}))_{ij}=\frac{1}{|\mathcal{M}|}\sum_{m_k\in\mathcal{M}}
\mathbf{\Psi}^{1:m_k}(\mathbf{x}_i)\mathbf{\Psi}^{1:m_k}(\mathbf{x}_j)^\top$ aggregates inner products across nested eigen-embeddings.
To improve generalization beyond observed data, we add extrapolation regularization via random samples $\mathbf{z}_{\text{sample}}$:
\small
\begin{equation}
\mathcal{L}_{\text{sample}}(g)
= \big\|\mathbf{z}_{\text{sample}}\mathbf{z}_{\text{sample}}^\top
- \mathbf{S}\!\left(\mathbf{\Psi}\!\big(g(\mathbf{z}_{\text{sample}})\big)\right)\big\|_2^2.
\label{eq:sample_loss}
\end{equation}
\normalsize
Finally, we bridge perception and generation by encoding DINO features into Vibe Space and decoding them to CLIP, retaining DINO’s semantic richness while remaining compatible with CLIP-conditioned diffusion generators \cite{ye2023ip}.
\small
\begin{equation}
\mathcal{L}_{\text{recon}}(f,g)=\|\mathbf{x}^{\text{clip}}-g(f(\mathbf{x}^{\text{dino}}))\|_2^2.
\label{eq:recon_loss}
\end{equation}
\normalsize
Since $\mathbf{z}\mathbf{z}^\top \approx \mathbf{S}(\mathbf{\Psi}(\mathbf{x}))$, linear interpolation in Vibe Space approximates the multiscale inverse-diffusion geodesic in closed form (Fig. \ref{fig:claim3}f).

\vspace{-8pt}
\paragraph{Vibe Blending.}
Given two input images $(I_A, I_B)$, our goal is to generate intermediate blended images $\{I_\alpha\}_{\alpha\in[0,1]}$. 
Vibe Blending operates in four steps: (1) train the Vibe Space, (2) determine which pair of attributes to blend by correspondence matching, (3) interpolate a path in Vibe Space, and (4) decode the path and generate images.
Algorithm~\ref{alg:vibe-path} summarizes the full procedure.

\input{figures/algorithm1}

\Cref{alg:vibe-path} lines 1-3 optionally accept additional related exemplars. Although two images suffice to train the Vibe space and identify the dominant attributes, adding related exemplars can enhance the dominant attributes.

\Cref{alg:vibe-path} lines 5-6 determine which attributes should be blended. Since concepts rarely align at the pixel level, we first cluster DINO tokens in each image into semantic segments using $k$-way NCut \cite{1238361}. We then compute a segment-level correspondence using the Hungarian algorithm. Each segment in $I_A$ is matched to a  segment in $I_B$, yielding a bijection $\pi : I_B \leftrightarrow I_A$. This mapping $\boldsymbol{\Delta}_{A\to B} = \pi(\mathbf{z}_B) - \mathbf{z}_A$ establishes how attributes in $I_A$ should morph toward $I_B$.

\Cref{alg:vibe-path} line 10 generates images with IP-Adapter \cite{ye2023ip}, a diffusion model conditioned on dense CLIP image features. No finetuning of the IP-Adapter is required.

\vspace{-8pt}
\paragraph{Vibe Analogy.}
In \Cref{fig:method}, after learning a transition $I_A\!\to\! I_B$, we reuse its displacement field $\boldsymbol{\Delta}_{A\to B}$. We align $I_{A'}$ to $I_A$ via the same region correspondences, apply $\boldsymbol{\Delta}_{A\to B}$ in Vibe Space to obtain $\mathbf{z}_{B'}$, then decode to $I_{B'}$. 

\subsection{Creative Control}
\input{figures/neg_vibe}

Just as how a set of positive exemplars can be used to drive Vibe Blending, we can also exclude ``negative vibes" with negative exemplars (\Cref{fig:neg_vibes}). Given a set of images $I_{pos}$ for blending and a set $I_{neg}$ that captures a “negative vibe” to remove, we suppress undesired attributes by projecting the positive attribute basis away from directions spanned by the negative exemplars. We compute flag-space eigenvectors for both positive ($\mathbf{\Psi}_{pos}$) and negative ($\mathbf{\Psi}_{neg}$) sets and orthogonalize the positive basis against the negative: $\mathbf{\Psi}_{\text{filtered}} = \mathbf{\Psi}_{pos} - \beta \cdot \mathbf{\Psi}_{neg} (\mathbf{\Psi}_{neg}^\top \mathbf{\Psi}_{pos})$. Training the Vibe Space to match the kernel of this filtered basis, $\mathbf{S}(\mathbf{\Psi}_{\text{filtered}})$, learns a representation that avoids the undesired attributes.

%% file: figures/algorithm1.tex
\begin{algorithm}[h]
\caption{Vibe Blending}
\label{alg:vibe-path}
\small{
\begin{algorithmic}[1]
\Require Images $(I_A, I_B)$, image features $(\mathbf{x}^{\text{dino}}, \mathbf{x}^{\text{clip}})$
\Ensure Generated intermediate images $\{I_\alpha\}_{\alpha\in[0,1]}$

\State $\mathbf{W}_{ij} = \text{exp}{(-\frac{\|\mathbf{x}_i^{\text{dino}} - \mathbf{x}_j^{\text{dino}}\|^2}{\sigma^2})}$, $\mathbf{D}_{ii} = \sum_j \mathbf{W}_{ij}$ \Comment{Graph}

\State $(\mathbf{D} - \mathbf{W})\mathbf{\Psi}(\mathbf{x}^{\text{dino}}) = \lambda \mathbf{D}\mathbf{\Psi}(\mathbf{x}^{\text{dino}})$ \Comment{Graph Diffusion Map}

\State $f, g \gets \text{Train}(\mathbf{x}^{\text{clip}}, \mathbf{x}^{\text{dino}}, \mathbf{\Psi}(\mathbf{x}^{\text{dino}}))$  \Comment{Train Vibe Space}

\State $\mathbf{z}_A = f(\mathbf{x}_A^{\text{dino}})$; $\mathbf{z}_B = f(\mathbf{x}_B^{\text{dino}})$ \Comment{Encode vibe}

\State $\pi \gets \text{Match}(\mathbf{x}_A^{\text{dino}}, \mathbf{x}_B^{\text{dino}})$ \Comment{Cluster correspondence}

\State $\boldsymbol{\Delta}_{A\to B} = \pi(\mathbf{z}_B) - \mathbf{z}_A$ \Comment{Path direction}

\For{$\alpha \in [0,1]$}
    \State $\mathbf{z}_\alpha = 
           \mathbf{z}_A + \alpha\,\boldsymbol{\Delta}_{A\to B}$ \Comment{Path interpolation}
    \State $\mathbf{x}_\alpha^{\text{clip}} = g(\mathbf{z}_\alpha)$ \Comment{Decode vibe}
    \State $I_\alpha \gets 
           \text{IPAdapter}(\mathbf{x}_\alpha^{\text{clip}})$ \Comment{Generate image}
\EndFor

\end{algorithmic}
}
\end{algorithm}

%% file: figures/neg_vibe.tex
\begin{figure}[!th]
    \vspace{-2pt}
    \centering
    \includegraphics[width=1.0\columnwidth]{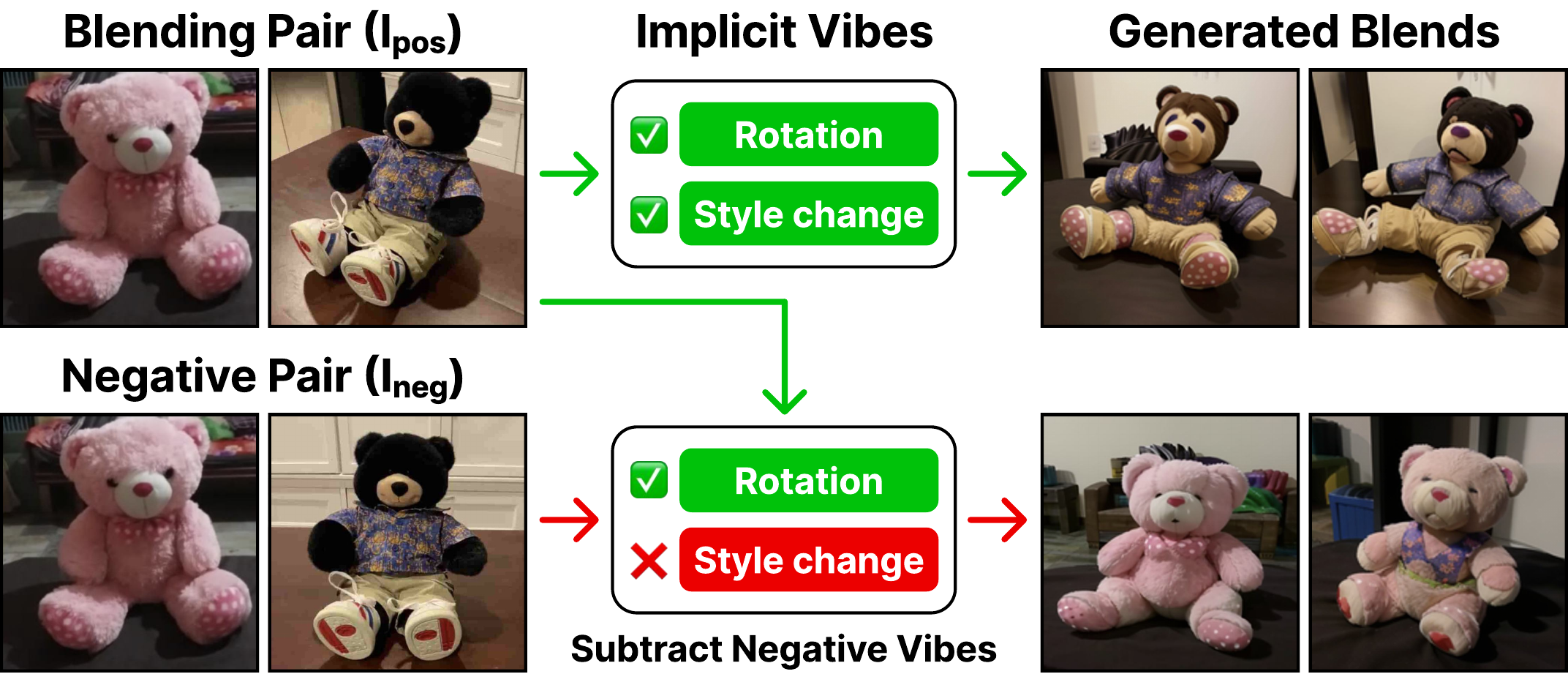}
    \vspace{-18pt}
    \caption{Negative vibe control. Vibe attributes are implicitly extracted by Vibe Space. The blending pair defines desired vibes (rotation + style). The negative pair defines vibes to suppress (style). Blending without negative examples transfers both attributes. Subtracting the negative vibe, only rotation is blended.}
    \vspace{-10pt}
    \label{fig:neg_vibes}
\end{figure}

%% file: sec/6_experiments.tex
\input{figures/human_creativity}

\section{How to Measure the Creativity of a Blend?}
\label{sec:experiments}

Measuring creativity is challenging and underdefined: what counts as ``creative'' is context-dependent and shifts as human expectations and model capabilities evolve. We take clues from cognitive psychology  \cite{beaty2023associative} to assess blend creativity, and we examine these insights through human evaluation. We also investigate how well LLMs can approximate human judgments of creativity via step-by-step reasoning. Specifically, we seek to answer the following questions:
\begin{enumerate}[noitemsep, nolistsep]
    \item How do humans rate the creativity of a blend?
    \item How to score blend difficulty using pretrained models?
    \item How do human preferences reflect blend creativity?
    \item How well do LLMs gauge the creativity of a blend?
\end{enumerate}

We construct evaluation datasets where each image pair contains distinct concepts that share visual attributes but are not spatially aligned. We curate 44 image pairs from the human-annotated Totally Looks Like dataset \cite{rosenfeld2018totally}, which exhibits humorous similarities in unrelated concepts. We also create 300 image pairs of architectural designs, where shared attributes exist but coherent blends are non-obvious.

We compare using our Vibe Space for blending images with recent multimodal LLMs: OpenAI's GPT Image 1 \cite{gpt-image-1} and Gemini 2.5 Flash Image \cite{comanici2025gemini} with step-by-step reasoning \cite{wei2022chain, peng2025probing}. We also implement ``CLIP Avg'', which averages the CLIP image embeddings of the inputs and feeds the resulting embedding into IP-Adapter \cite{ye2023ip}.

\input{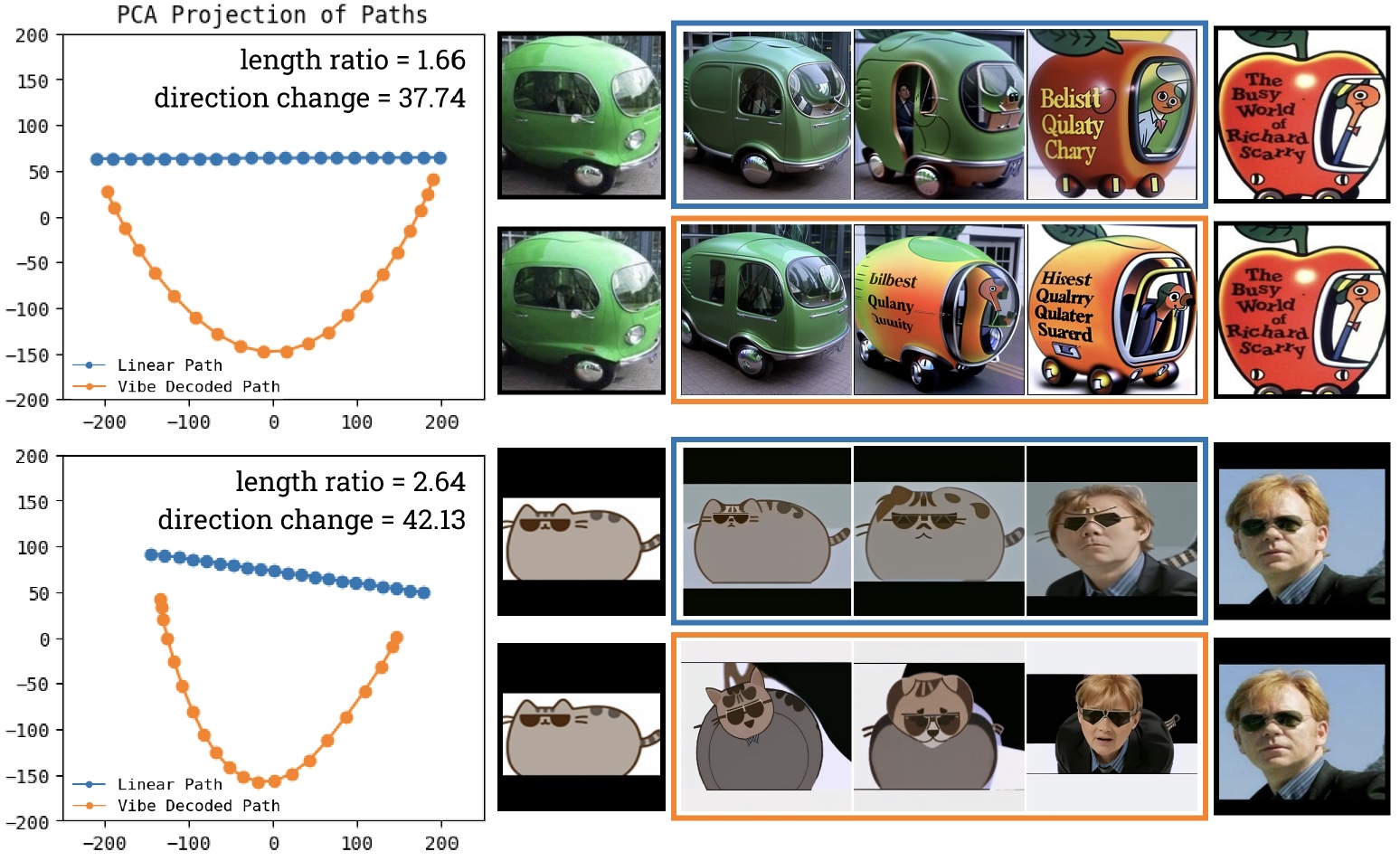}

\input{figures/blending_qualitative}

\subsection{How Do Humans Rate Blend Creativity?}

To gauge human perceptions of creativity, we ask raters to compare different image pairs along two related but distinct axes, independent of any method's output:

\begin{enumerate}[noitemsep, nolistsep]
    \item \textbf{Creative Potential:} How much the content of the image pair allows for an \textit{interesting or compelling} blend.
    \item \textbf{Blend Difficulty:} How \textit{challenging} it would be to create a coherent hybrid that fuses the shared attributes.
\end{enumerate}

These axes are informed by cognitive psychology studies suggesting that human creativity involves both attention and cognitive control to produce meaningful connections between concepts \cite{beaty2023associative, benedek2019toward}. Creative Potential reflects attention to the key visual attributes shared by the images---the ``vibe'' that makes a blend interesting or compelling---while Blend Difficulty reflects the cognitive control required to integrate these attributes into a coherent hybrid.

We conducted a user study on image pairs from Totally Looks Like \cite{rosenfeld2018totally}. Humans were presented with two image pairs at a time and asked to indicate which pair has higher Creative Potential and which has higher Blend Difficulty, or select tie if they are similar. Each image pair was compared multiple times across ten raters, and these pairwise judgments were aggregated to assign each image pair a score along both axes. The scores were discretized into low, medium, and high bins of Creative Potential and Blend Difficulty, as illustrated in \Cref{fig:human_creativity}. We observe that image pairs with higher Blend Difficulty tend to have higher Creative Potential and are generally more conceptually different.

\subsection{How to Estimate Blend Difficulty?}

Cognitive psychology suggests that creatively blending distant concepts requires humans to flexibly traverse semantic networks via intermediate associations rather than direct links \cite{beaty2023associative, kenett2016examining}. Following this insight, we hypothesize that blending conceptually distant pairs (e.g., man and banana) involves traversing longer, curved paths in pretrained feature spaces, while blending nearby pairs (e.g., cat and dog) follows simpler, approximately linear paths.

To estimate the difficulty of blending two images, we define a \textit{path nonlinearity score} (PNS) for each image pair based on how much a path decoded from Vibe Space deviates from linear interpolation in CLIP space (Fig. \ref{fig:combined_curved_paths}). Given a vibe decoded path in CLIP space $\gamma(\alpha)$ with $\alpha \in [0, 1]$, we sample $n$ equally-spaced points $\alpha_0 = 0, \alpha_1, \ldots, \alpha_{n-1} = 1$, and quantify excess path length and directional changes:

\small

\vspace{-12pt}
\begin{equation}
\hspace{-6pt}\textit{length ratio}=\frac{\gamma_{\text{curved}}}{\gamma_{\text{linear}}},
\quad
\raisebox{1.0ex}{%
\(\begin{aligned}
\gamma_{\text{curved}} &= \sum_{i=0}^{n-2} \bigl\|\gamma(\alpha_{i+1})-\gamma(\alpha_{i})\bigr\|_2, \\
\gamma_{\text{linear}} &= \bigl\|\gamma(\alpha_{n-1})-\gamma(\alpha_0)\bigr\|_2,
\end{aligned}\)
}
\label{eq:length_ratio}
\end{equation}

\vspace{-8pt}
\begin{equation}
\textit{direction change} = \frac{1}{n-2} \sum_{i=0}^{n-3} \cos^{-1}\left(\frac{\langle \boldsymbol{\delta}_i, \boldsymbol{\delta}_{i+1} \rangle}{\|\boldsymbol{\delta}_i\|_2 \|\boldsymbol{\delta}_{i+1}\|_2}\right),
\label{eq:direction_change}
\end{equation}

\normalsize 

\noindent where $\boldsymbol{\delta}_i = \gamma(\alpha_{i+1}) - \gamma(\alpha_i)$. We normalize and average the \emph{length ratio} and \emph{direction change} scores to obtain the PNS for an image pair. This metric serves as a computational proxy for conceptual distance: higher PNS means the path traverses multiple intermediate regions of feature space.

To check whether PNS reflects human-perceived Blend Difficulty, we compare two image pairs at a time and test whether the pair humans rate as harder to blend also has a higher PNS. To reduce noise, we use comparisons with high human consensus ($\geq 66\%$). In \Cref{fig:combined_curved_paths}, as humans agree more on which image pair is harder, the PNS of the human-preferred pair is correspondingly higher. We find \textbf{$80.0\%$} agreement between PNS and human-rated Blend Difficulty, indicating that PNS effectively estimates perceived difficulty and can help curate challenging image pairs.

\subsection{How Do Human Preferences Reflect Creativity?}
\label{sec:human_preferences}

\input{tables/blending_quantitative}

To compare the creativity of blends from different methods, we conduct a human preference study in which participants first identify the main attributes that are similar between images (\Cref{fig:human_creativity}), and then rank the outputs based on how well they coherently blend those attributes. We determine consensus using first-place votes, followed by second-place votes to break ties. We recruit six raters for Totally Looks Like \cite{rosenfeld2018totally} and four raters for our Architecture dataset.

As reported in \Cref{tab:blending_quantitative}, humans prefer our method 3$\times$ more often as the second-ranked GPT \cite{gpt-image-1} on high Blend Difficulty examples from Totally Looks Like \cite{rosenfeld2018totally}, and 2.4$\times$ more often on medium difficulty examples. On easier image pairs, human preference for GPT and CLIP Avg increases. This may explain the smaller margin in performance between our method and CLIP Avg on Architecture, in which the paired images are more conceptually related than in Totally Looks Like.
We also provide qualitative comparisons in \Cref{fig:blending_qualitative}. Our method effectively identifies the relevant attributes from both input images and creatively blends them, while baselines like Gemini and GPT often miss key attributes or merge them less successfully.

\input{tables/human_llm_agreement}

Humans show notable agreement on which method's blend is preferred within each pairwise comparison of methods in \Cref{tab:human_llm_agreement}. We compute ``Agreement'' as the fraction of matching judgments between two raters, and we report Cohen's $\kappa$ \cite{cohen1960coefficient} to account for chance. For instance, 76.6\% agreement on ``Gemini vs. Ours'' means, on average, raters chose the same method's blend as preferred for 76.6\% of input image pairs. We observe that agreement ranges from moderate to high (63--77\% on Totally Looks Like; 66--75\% on Architecture), demonstrating that humans are fairly consistent even on this subjective task. Both agreement and Cohen’s $\kappa$ show greater variation on Totally Looks Like than on Architecture, suggesting that humans are more consistent when evaluating blends of conceptually related images.

\subsection{How Well Do LLMs Gauge Blend Creativity?}
\label{sec:llm_preferences}

\input{figures/llm_failures}

LLMs have recently been used to approximate human judgments \cite{gu2024survey}, which can be expensive to obtain. To probe whether LLMs can assess the creativity of a blend, we use GPT-5 (``LLM'') \cite{gpt-5} with step-by-step reasoning \cite{wei2022chain, peng2025probing} to select the best blend among outputs from different methods for each input image pair. Specifically, we prompt the LLM to (1) identify the main objects and shared attributes between the image pair, (2) evaluate how well each output merges those attributes, and (3) choose the blend it judges as best, providing its reasoning at each step. We adopt this structured setup to encourage the LLM reason explicitly about the ``vibe'' of the image pair. As shown in \Cref{tab:blending_quantitative}, the LLM most frequently prefers our method (and GPT) for high Blend Difficulty examples in Totally Looks Like \cite{rosenfeld2018totally}, and consistently favors our method on Architecture.

The LLM judge offers a promising approximation of human preferences for the creativity of a blend, but it might incorrectly infer the underlying ``vibe'' in the input image pair. We observe two primary failure modes: (1) the LLM identifies the wrong shared attributes, or (2) it identifies the correct attributes but overemphasizes irrelevant ones, leading it to prefer a different blend than humans. For example, in the map--chicken pair in \Cref{fig:blending_qualitative}, the LLM focuses on attributes such as a centered subject and earthy color palette but overlooks the shared object shape. In \Cref{fig:llm_failures}, the LLM bases its preferred blend on color and texture rather than the hairstyle, but humans consistently identify the hairstyle. These observations suggest that while LLM can approximate human judgments of creative blends, improving their ability to prioritize the most relevant shared attributes between images is essential for reliable evaluation.

In \Cref{tab:human_llm_agreement}, we quantify agreement between humans and the LLM on which blend is preferred among outputs from different methods, where random chance agreement is 25\%. Using only the human most preferred blend (top-1) yields lower agreement with the LLM (35.7\% on Totally Looks Like; 31.3\% on Architecture). Expanding to the human top-2 preferred blends increases agreement to 55.1\% and 51.8\%, respectively, suggesting that the LLM tends to select a subset of blends that humans rate highly. Overall, humans are more consistent with one another, and the LLM captures a partial but meaningful overlap with human preferences.

%% file: figures/human_creativity.tex
\begin{figure*}[!th]
    \vspace{-22pt}
    \centering
    \includegraphics[trim=0in 0in 0in 0in, clip,width=\textwidth]{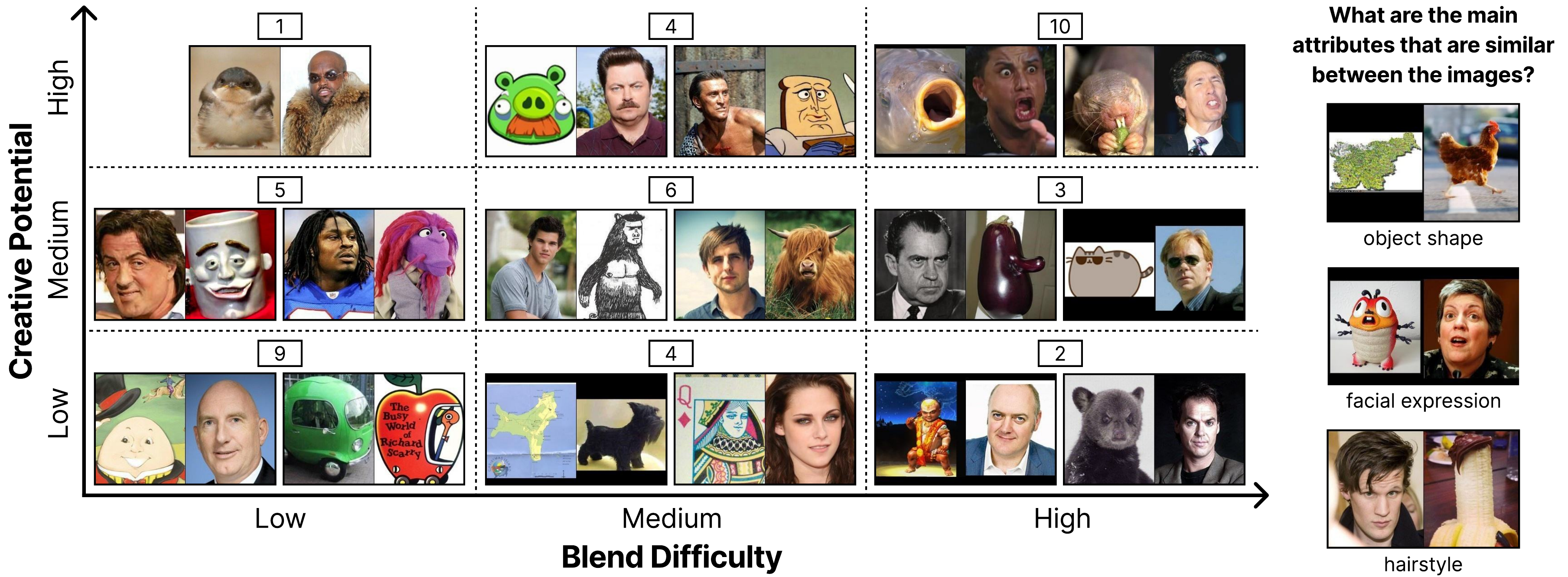}
    \vspace{-20pt}
    \caption{\textbf{Left:} To gauge human perceptions of creativity, we ask raters to compare image pairs along two axes: \emph{Creative Potential} refers to how interesting a blend might be, and \emph{Blend Difficulty} indicates how challenging it is to form a coherent blend. Image pairs with higher Blend Difficulty tend to have higher Creative Potential and are often more conceptually different. Numbers in each cell indicate the number of examples. \textbf{Right:} When evaluating blend creativity across methods, raters first identify the key shared attributes in the input images.}
    \vspace{-15pt}
    \label{fig:human_creativity}
\end{figure*}

%% file: figures/curved_paths.tex
\begin{figure}[!ht]
    \centering
    \includegraphics[width=\linewidth]{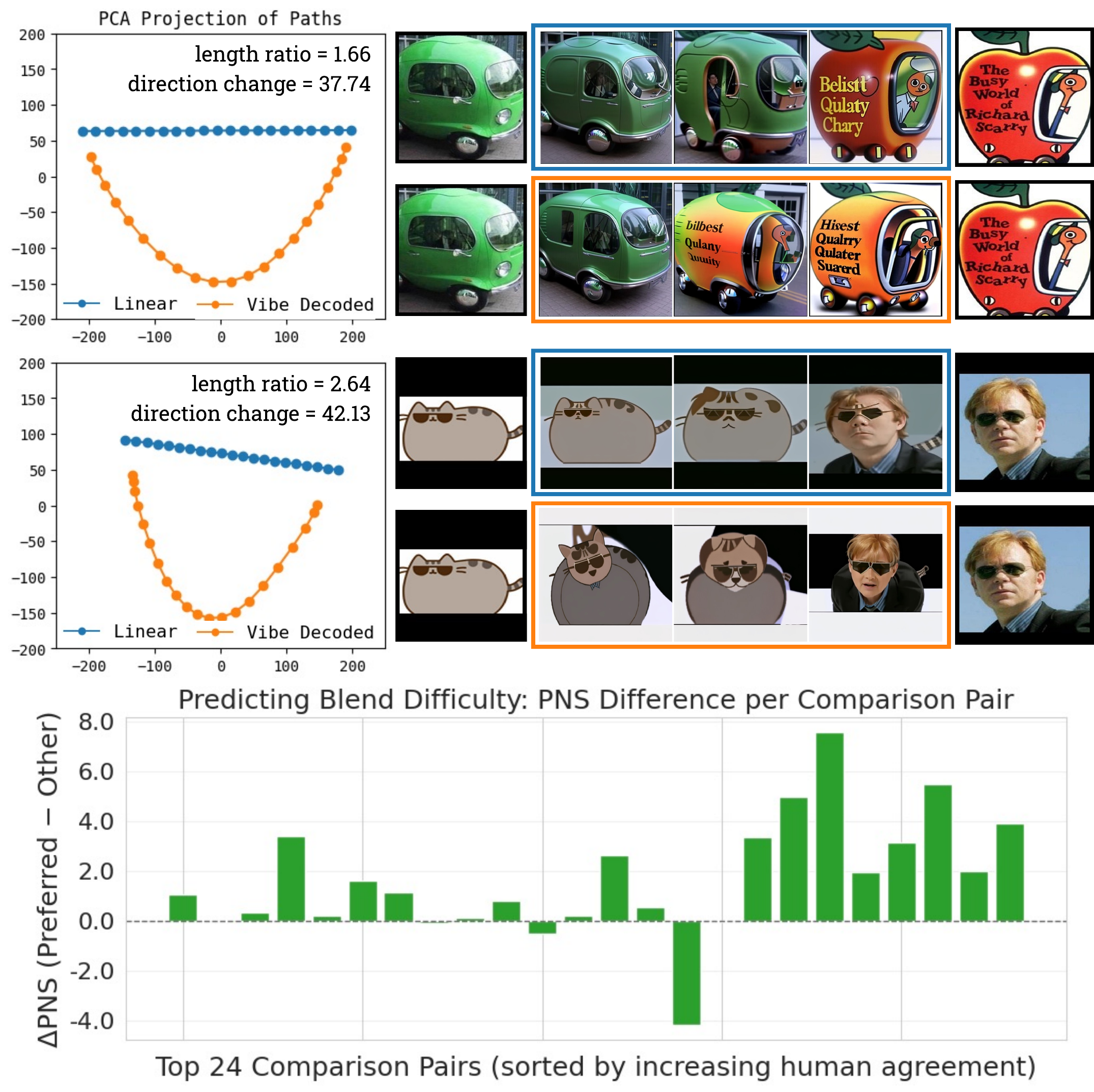}
    \vspace{-20pt}
    \caption{\textbf{Top:} Vibe-decoded (curved) vs.\ straight path in CLIP space for a \emph{single} image pair; we sample 20 points to capture curvature. Pairs that humans judge as more difficult exhibit more curvature. \textbf{Bottom:} For each \emph{comparison pair} (two different image pairs shown together), we plot $\Delta\text{PNS} = \text{PNS}_{\text{preferred}} - \text{PNS}_{\text{other}}$. Positive values indicate that PNS agrees with the human choice. Larger $\Delta\text{PNS}$ correlates with higher rater consensus. Negative values indicate disagreement with human consensus.
    }
    \vspace{-14pt}
    \label{fig:combined_curved_paths}
\end{figure}

%% file: figures/blending_qualitative.tex
\begin{figure*}[!th]
    \vspace{-20pt}
    \centering
    \includegraphics[trim=0in 0in 0in 0in, clip,width=\textwidth]{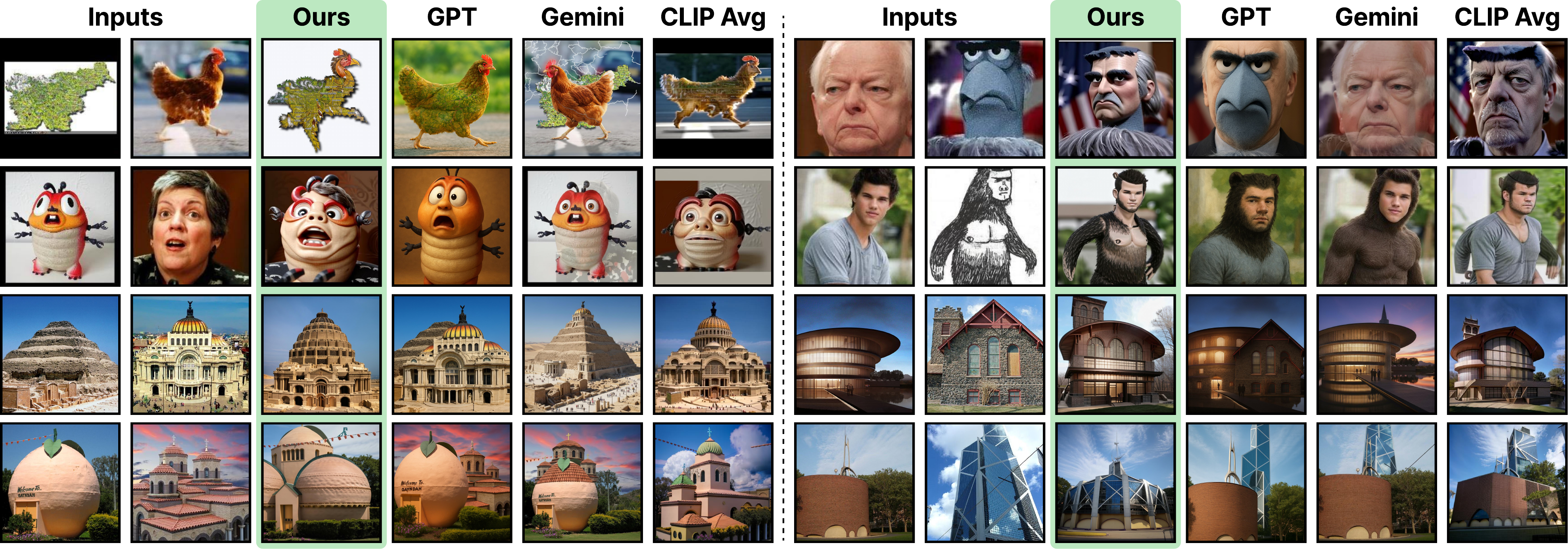}
    \vspace{-20pt}
    \caption{Our method creatively blends the relevant visual attributes that are similar between the inputs, or ``vibe''. For example, the vibe between the map and chicken is the shape, not color or texture. In contrast, the baselines often fail to recognize the key attributes or fuse them effectively. CLIP Avg refers to averaging the input image embeddings and feeding the resulting embedding into IP-Adapter \cite{ye2023ip}, akin to using weight 0.5 in CLIP linear interpolation. Gemini \cite{comanici2025gemini} and GPT \cite{gpt-image-1} often perform a part-level composition or style transfer.}
    \vspace{-13pt}
\label{fig:blending_qualitative}
\end{figure*}

%% file: tables/blending_quantitative.tex
\begin{table}[!th]
\centering
\resizebox{\columnwidth}{!}{%
\setlength{\tabcolsep}{3pt}
\begin{tabular}{lccccccccc}
\toprule

& \multicolumn{6}{c}{Totally Looks Like} \\
\cmidrule(lr){2-7} \addlinespace[2pt]

& \multicolumn{2}{c}{High Difficulty} & \multicolumn{2}{c}{Medium Difficulty} & \multicolumn{2}{c}{Low Difficulty} & \multicolumn{2}{c}{Architecture} \\
\cmidrule(lr){2-3} \cmidrule(lr){4-5} \cmidrule(lr){6-7} \cmidrule(lr){8-9} \addlinespace[2pt]

Method & Human & LLM & Human & LLM & Human & LLM & Human & LLM \\
\midrule

CLIP Avg & 13.3\% & 13.3\% & 21.4\% & 7.14\% & \underline{26.7\%} & 0.00\% & 39.0\% & 21.7\% \\ 
Gemini \cite{comanici2025gemini} & 6.67\% & 6.67\% & 7.14\% & 7.14\% & 6.67\% & 13.3\% & 5.00\% & 8.33\% \\ 
GPT \cite{gpt-image-1} & 20.0\% & \textbf{40.0\%} & 21.4\% & \textbf{50.0\%} & \textbf{40.0\%} & \textbf{66.7\%} & 14.0\% & 30.3\% \\ 
Ours & \textbf{60.0\%} & \textbf{40.0\%} & \textbf{50.0\%} & \underline{35.7\%} & \underline{26.7\%} & \underline{20.0\%} & \textbf{42.0\%} & \textbf{39.7\%} \\ 

\bottomrule
\end{tabular}
}%
\vspace{-5pt}
\caption{We report human preferences for blends from different methods using input image pairs from Totally Looks Like \cite{rosenfeld2018totally} (grouped by high, medium, and low Blend Difficulty via human ratings) and from our Architecture dataset. We also report LLM preferences and compare human and LLM judgments in \Cref{fig:llm_failures}. Our method is effective overall and most preferred by humans and the LLM on image pairs that are more difficult to blend.}
\vspace{-10pt}
\label{tab:blending_quantitative}
\end{table}

%% file: tables/human_llm_agreement.tex
\begin{table}[!th]
\vspace{-6pt}
\centering
\resizebox{\columnwidth}{!}{%
\setlength{\tabcolsep}{3pt}
\begin{tabular}{lcccc}
\toprule

& \multicolumn{2}{c}{Totally Looks Like} & \multicolumn{2}{c}{Architecture} \\
\cmidrule(lr){2-3} \cmidrule(lr){4-5} \addlinespace[2pt]

Comparison & Agreement (\%) & Cohen's $\kappa$ & Agreement (\%) & Cohen's $\kappa$ \\
\midrule

CLIP Avg vs. Gemini &  74.9 $\pm$ 6.0  &  0.23 $\pm$ 0.22  & 74.0 $\pm$ 4.5  &  0.18 $\pm$ 0.09 \\
CLIP Avg vs. GPT    &  63.1 $\pm$ 9.1  &  0.18 $\pm$ 0.16  & 70.0 $\pm$ 3.6  &  0.26 $\pm$ 0.14 \\
CLIP Avg vs. Ours   &  63.1 $\pm$ 7.9  &  0.20 $\pm$ 0.17  & 65.5 $\pm$ 3.0  &  0.31 $\pm$ 0.06 \\
Gemini vs. GPT      &  72.9 $\pm$ 9.5  &  0.33 $\pm$ 0.23  & 66.8 $\pm$ 4.7  &  0.31 $\pm$ 0.09 \\
Gemini vs. Ours     &  76.6 $\pm$ 9.1  &  0.16 $\pm$ 0.23  & 75.0 $\pm$ 4.0  &  0.17 $\pm$ 0.08 \\
GPT vs. Ours        &  ~~65.8 $\pm$ 11.1 &  0.15 $\pm$ 0.20  & 70.8 $\pm$ 4.7  &  0.26 $\pm$ 0.12 \\
\midrule

Human Top-1 vs. LLM &  35.7 $\pm$ 7.9  & -- &  31.3 $\pm$ 5.6  & -- \\
Human Top-2 vs. LLM &  55.1 $\pm$ 4.7  & -- &  51.8 $\pm$ 3.6  & -- \\

\bottomrule
\end{tabular}
}%
\vspace{-5pt}
\caption{\textbf{Top:} Mean $\pm$ standard deviation (SD) of human agreement on which method's blend is preferred within each pairwise comparison of methods. Reported as fraction of matching judgments (``Agreement'') and Cohen's $\kappa$ \cite{cohen1960coefficient} ($-1$ = perfect disagreement; $1$ = perfect agreement). For example, 76.6\% agreement on ``Gemini vs. Ours'' indicates that, on average, humans chose the same method's blend as preferred for 76.6\% of input image pairs. \textbf{Bottom:} Mean $\pm$ SD of agreement between human top-1 / top-2 preferred blends and the LLM-selected blend, where top-2 counts a match if the LLM's choice ranks first or second for the human.}
\vspace{-12pt}
\label{tab:human_llm_agreement}
\end{table}

%% file: figures/llm_failures.tex
\begin{figure}[!h]
    \vspace{-13pt}
    \centering
    \includegraphics[trim=0in 0in 0in 0in, clip,width=\columnwidth]{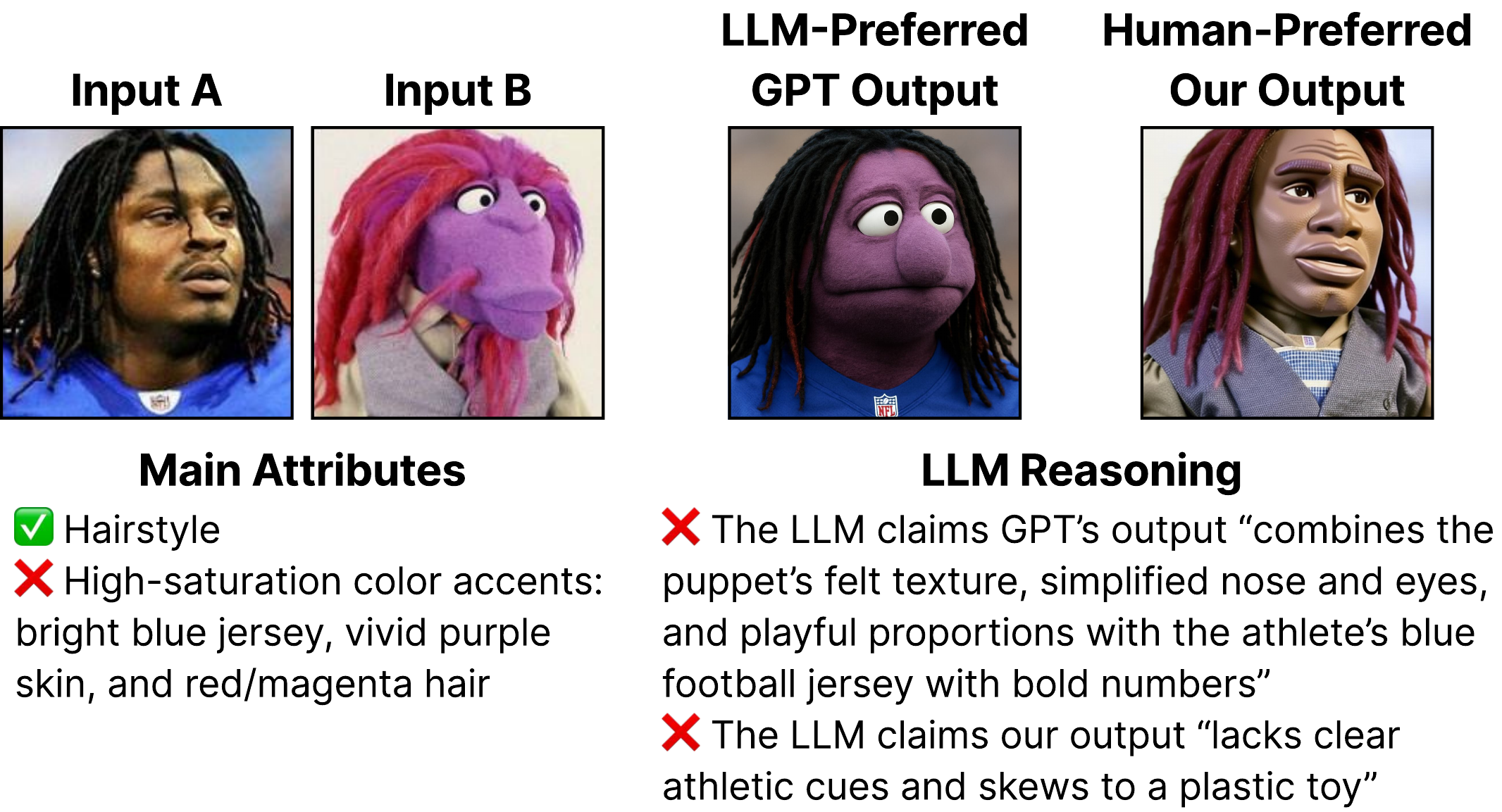}
    \vspace{-20pt}
    \caption{LLM judges offer a useful approximation of human preferences but can exhibit failure modes. The LLM correctly identifies the key attribute as the hairstyle. However, the LLM also references irrelevant attributes like color and bases its preference on texture and body composition rather than focusing on hairstyle.}
    \vspace{-9pt}
    \label{fig:llm_failures}
\end{figure}

%% file: sec/7_conclusion.tex
\section{Conclusion}
\label{sec:conclusion}

Creatively blending visual concepts involves identifying the most relevant shared attributes for an input pair and merging them into a coherent and meaningful hybrid, which we call Vibe Blending. We propose Vibe Space, a hierarchical graph manifold that guides geodesic paths in ambient feature spaces like CLIP to produce such blends.
To evaluate the creativity of a blend, we introduce a cognitively-inspired framework using human and LLM preferences, and a path nonlinearity score for blend difficulty based on path geometry in feature space. Our method generates blends that humans prefer more often than those from strong baselines like Gemini and GPT, especially on difficult image pairs.

Curating compelling and challenging input image pairs remains an open problem. Our path nonlinearity score provides a principled approach to identifying such examples, offering a direction for dataset expansion. Another promising direction is improving LLM judges to better attend to the key visual attributes that underlie creative blending, enabling more reliable and scalable evaluation in future work.

\section*{Acknowledgments}
This work is supported by the funds provided by the National Science Foundation and by DoD OUSD (R\&E) under Cooperative Agreement PHY-2229929 (The NSF AI Institute for Artificial and Natural Intelligence).

%% file: sec/X_appendix.tex
\clearpage
\setcounter{page}{1}
\maketitlesupplementary

\appendix

\input{sec_appendix/baseline_comparison}

\input{sec_appendix/ablations}

\input{sec_appendix/path_finding}

\section{Vibe Space Details}
\label{appsec:vibespace}

\subsection{Correspondence Matching}
\label{sec:appendix_correspondence}
To ensure that the ``vibe'' transition preserves semantic structure (e.g., blending a head to another head rather than to a background element), we establish a semantic correspondence $\pi$ between the input images. This corresponds to the `Match' function in Algorithms 1 and 2. Since pixel-wise matching is computationally expensive and noisy, we operate on semantic regions derived from DINO features.

\begin{figure}[h]
    \centering
    \includegraphics[width=1\linewidth]{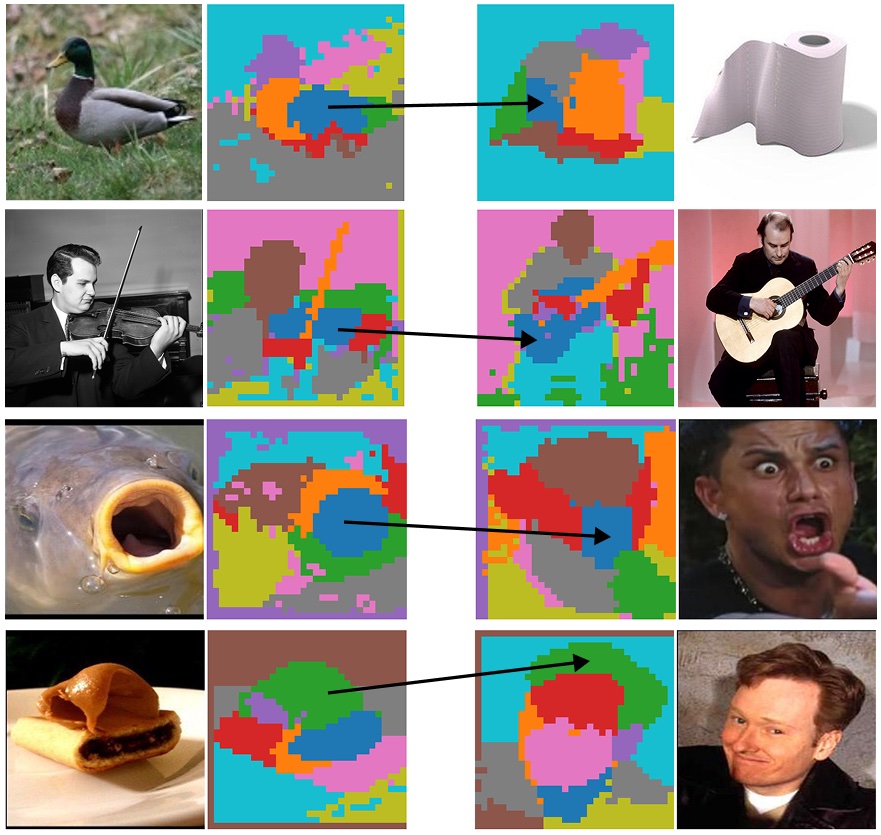}
    \caption{Examples of correspondence matching. We use NCut clustering \cite{ncut-pytorch} and Hungarian matching to compute correspondence between segments.}
    \label{appfig:correspondence-example}
\end{figure}

As shown in \Cref{appfig:correspondence-example}, we first segment each image into $k$ distinct semantic regions using NCut \cite{ncut-pytorch}. Given the DINO feature tokens $\mathbf{x}^{\text{dino}}$, we construct a token-wise affinity graph and compute the leading eigenvectors of the graph Laplacian. We then discretize these eigenvectors into segmentation masks using a $k$-way clustering approach. This results in a set of masks $\{\text{Mask}_1, \dots, \text{Mask}_k\}$ for each image, grouping visually and semantically similar patches.

For each identified segment $i$, we compute a representative feature centroid $\mathbf{c}_i$. This is calculated by averaging the DINO feature embeddings of all patch tokens belonging to that segment:
\begin{equation}
\mathbf{c}_i = \frac{1}{|\text{Mask}_i|} \sum_{p \in \text{Mask}_i} \mathbf{x}^{\text{dino}}_p,
\end{equation}
where $p$ represents the token indices within the mask $\text{Mask}_i$.

To define the correspondence mapping $\pi$, we formulate the problem as a linear assignment task. We compute a cost matrix $C \in \mathbb{R}^{k \times k}$ representing the pairwise feature distances between the cluster centroids of the source image $I_A$ and the target image $I_B$:
\begin{equation}
C_{ij} = \| \mathbf{c}_i^{(A)} - \mathbf{c}_j^{(B)} \|_2.
\end{equation}
We apply the Hungarian algorithm to find the optimal bijection that minimizes the total feature distance between matched segments. This mapping $\pi$ allows us to compute the vibe displacement $\Delta_{A \to B}$ relative to the corresponding semantic structures in the Vibe Space.

\input{figures/more_analogies}
\subsection{Vibe Analogy}
\label{sec:appendix_analogy}

\input{sec_appendix/algo_analogy}

Vibe Analogy applies the semantic transformation observed between a reference pair ($I_A, I_B$) to a new subject $I_{A'}$, as shown in \Cref{fig:more_analogies}. The procedure is detailed in Algorithm \ref{alg:vibe-analogy}.

We first compute the Vibe Space for all three images $I_A, I_B, I_{A'}$ simultaneously (Lines 1-5). We extract DINO features $\mathbf{x}^{\text{dino}}$, construct the joint affinity graph, and train the encoder-decoder pair $(f, g)$ to map between the shared spectral manifold and CLIP space. We then encode all three inputs into the Vibe Space to obtain $\mathbf{z}_A, \mathbf{z}_B$, and $\mathbf{z}_{A'}$.

To transfer the ``vibe'' from the reference pair to the new subject, we must establish a correspondence chain (Lines 6-7). Using the matching procedure defined in \Cref{sec:appendix_correspondence}, we compute two distinct mappings:
(1) $\pi_{A \leftrightarrow B}$ matches semantic segments in the reference source $I_A$ to the reference target $I_B$.
(2) $\pi_{A \leftrightarrow A'}$ matches semantic segments in the new subject $I_{A'}$ to the reference source $I_A$.

We define the ``vibe'' as the displacement field $\boldsymbol{\Delta}_{A \to B}$ between semantically corresponding regions in the reference pair. For a cluster centroid $\mathbf{c}_i^{(A)}$ and its match $\mathbf{c}_j^{(B)}$ (where $j = \pi_{A \leftrightarrow B}(i)$), the local displacement is $\mathbf{v}_i = \mathbf{c}_j^{(B)} - \mathbf{c}_i^{(A)}$.

To transfer this vibe to $I_{A'}$, we use the mapping $\pi_{A \leftrightarrow A'}$ to pull the appropriate displacement vector for each segment in $A'$ (Lines 8-9). For a segment $k$ in $I_{A'}$, if it maps to segment $i$ in $I_A$ (i.e., $i = \pi_{A \leftrightarrow A'}(k)$), we assign it the displacement $\mathbf{v}_i$. This constructs the target displacement field $\boldsymbol{\Delta}_{A' \to B'}$, ensuring that the specific semantic transformation (e.g., ``turning a face into a flower'') is applied to the correct anatomical regions of the new subject.

Finally, we generate the output image $I_{B'}$ by applying the transferred displacement field to the latent code of the new subject (Lines 10-13): $\mathbf{z}_{\alpha} = \mathbf{z}_{A'} + \alpha \boldsymbol{\Delta}_{A' \to B'}$. This latent path is decoded by $g$ into CLIP space and rendered via IP-Adapter \cite{ye2023ip}.

\subsection{Negative Vibe Control}
\label{sec:appendix_negative_vibe}
\input{figures/neg_vibes_extended}

\paragraph{The Principle of Vibe Subtraction.} Conceptually, our goal is to ``subtract'' the negative vibe from the positive vibe. In the geometric framework of Vibe Space, both positive and negative vibes can be represented as subspaces spanned by the eigenvectors of their respective graph Laplacians. The positive subspace, spanned by $\mathbf{\Psi}_{\text{pos}}$, contains the desired attributes, while the negative subspace, spanned by $\mathbf{\Psi}_{\text{neg}}$, contains the unwanted ones.

It is highly probable that these subspaces are not orthogonal; they share common components. For instance, if the positive vibe is ``rotation'' and ``style change'' (\Cref{fig:neg_vibes}), and the negative vibe is ``style change,'' both subspaces will contain eigenvectors related to texture, color palette, and artistic rendering. To isolate ``rotation,'' we must remove the ``style change'' components from the positive vibe's basis.

We achieve this by orthogonalizing the positive basis vectors against the negative basis vectors. This process effectively projects $\mathbf{\Psi}_{\text{pos}}$ onto a new basis, $\mathbf{\Psi}_{\text{filtered}}$, that is orthogonal to the subspace of $\mathbf{\Psi}_{\text{neg}}$. The mathematical formulation for this operation in the flag-space hierarchy is:
\begin{equation}
\mathbf{\Psi}_{\text{filtered}} = \mathbf{\Psi}_{\text{pos}} - \beta \cdot \mathbf{\Psi}_{\text{neg}} \left( (\mathbf{\Psi}_{\text{neg}})^\top \mathbf{\Psi}_{\text{pos}} \right).
\label{eq:vibe_subtraction}
\end{equation}
Here, we assume the eigenvectors forming the columns of $\mathbf{\Psi}_{\text{neg}}$ are orthonormal. The term $\mathbf{\Psi}_{\text{neg}} ((\mathbf{\Psi}_{\text{neg}})^\top \mathbf{\Psi}_{\text{pos}})$ thus represents the orthogonal projection of the positive eigenvectors onto the subspace spanned by the negative eigenvectors. By subtracting this projection, we are left with the component of the positive vibe that is orthogonal to the negative vibe. The hyperparameter $\beta$ controls the strength of this subtraction: $\beta=1$ performs full orthogonalization, while $\beta > 1$ can be used to actively push the resulting vibe away from the negative attributes.

\paragraph{Learning a Filtered Vibe Space.} To incorporate negative vibes into our framework, we modify the training objective of the Vibe Space autoencoder. Instead of training the encoder $f$ to match the geometry of the positive flag-space kernel $S(\mathbf{\Psi}(\mathbf{x}_{\text{pos}}))$, we train it to match the kernel of the \textit{filtered} basis, $S(\mathbf{\Psi}_{\text{filtered}})$.

The flag-space encoder loss from Eq.~(4) becomes:
\begin{equation}
\mathcal{L}_{\text{flag\_enc}}(f) = \| \mathbf{z}\mathbf{z}^\top - S(\mathbf{\Psi}_{\text{filtered}}) \|_F^2,
\label{eq:filtered_loss}
\end{equation}
where $\mathbf{z} = f(\mathbf{x}_{\text{pos}})$. The encoder $f$ is trained only on the positive exemplars $\mathbf{x}_{\text{pos}}$, but the target geometry $S(\mathbf{\Psi}_{\text{filtered}})$ is now computed from both the positive ($\mathbf{\Psi}_{\text{pos}}$) and negative ($\mathbf{\Psi}_{\text{neg}}$) eigenvector sets. \Cref{fig:neg_vibes_extended} shows an example of a vibe blending path with and without negative exemplars. 

\input{sec_appendix/algo_negative_vibe}
\input{figures/grad_channels}

\subsection{Vibe Visualization by Gradient}

To understand which specific visual attributes drive the geometric alignment in Vibe Space, we propose a gradient-based visualization method. Since our framework relies on DINO features for semantic correspondence and graph construction, we identify the specific feature channels within the high-dimensional DINO embedding that are most responsible for a specific semantic region or ``vibe''. 

We leverage the differentiability of the spectral clustering process. Let $\mathbf{X} \in \mathbb{R}^{N \times C}$ denote the extracted DINO feature tokens for an images, where $N$ is the number of patches and $C$ is the feature dimension. We compute the generalized eigenvectors $\mathbf{\Psi}$ of the graph Laplacian $\mathbf{L}$. 

To visualize the features corresponding to a specific semantic cluster $k$ (e.g., the foreground object), we first obtain the discretized cluster indicator $\mathbf{y}_k \in \{0, 1\}^N$ from the eigenvectors via $k$-way Ncut \cite{1238361}. We the define a maximization objective $\mathcal{L}_{\text{vis}}$ to identify the feature channels contributing to this cluster:

\begin{equation}
    \mathcal{L}_{\text{vis}} = - \frac{1}{|\text{Mask}_k|} \sum_{i \in \text{Mask}_k} |\mathbf{\Psi}_{i, k}|,
\end{equation}

where $\text{Mask}_k$ is the set of indices belonging to cluster $k$, and $\mathbf{\Psi}_{\cdot, k}$ is the eigenvector corresponding to that cluster. We treat the input features $\mathbf{X}$ as a learnable parameter with respect to this loss. By computing the gradient $\nabla_{\mathbf{X}} \mathcal{L}_{\text{vis}}$ via backpropagation through the eigendecomposition, we obtain a saliency map $\mathbf{G} \in \mathbb{R}^{N \times C}$:

\begin{equation}
    \mathbf{G} = \nabla_{\mathbf{X}} \mathcal{L}_{\text{vis}}.
\end{equation}

We then aggregate the gradients across the spatial dimensions of the cluster mask to score each feature channel $c$:

\begin{equation}
s_c = \frac{1}{|\text{Mask}_k|} \sum_{i \in \text{Mask}_k} \mathbf{G}_{i, c}.
\end{equation}

The channels with the highest scores $s_c$ represent the specific DINO feature dimensions encoding the attribute. We visualize these top-ranked channels as heatmaps as shown in \Cref{fig:grad_channels}.

\input{figures/clip_dip_plot}
\input{figures/clip_dip}

\subsection{Vibe Blending: Selecting the Optimal Blend Weight $\alpha$ via Dip in CLIP Consistency}
\label{sec:appendix_alpha_selection}

Selecting a single representative frame from the generated continuous blend path is non-trivial. A naive choice, such as the midpoint $\alpha=0.5$, often fails to capture the most compelling hybrid. This stems from the inherent asymmetry of our correspondence-based interpolation: we compute a semantic displacement field $\boldsymbol{\Delta}_{A \to B}$ and add it to the token clusters of the source image $I_A$. Since the transformation is anchored to the source image's semantic structure, the conceptual transition is not perfectly linear with respect to $\alpha$. Consequently, the point of optimal blending may shift away from $\alpha=0.5$, requiring an adaptive selection strategy.

To address this, we propose an automated procedure to select the optimal $\alpha$ by identifying the point of greatest conceptual transition along the blend path. Our intuition is that the most challenging point for the generative model to render often represents the most novel synthesis of attributes.

We formalize this procedure in Algorithm~\ref{alg:alpha-selection}, which measures the consistency between an ``ideal'' path decoded from Vibe Space and a ``realized'' path obtained by re-encoding the generated images. The point of minimum consistency, or the ``dip'' in the score, identifies our target $\alpha^*$.

\begin{algorithm}[h]
\caption{Optimal Blend Selection ($\alpha^*$)}
\label{alg:alpha-selection}
\small{
\begin{algorithmic}[1]
\Require Vibe decoder $g$, vibe endpoints $\mathbf{z}_A, \mathbf{z}_B$, source image $I_A$, candidate steps $A_{\text{steps}} \subset [0,1]$
\Ensure Optimal interpolation weight $\alpha^*$

\State $S_{\text{list}} \gets []$ \Comment{Initialize list to store scores}
\For{$\alpha \in A_{\text{steps}}$}
    \State $\mathbf{z}_\alpha \gets (1-\alpha) \mathbf{z}_A + \alpha \mathbf{z}_B$ \Comment{Interpolate in Vibe Space}
    \State $\mathbf{x}_{\alpha}^{\text{ideal}} \gets g(\mathbf{z}_\alpha)$ \Comment{\textbf{Ideal Path}: Decode to CLIP space}
    
    \State $I_\alpha \gets \text{IPAdapter}(\mathbf{x}_{\alpha}^{\text{ideal}})$ \Comment{Generate image}
    \State $\mathbf{x}_{\alpha}^{\text{realized}} \gets \text{CLIP}(I_\alpha)$ \Comment{\textbf{Realized Path}: Re-encode image}

    \State $\pi_\alpha, \mathcal{C}_A^{\text{down}}, \mathcal{C}_\alpha^{\text{down}} \gets \text{ComputeCorrespondence}(I_A, I_\alpha)$ \Comment{\textbf{Correspondence}: See Appendix~\ref{sec:appendix_correspondence}}
    
    \State $S_\alpha \gets \text{ComputeConsistency}(\mathbf{x}_{\alpha}^{\text{ideal}}, \mathbf{x}_{\alpha}^{\text{realized}}, \pi_\alpha, \mathcal{C}_A^{\text{down}})$ \Comment{\textbf{Score}: See Eq.~\ref{eq:consistency_score}}
    \State $S_{\text{list}}.append(S_\alpha)$
\EndFor
\State $i^* \gets \arg \min(S_{\text{list}})$ \Comment{\textbf{Identify Dip}: Find index of min score}
\State $\alpha^* \gets A_{\text{steps}}[i^*]$
\State \Return $\alpha^*$
\end{algorithmic}
}
\end{algorithm}

The core of this procedure involves generating and comparing two conceptual paths for each candidate $\alpha$. As shown in Algorithm~\ref{alg:alpha-selection} (lines 3-6), the \textbf{ideal path} is formed by decoding the Vibe Space interpolation directly into CLIP space, representing the intended manifold-respecting trajectory. The \textbf{realized path} is formed by generating a pixel-space image with the IP-Adapter and then re-encoding it with CLIP, representing what the model actually produces.

A direct token-wise comparison of these paths is brittle to minor spatial shifts. Therefore, we establish a robust semantic correspondence between the source image $I_A$ and each generated image $I_\alpha$ (line 7). We use DINO features to perform spectral clustering on both images, segmenting them into meaningful regions. The Hungarian algorithm is then used to find an optimal matching $\pi_\alpha$ between the cluster centers.
With the correspondence established, we compute the consistency score $S_\alpha$ (line 8). For each semantic region $c$ in the source image, we calculate its mean CLIP feature vector along both the ideal path, $\mu^{ideal}_c$, and the corresponding realized path, $\mu^{realized}_c$:
\begin{align*}
    \mu^{ideal}_c(\alpha)   &= \text{mean}(\mathbf{x}_{\alpha}^{\text{ideal}}[c]) \\
    \mu^{realized}_c(\alpha) &= \text{mean}(\mathbf{x}_{\alpha}^{\text{realized}}[\pi_\alpha(c)])
\end{align*}
The overall score is the average cosine similarity over all matched cluster pairs:
\begin{equation}
\label{eq:consistency_score}
    S(\alpha) = \frac{1}{|\mathcal{C}_A|} \sum_{c \in \mathcal{C}_A} \cos\left( \mu^{ideal}_c(\alpha), \mu^{realized}_c(\alpha) \right)
\end{equation}
where $\mathcal{C}_A$ is the set of source clusters and $\mathbf{x}[c]$ denotes the CLIP tokens belonging to cluster $c$.

Finally, the optimal interpolation weight $\alpha^*$ is selected as the one that minimizes this consistency score (lines 10-11). This ``dip'' in consistency marks the point of greatest semantic tension and, frequently, the most creative and compelling synthesis of the two concepts (\Cref{fig:clip_dip_plot}). \Cref{fig:clip_dip} shows examples of selecting $\alpha^*$ across a blend trajectory.

\subsection{Vibe Blending: Training with Many Images}

Vibe Space can be trained with more than two images, and using additional related images helps in identifying the main attributes. \Cref{alg:vibe-extra-image} outlines the steps for training Vibe Space with multiple additional images. The key difference between \Cref{alg:vibe-path} and \Cref{alg:vibe-extra-image} is that the latter uses more images to solve the graph diffusion map $\mathbf{\Psi}$ and train the Vibe Space. However, the blending process is still executed using two images.

\begin{figure*}[h]
    \centering
    \includegraphics[width=0.9\linewidth]{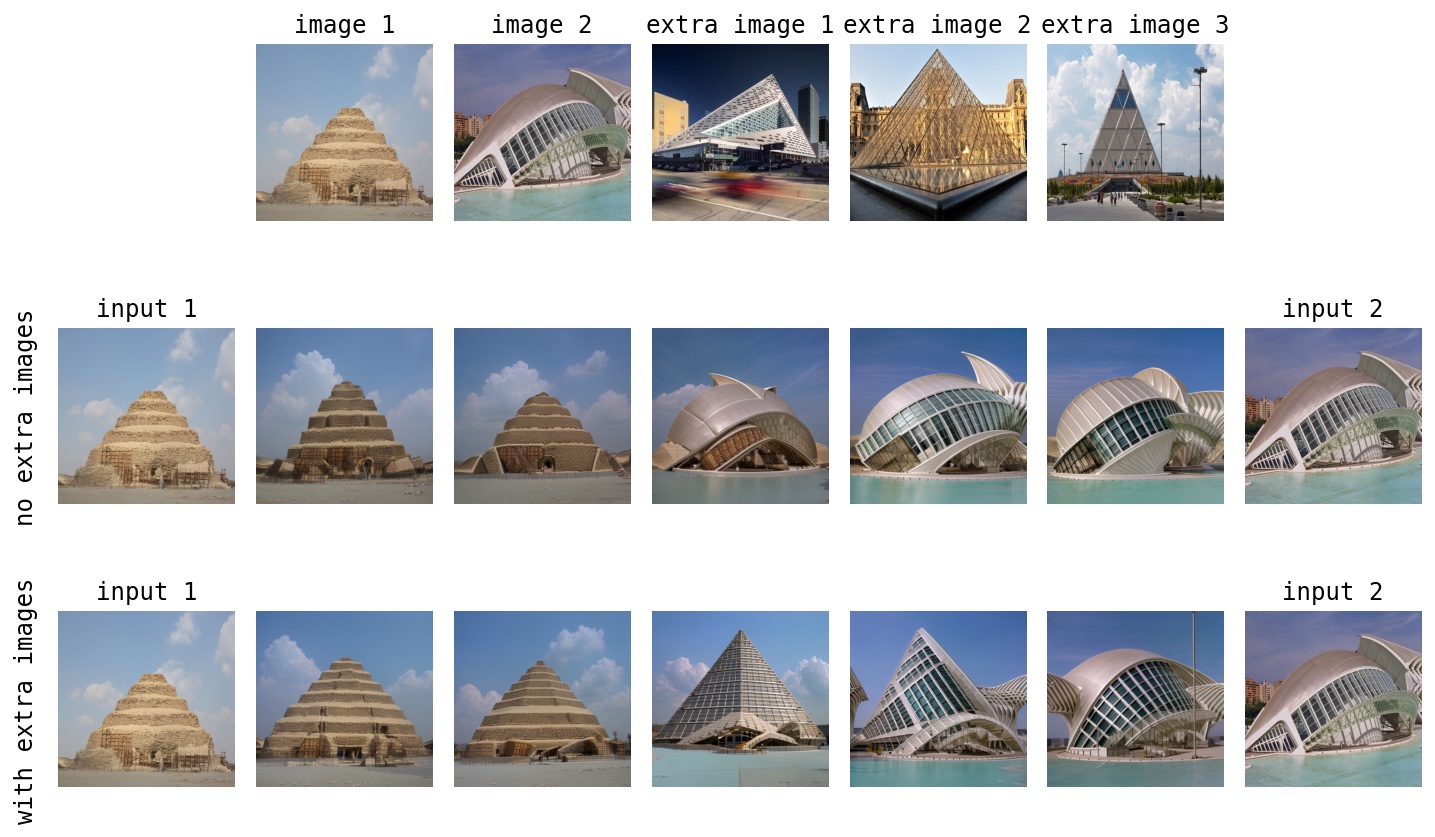}
    \caption{Vibe Blending with extra image training. We compare training Vibe Space with only two images (middle row) and with five images (bottom row), input images and extra images are in the first row. Training with extra glass window images helps capture the ``glass window'' vibe, resulting in a glass texture pyramid in the middle-blend image.}
    \label{fig:vibe-blending-extra-image}
\end{figure*}

\Cref{fig:vibe-blending-extra-image} presents qualitative examples comparing training with only two images to training with additional images. The middle blend, when additional images are included, successfully captures the ``glass window'' vibe when trained with extra images featuring glass windows.

\begin{algorithm}[h]
\caption{Vibe Blending with extra images}
\label{alg:vibe-extra-image}
\small{
\begin{algorithmic}[1]
\Require Multiple images $(I_A, I_B, \dots)$, image features for all images $(\mathbf{x}^{\text{dino}}, \mathbf{x}^{\text{clip}})$
\Ensure Generated intermediate images $\{I_\alpha\}_{\alpha\in[0,1]}$

\State $\mathbf{W}_{ij} = \text{exp}{(-\frac{\|\mathbf{x}_i^{\text{dino}} - \mathbf{x}_j^{\text{dino}}\|^2}{\sigma^2})}$, $\mathbf{D}_{ii} = \sum_j \mathbf{W}_{ij}$ \Comment{Graph}

\State $(\mathbf{D} - \mathbf{W})\mathbf{\Psi}(\mathbf{x}^{\text{dino}}) = \lambda \mathbf{D}\mathbf{\Psi}(\mathbf{x}^{\text{dino}})$ \Comment{Graph Diffusion Map}

\State $f, g \gets \text{Train}(\mathbf{x}^{\text{clip}}, \mathbf{x}^{\text{dino}}, \mathbf{\Psi}(\mathbf{x}^{\text{dino}}))$  \Comment{Train Vibe Space (all images)}

\State $\mathbf{z}_A = f(\mathbf{x}_A^{\text{dino}})$; $\mathbf{z}_B = f(\mathbf{x}_B^{\text{dino}})$ \Comment{Encode vibe (two images)}

\State $\pi \gets \text{Match}(\mathbf{x}_A^{\text{dino}}, \mathbf{x}_B^{\text{dino}})$ \Comment{Cluster correspondence}

\State $\boldsymbol{\Delta}_{A\to B} = \pi(\mathbf{z}_B) - \mathbf{z}_A$ \Comment{Path direction}

\For{$\alpha \in [0,1]$}
    \State $\mathbf{z}_\alpha = 
           \mathbf{z}_A + \alpha\,\boldsymbol{\Delta}_{A\to B}$ \Comment{Path interpolation}
    \State $\mathbf{x}_\alpha^{\text{clip}} = g(\mathbf{z}_\alpha)$ \Comment{Decode vibe}
    \State $I_\alpha \gets 
           \text{IPAdapter}(\mathbf{x}_\alpha^{\text{clip}})$ \Comment{Generate image}
\EndFor

\end{algorithmic}
}
\end{algorithm}

\input{figures/extrapolate}
\subsection{Extrapolating Vibe Blending Paths}
Since our framework constructs a locally linear manifold, we are not limited to interpolation within the convex hull of the input images ($\alpha \in [0, 1]$). We can perform \textit{extrapolation} by setting the interpolation weight $\alpha > 1$, effectively continuing along the geodesic path defined by the displacement vector $\boldsymbol{\Delta}_{A \to B}$. We observe that this can yield an exaggerating effect, where specific visual characteristics are amplified beyond their presence in the target image $I_B$ (\Cref{fig:extrapolate}).

\input{figures/dog_ram_horse}
\subsection{N-Image Blending}

We extend our framework to blend $N$ images $\{I_1, \dots, I_N\}$ simultaneously, enabling applications such as barycentric interpolation within a triangle of concepts (\Cref{fig:n_image_triangle}).

Unlike pairwise blending, establishing consistent multi-way correspondence is non-trivial. We designate one input as the base image ($I_{\text{base}}$), which serves as the structural anchor for the blend. We then compute $N-1$ pairwise correspondence mappings, aligning the semantic clusters of every other image $I_k$ to the clusters of the base image:
$$ \pi_{k \to \text{base}} = \text{Match}(\mathbf{x}^{\text{dino}}_{\text{base}}, \mathbf{x}^{\text{dino}}_k) $$
where $\pi_{k \to \text{base}}$ maps a cluster index in the base image to the corresponding cluster index in image $k$.

With semantic correspondences established across all inputs, we generalize the path interpolation logic to a multi-way vector summation. Let $\{\alpha_1, \dots, \alpha_N\}$ be the scalar interpolation weights for the $N$ images. We generate the blended latent code $\mathbf{z}_{\text{blend}}$ by starting with the base embedding and accumulating the weighted semantic displacements from all other images.

For a specific semantic region (cluster) $i$ in the base image, let $\mathbf{c}_i^{(\text{base})}$ be its centroid in Vibe Space. For every other image $k$, the centroid of the corresponding matched cluster is $\mathbf{c}_{j}^{(k)}$, where $j = \pi_{k \leftrightarrow \text{base}}(i)$. The blending equation for tokens belonging to region $i$ is:
$$ \mathbf{z}_{\text{blend}}[i] = \mathbf{z}_{\text{base}}[i] + \sum_{k=1}^{N} \alpha_k \left( \mathbf{c}_{\pi_{k \leftrightarrow \text{base}}(i)}^{(k)} - \mathbf{c}_{i}^{(\text{base})} \right) $$
This formulation computes a weighted average of the ``vibe'' displacements relative to the base structure. If $\alpha_k=1$ (and all others 0), the term simplifies to shifting the base embedding by the exact vector difference between the base and target $k$, approximating a full morph to image $k$. The resulting $\mathbf{z}_{\text{blend}}$ is decoded into CLIP space and rendered via the IP-Adapter \cite{ye2023ip}.

\input{sec_appendix/implementation_details}

\section{Additional Evaluation Details}
\label{sec:evaluation_details}

\subsection{LLMs for Generating and Evaluating Blends}

Vibe Blending involves reasoning about the relevant visual attributes shared between images, and then coherently fusing these attributes. Accordingly, we compare our method with multimodal LLMs that exhibit strong visual reasoning capabilities, such as Gemini 2.5 Flash Image \cite{comanici2025gemini} and GPT Image 1 \cite{gpt-image-1}. To promote their reasoning for this task, we use chain-of-thought prompting \cite{wei2022chain} inspired by Peng et al. \cite{peng2025probing}. Specifically, we ask Gemini to (1) identify the main objects in each input image, (2) recognize the main visual attributes that are similar between the images, and (3) generate an image that creatively blends these attributes, as shown in \Cref{fig:gemini_blend}. Since the OpenAI API does not provide a public model that supports multi-turn conversations with both image and text inputs---and image and text outputs---we combine these steps into a single prompt for GPT \cite{gpt-image-1}. The full prompts for generating blends are presented below. The parentheses indicate ($\text{inputs} \to \text{outputs}$) where $I_A$ and $I_B$ are input images, $I_{blend}$ is the blended output, T denotes text, and $\varnothing$ denotes no additional inputs besides the prompt.

\input{prompts/prompt_gemini_blend}
\input{prompts/prompt_gpt_blend}

\input{figures/gemini_blend}

We also employ an LLM to judge the output blends from different methods using a similar step-by-step reasoning approach. We ask GPT-5 \cite{gpt-5} to (1) identify the main objects in each input image; (2) recognize the main visual attributes that are similar between the images; (3) for each output, assess how well it blends the main attributes; and (4) choose the best blended output and provide reasoning. An abbreviated example chat is shown in \Cref{fig:llm_eval}. The full prompt for evaluating blends is provided below.

\input{prompts/prompt_llm_judge}

\input{figures/llm_eval}

\Cref{fig:human_llm_agreement} illustrates examples where human raters and the LLM judge agree and disagree on the best blended output. We observe that common LLM failure cases include identifying extra irrelevant attributes such as body pose, color palette, and image composition.

\input{figures/human_llm_agreement}

\subsection{User Study Details}
\label{sec:user_study_details}

\input{figures/full_human_creativity}

\paragraph{Human-Rated Creative Potential vs. Blend Difficulty.}

\input{figures/user_study_form}

We conducted a user study with ten human participants to compare 44 image pairs sourced from the Totally Looks Like dataset \cite{rosenfeld2018totally}. \Cref{fig:user_study_form} shows the prompt to users and examples with high rater agreement. Participants were presented with two image pairs side-by-side and asked to select which of the two pairs exhibited a higher Creative Potential and which exhibited a higher Blend Difficulty. They performed a total of 110 side-by-side comparisons, ensuring each of the 44 image pairs was included in 5 unique comparisons. We then fit a Bradley-Terry model \cite{bradley1952rank} to these comparison ratings to produce a robust ranking of Creative Potential and Blend Difficulty across image pairs. Finally, we used these continuous Bradley-Terry scores to categorize the image pairs into low, medium, and high bins, as depicted in \Cref{fig:full_human_creativity}.

\paragraph{Human Preference of Creative Blends.}

\input{figures/user_study_ttl}

We conducted another user study to assess the quality of blended outputs generated by different methods using input image pairs from Totally Looks Like \cite{rosenfeld2018totally}. \Cref{fig:user_study_ttl} (a) displays the prompt given to participants, who answered two subquestions. First, users identified the main attributes shared by the images (\Cref{fig:user_study_ttl} (b)). Then, users ranked the outputs based on how well they coherently and creatively blended those main attributes (\Cref{fig:user_study_ttl} (c)). This two-step process encourages users to actively consider the relevant attributes for blending before making their selection. We determined the final ranking for each example via majority vote based on first-place votes, followed by second-place votes.

To encourage participants to focus on blending coherence and creativity, instead of other irrelevant characteristics like image quality, we utilized the following enhancement prompt with Gemini 2.5 Flash Image \cite{comanici2025gemini} to standardize the image quality of our blended output to match that of the outputs from models like Gemini and GPT \cite{gpt-image-1}.

\input{prompts/prompt_enhance}

\subsection{How Does Creativity Relate to Diversity?}
\label{sec:diversity}

\input{tables/supp_blending_quantitative}

While Creative Potential and Blend Difficulty describe how a blend could be formed, creativity can also be evaluated in the outputs. We examine one complementary metric, output diversity, echoing Hofstadter’s insight that ``making variations on a theme is the crux of creativity'' \cite{hofstadter2008metamagical}. We quantify output diversity by generating multiple blends per input image pair and computing pairwise perceptual distances using DreamSim \cite{fu2023dreamsim}, denoted by $F$. The diversity $V$ for one example is:
\small
\begin{equation}
V = \frac{1}{\binom{n}{2}} \sum_{i < j} \text{dist}\!\left(F(I_i), F(I_j)\right).
\label{eq:diversity}
\end{equation}
\normalsize
where $I_i$ and $I_j$ are different outputs from the same input, $\text{dist}(\cdot,\cdot)$ denotes DreamSim distance, and $n=3$.
Higher $V$ indicates greater output diversity, and we report the mean $V$ across all examples. We also implement a comparable measure of output diversity using CLIP image similarity \cite{radford2021learning}.

Shown in \Cref{tab:supp_blending_quantitative}, our method generates the most varied blends, in addition to the best creative blends according to humans and the LLM in \Cref{tab:blending_quantitative}. Interestingly, Gemini creates the second-most diverse outputs, even though it lags behind CLIP Avg and GPT in terms of blend quality.

\subsection{Curated Datasets}
\label{sec:curated_datasets}

The Totally Looks Like dataset \cite{rosenfeld2018totally} contains image pairs that humans judge as visually similar. To curate a subset suitable for our evaluation, we focus on image pairs depicting distinct concepts while ensuring high image quality. We first automatically exclude pairs with overly similar content by removing those with both CLIP image similarity \cite{radford2021learning} above 0.65 and DreamSim distance \cite{fu2023dreamsim} below 0.65. A human annotator then filters out any remaining pairs containing text or low-quality images. This process yields 44 high-quality and semantically distinct image pairs.

Additionally, we curate 300 pairs of architectural design images. We begin by collecting images spanning diverse architectural styles and architects from several existing datasets \cite{xu2014architectural, kaggle-architects, kaggle-architectural-styles}. We then form random pairs such that the two images in each pair come from different architectural styles and different architects, and we recruit a human annotator to verify the quality of each image pair.

\input{sec_appendix/failure_cases}

%% file: sec_appendix/baseline_comparison.tex
\begin{table*}[!h]
\centering
\resizebox{\linewidth}{!}{
\begin{tabular}{lccccccccc}
\hline
& \multicolumn{9}{c}{Attribute Masked DreamSim} \\   
\cmidrule(lr){2-10}                                           
& \multicolumn{3}{c}{High Difficulty}
& \multicolumn{3}{c}{Medium Difficulty}
& \multicolumn{3}{c}{Low Difficulty} \\
\cmidrule(lr){2-4}
\cmidrule(lr){5-7}
\cmidrule(lr){8-10}
Method
& Input 1 & Input 2 & Mean
& Input 1 & Input 2 & Mean
& Input 1 & Input 2 & Mean \\ \hline

\multicolumn{10}{l}{\textit{Concept-level blending methods}} \\

VibeSpace (ours)
& 0.632 & 0.540 & \textbf{0.586}
& 0.642 & 0.562 & \textbf{0.602}
& 0.708 & 0.596 & \textbf{0.652} \\

AID \cite{he2024aid}
& 0.483 & 0.531 & \underline{0.507}
& 0.570 & 0.468 & \underline{0.519}
& 0.599 & 0.531 & \underline{0.565} \\

GPT \cite{gpt-image-1}
& 0.412 & 0.336 & 0.374
& 0.314 & 0.290 & 0.302
& 0.444 & 0.355 & 0.399 \\

Gemini \cite{comanici2025gemini}
& 0.410 & 0.388 & 0.399
& 0.300 & 0.285 & 0.292
& 0.518 & 0.454 & 0.486 \\[2pt]

\hline
\multicolumn{10}{l}{\textit{Pixel-level blending methods}} \\

VibeSpace (pixel-level blending)
& 0.552 & 0.621 & \textbf{0.586}
& 0.600 & 0.544 & \underline{0.572}
& 0.660 & 0.591 & \underline{0.626} \\

CLIP Avg (our baseline)
& 0.594 & 0.573 & \underline{0.584}
& 0.584 & 0.581 & \textbf{0.583}
& 0.632 & 0.648 & \textbf{0.640} \\

Yu et al.\ \cite{yu2025probability}
& 0.554 & 0.505 & 0.530
& 0.493 & 0.482 & 0.487
& 0.571 & 0.581 & 0.576 \\

DiffMorpher \cite{zhang2024diffmorpher}
& 0.390 & 0.483 & 0.437
& 0.440 & 0.586 & 0.513
& 0.547 & 0.591 & 0.569 \\ \hline

\end{tabular}}
\caption{
Comparison to current methods on the Totally Looks Like dataset \cite{rosenfeld2018totally}. We report \emph{Attribute-Masked DreamSim} computed in three steps: 
(1) the main shared attribute for each image pair is obtained from our user study; 
(2) an open-vocabulary segmentation model~\cite{lai2024lisa} is used to generate a mask for this attribute; 
(3) DreamSim features are extracted and cosine similarity is computed only over the masked region. 
For each method, the blended midpoint image is compared to both input images: the ``Input 1'' and ``Input 2'' columns report DreamSim similarity between the midpoint and each input, respectively, and the ``Mean'' column reports their average.
Bolded numbers denote the best-performing method among concept-level methods and among pixel-level methods within their respective groups, underlined numbers mark the second best. 
\\\textbf{Insights:} (1) Vibe Space achieves the strongest performance among concept-level blending methods; (2) CLIP Avg---our baseline---is a strong pixel-level method and performs best within its category.}
\label{apptab:baselines}
\end{table*}

\section{Comparisons to Current Methods}
\label{appsec:baseline}

In the main paper, we compared Vibe Space with CLIP Avg (our baseline), GPT \cite{gpt-image-1}, and Gemini \cite{comanici2025gemini}. In this section, we also compare with recent concept-level and pixel-level blending methods, including AID~\cite{he2024aid}, Yu et al.~\cite{yu2025probability}, and DiffMorpher~\cite{zhang2024diffmorpher}. Our experiments demonstrate that CLIP Avg is stronger than these additional baselines.

We use image pairs from the Totally-Looks-Like dataset. Our user studies provide two annotations for each pair: (1) a Blend Difficulty level (low, medium, high) and (2) a short text description of the main shared attribute, or ``vibe''.

\paragraph{Qualitative Comparison.}

\begin{figure*}
    \centering
    \begin{subfigure}{0.48\linewidth}
        \centering
        \includegraphics[width=\linewidth]{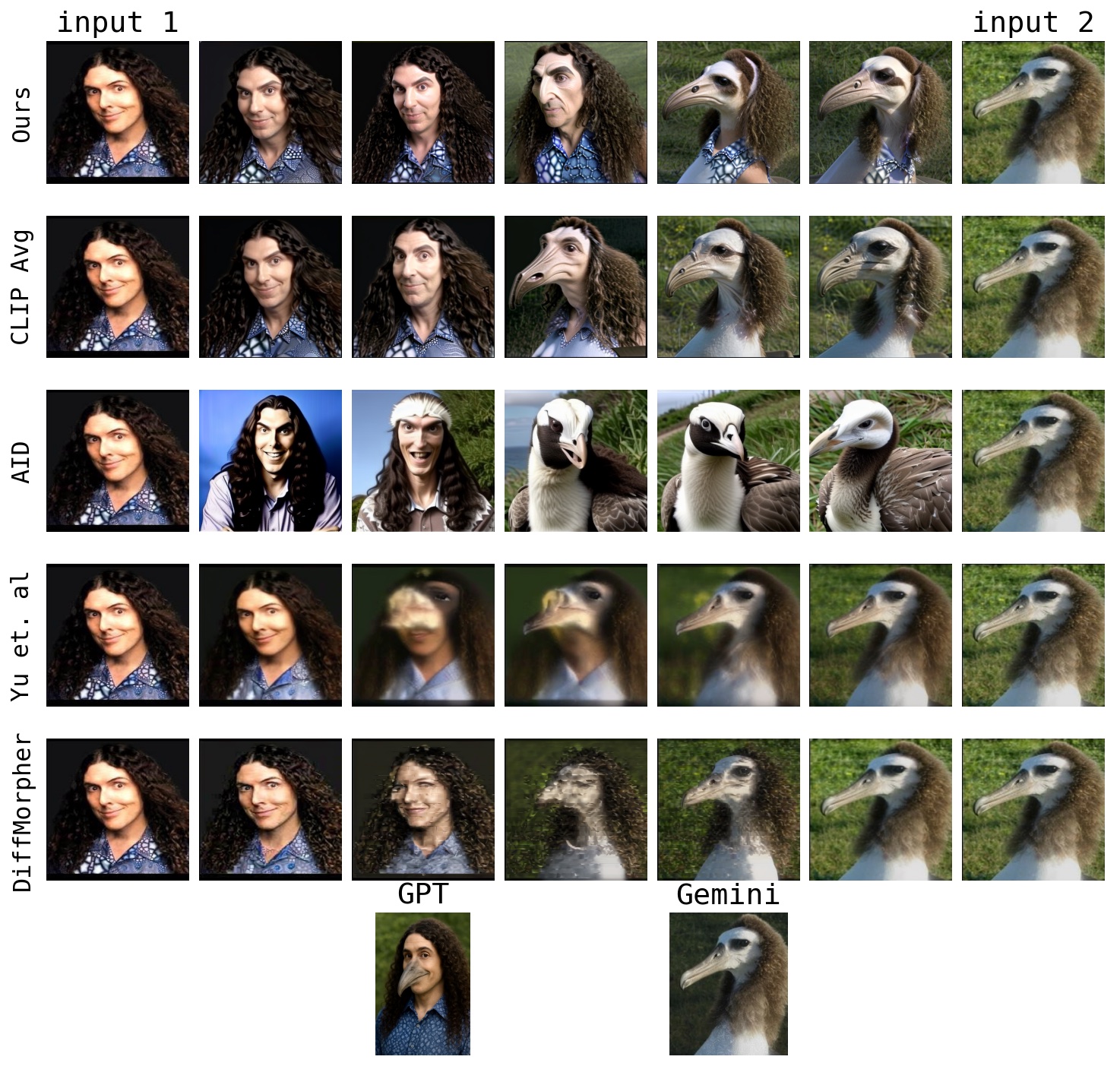}
        \caption{User rated blend difficulty is: High. \\ User annotated main attribute: ``Hair Style''}
    \end{subfigure}
    \hfill
    \begin{subfigure}{0.48\linewidth}
        \centering
        \includegraphics[width=\linewidth]{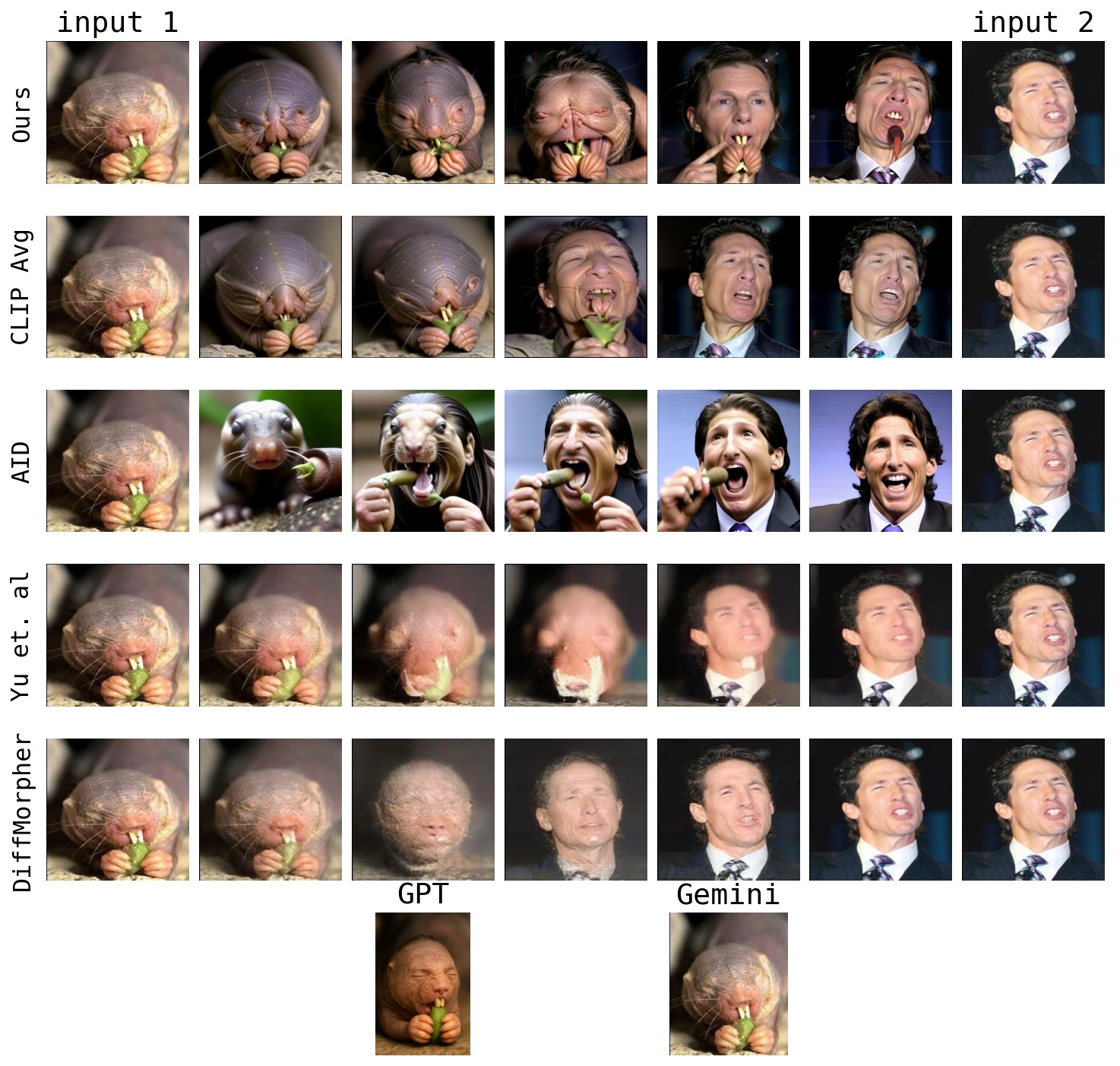}
        \caption{User rated blend difficulty is: High. \\ User annotated main attribute: ``Teeth and Eyes''}
    \end{subfigure}

    \vspace{1em}

    \begin{subfigure}{0.48\linewidth}
        \centering
        \includegraphics[width=\linewidth]{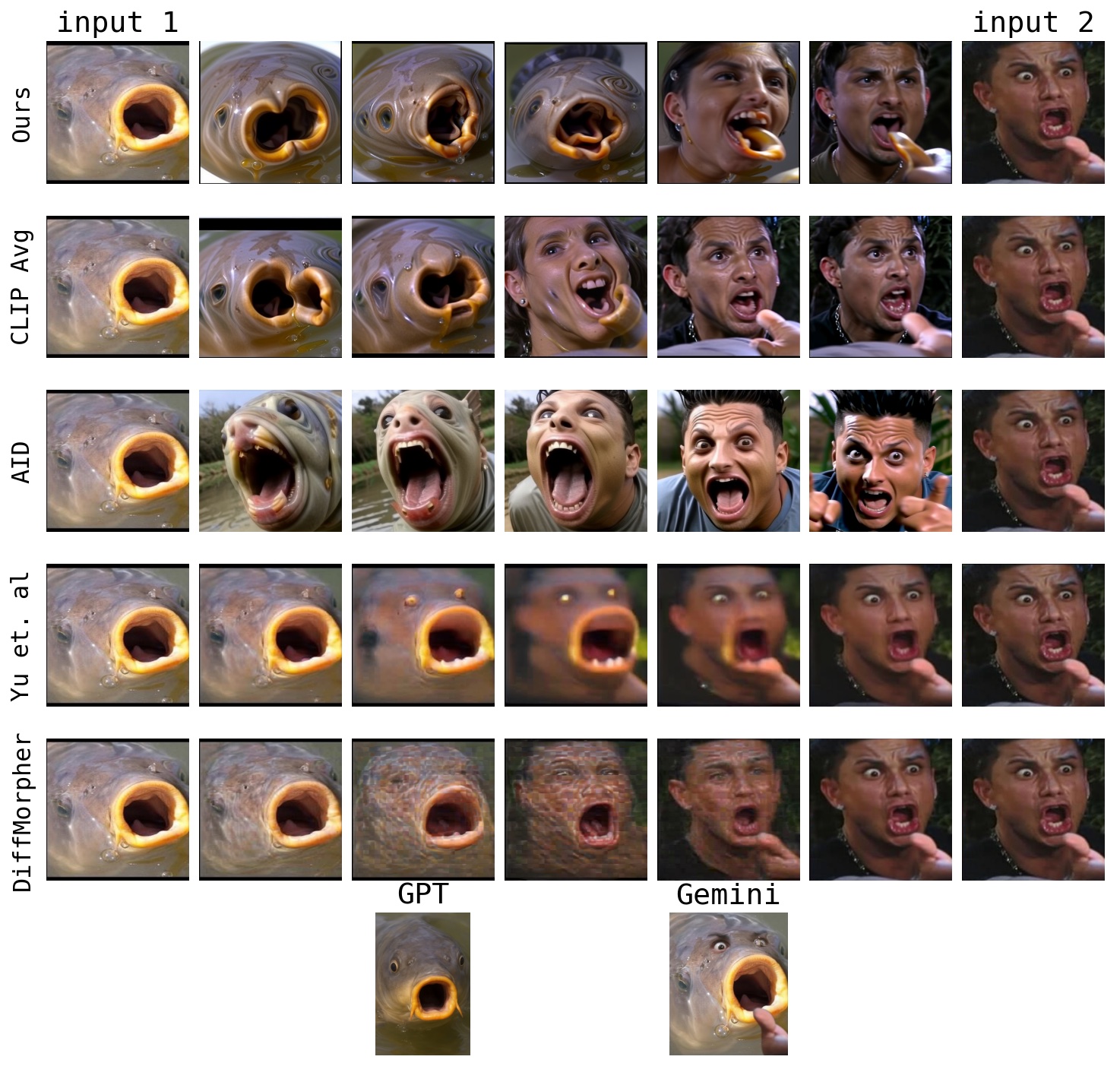}
        \caption{User rated blend difficulty is: High. \\ User annotated main attribute: ``Mouth Shape''}
    \end{subfigure}
    \hfill
    \begin{subfigure}{0.48\linewidth}
        \centering
        \includegraphics[width=\linewidth]{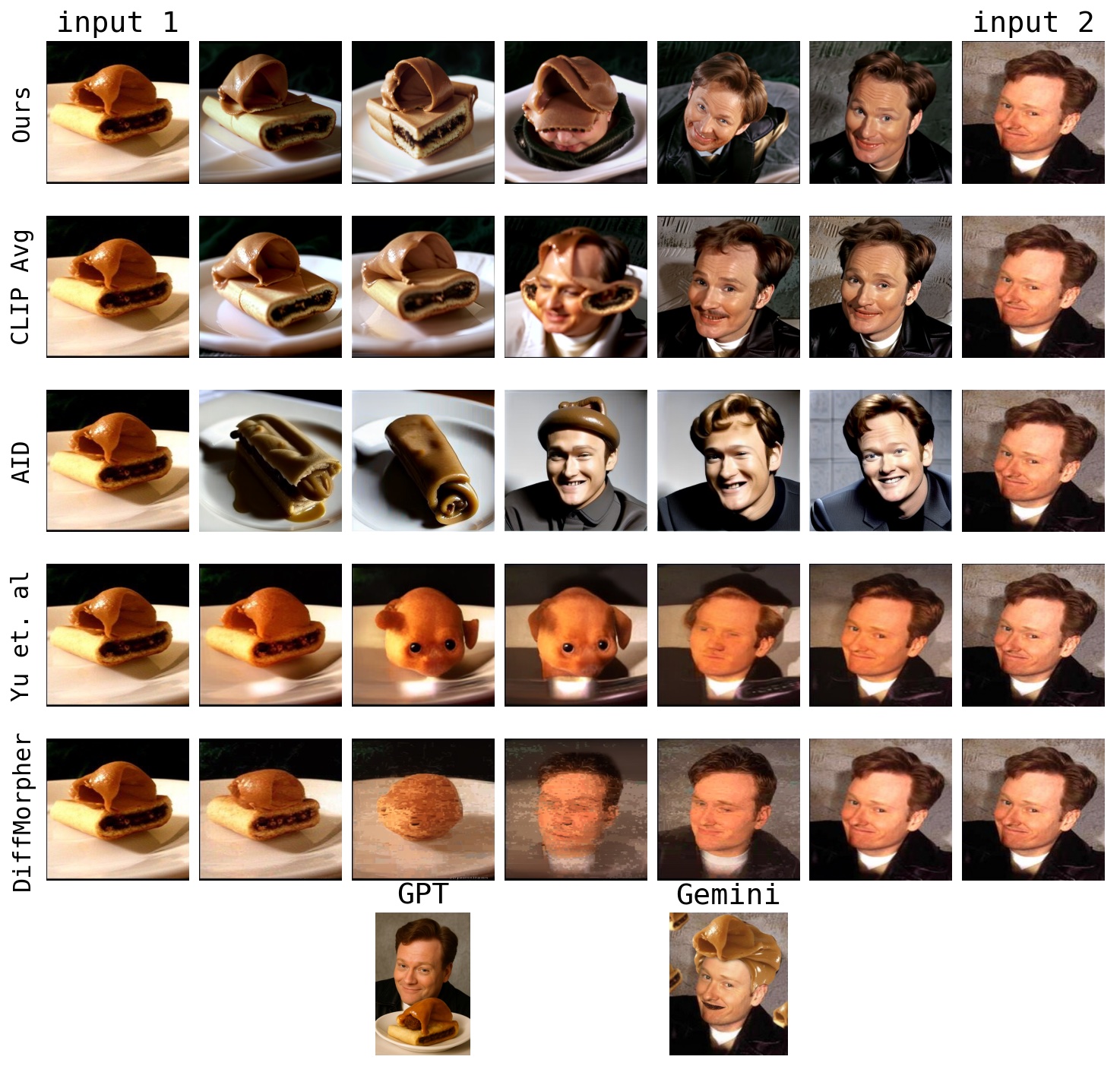}
        \caption{User rated blend difficulty is: High. \\ User annotated main attribute: ``Curly Hair''}
    \end{subfigure}
    \caption{Comparison to current methods. In this set of examples, blend difficulty is ``High''. \textbf{Insight:} Our method captures the main attribute, AID \cite{he2024aid} failed to capture the main attribute and sometimes generate images unrelated to input 1 and input 2, Yu et. al \cite{yu2025probability} and DiffMorpher \cite{zhang2024diffmorpher} produce low-quality images. Although the visual difference between ours vs CLIP Avg is high---CLIP Avg does not capture the main attribute while ours does---the quantitative measure in \Cref{apptab:baselines} shows small difference between Ours and CLIP Avg, this highlights the limitation of quantitative metrics in evaluating Vibe Blending.}
    \label{appfig:compare-high}
\end{figure*}

\begin{figure*}
    \centering
    \begin{subfigure}{0.48\linewidth}
        \centering
        \includegraphics[width=\linewidth]{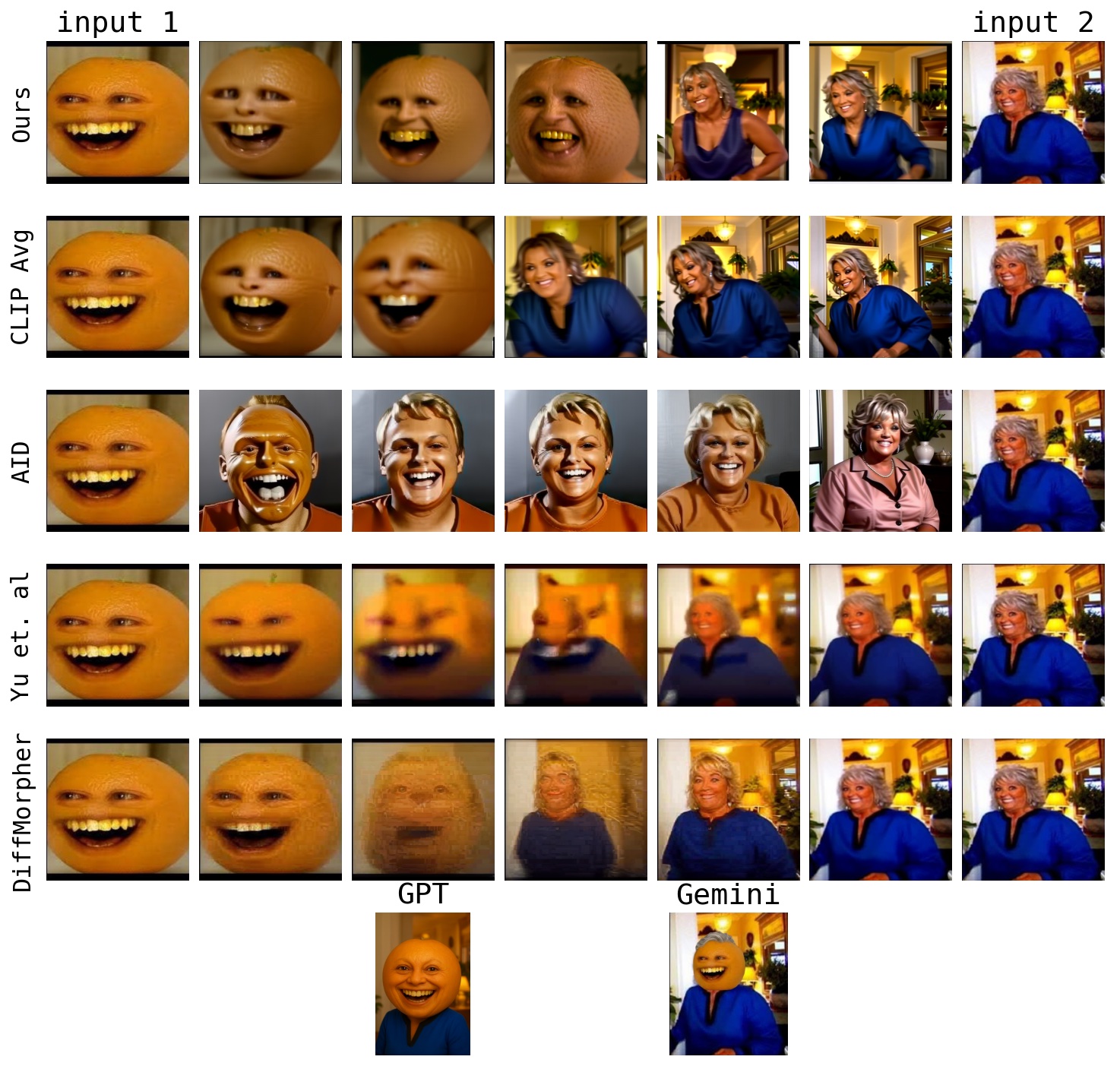}
        \caption{User rated blend difficulty is: Medium. \\ User annotated main attribute: ``Facial Expression''}
    \end{subfigure}
    \hfill
    \begin{subfigure}{0.48\linewidth}
        \centering
        \includegraphics[width=\linewidth]{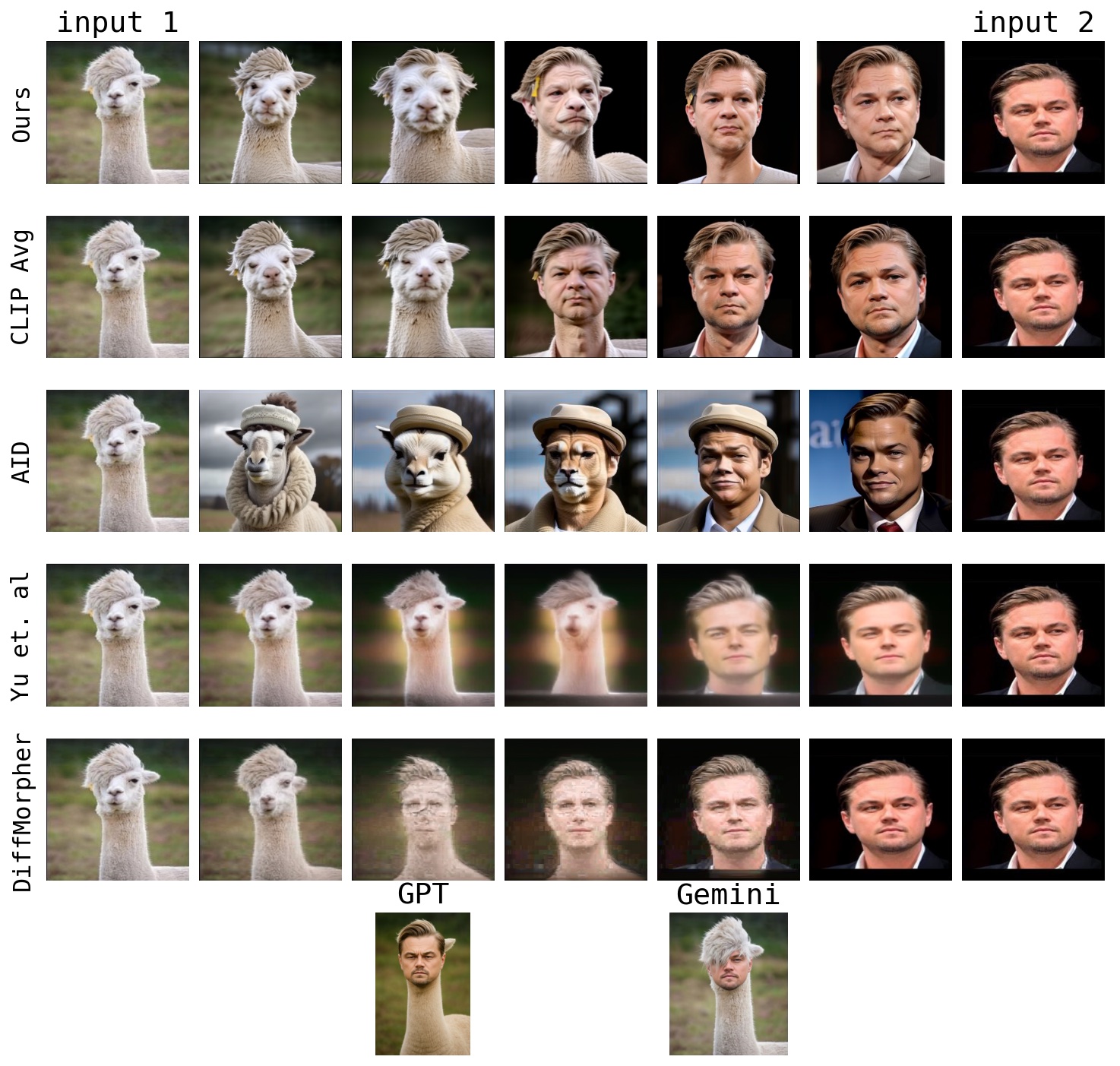}
        \caption{User rated blend difficulty is: Medium. \\ User annotated main attribute: ``Hair Style''}
    \end{subfigure}

    \vspace{1em}

    \begin{subfigure}{0.48\linewidth}
        \centering
        \includegraphics[width=\linewidth]{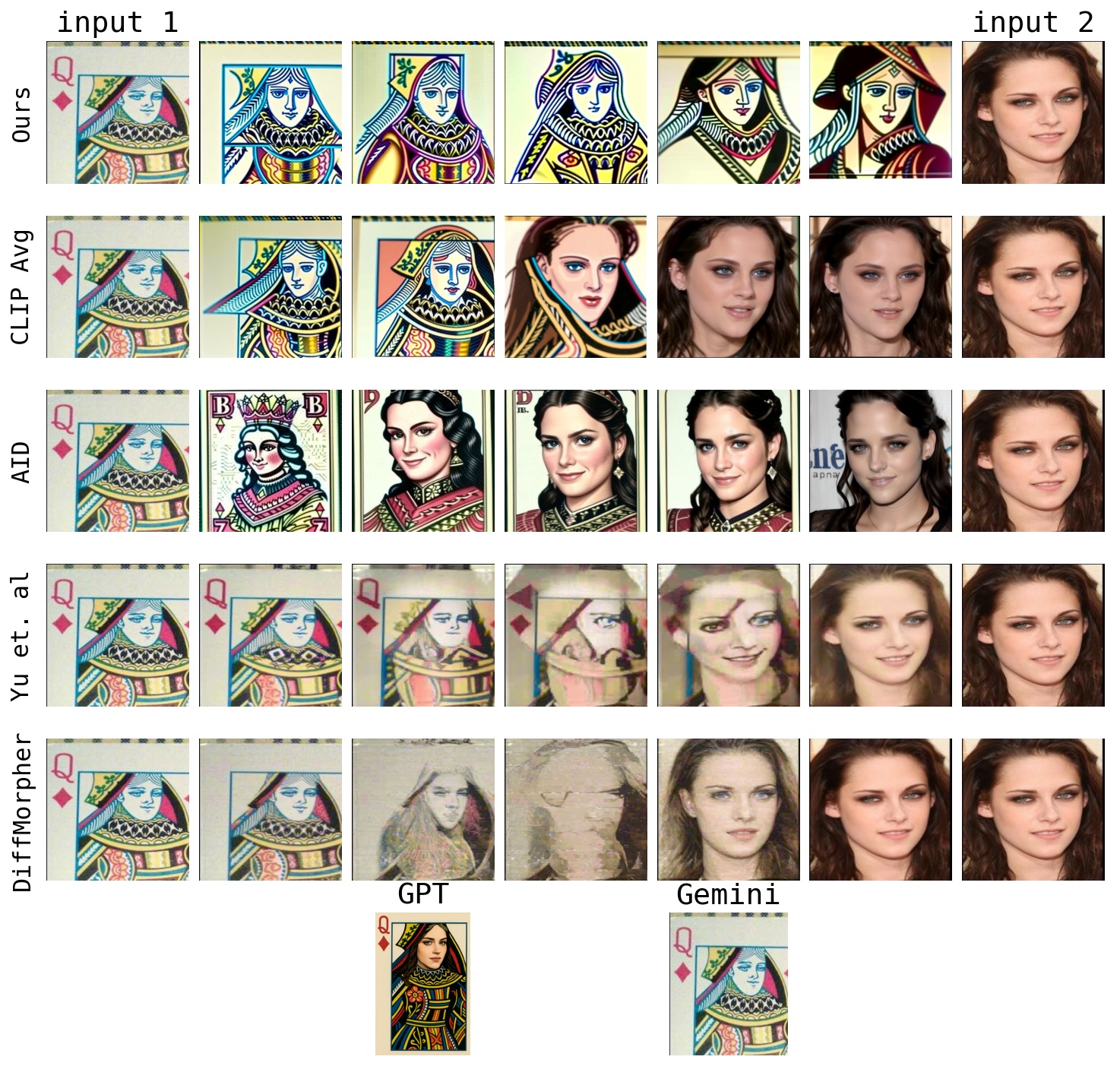}
        \caption{User rated blend difficulty is: Medium. \\ User annotated main attribute: ``Grin''}
    \end{subfigure}
    \hfill
    \begin{subfigure}{0.48\linewidth}
        \centering
        \includegraphics[width=\linewidth]{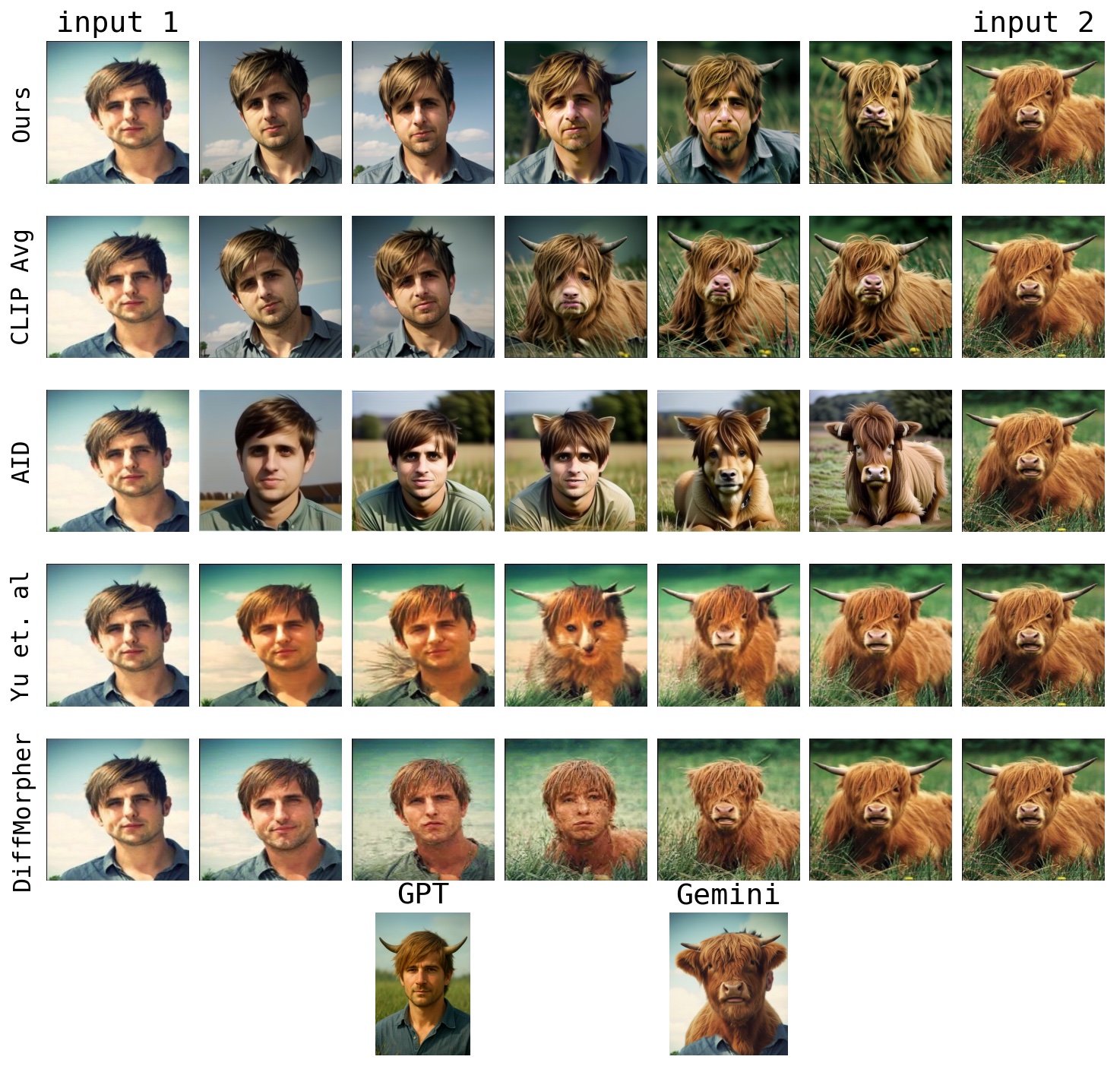}
        \caption{User rated blend difficulty is: Medium. \\ User annotated main attribute: ``Hair Style''}
    \end{subfigure}
    \caption{Comparison to current methods. In this set of examples, blend difficulty is ``Medium''. \textbf{Insight:} Our method captures the main attribute, AID \cite{he2024aid} failed to capture the main attribute and sometimes generate images unrelated to input 1 and input 2, Yu et. al \cite{yu2025probability} and DiffMorpher \cite{zhang2024diffmorpher} produce low-quality images. }
    \label{appfig:compare-medium}
\end{figure*}

\begin{figure*}
    \centering
    \begin{subfigure}{0.48\linewidth}
        \centering
        \includegraphics[width=\linewidth]{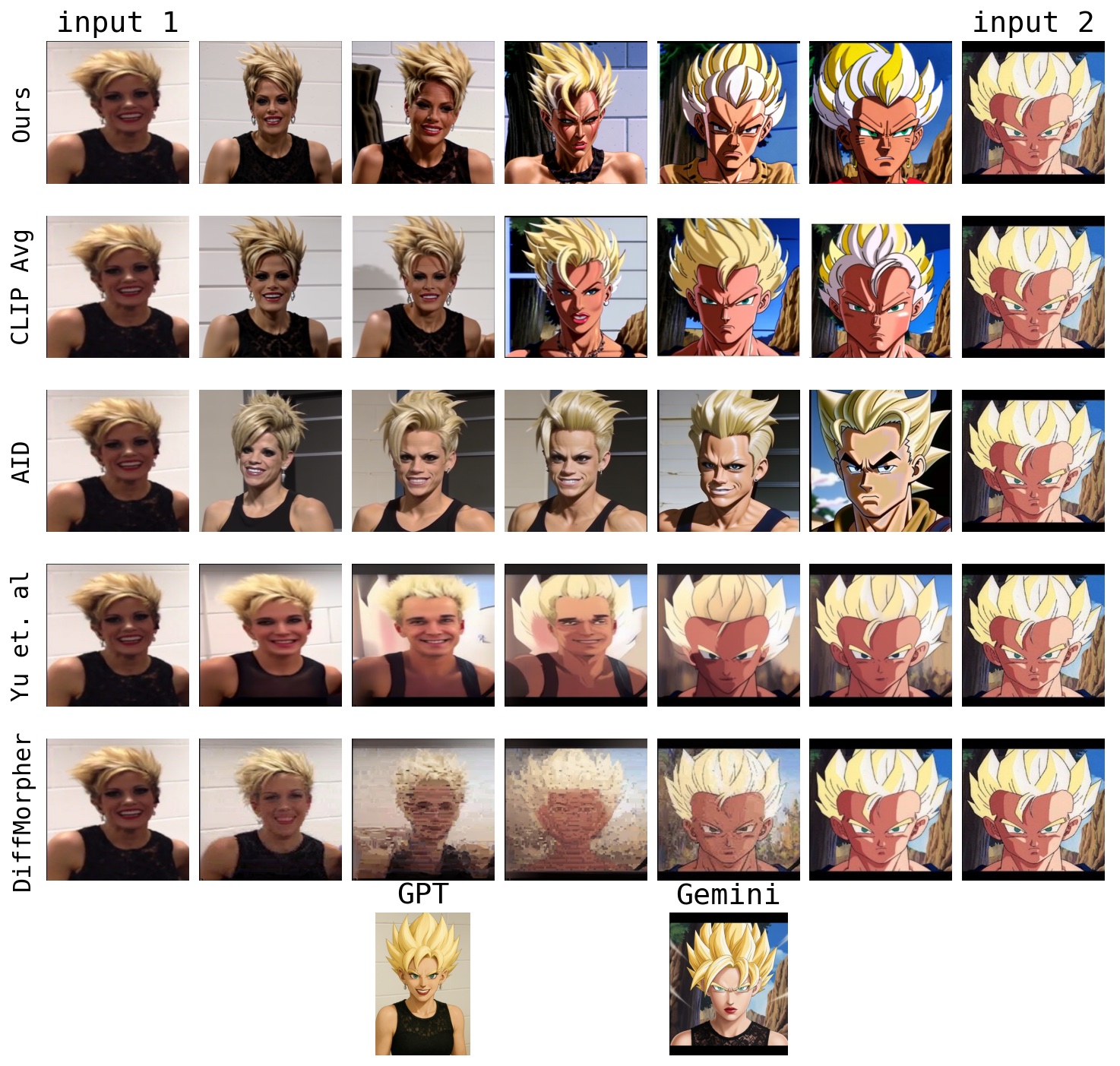}
        \caption{User rated blend difficulty is: Low. \\ User annotated main attribute: ``Spiky Hair''}
    \end{subfigure}
    \hfill
    \begin{subfigure}{0.48\linewidth}
        \centering
        \includegraphics[width=\linewidth]{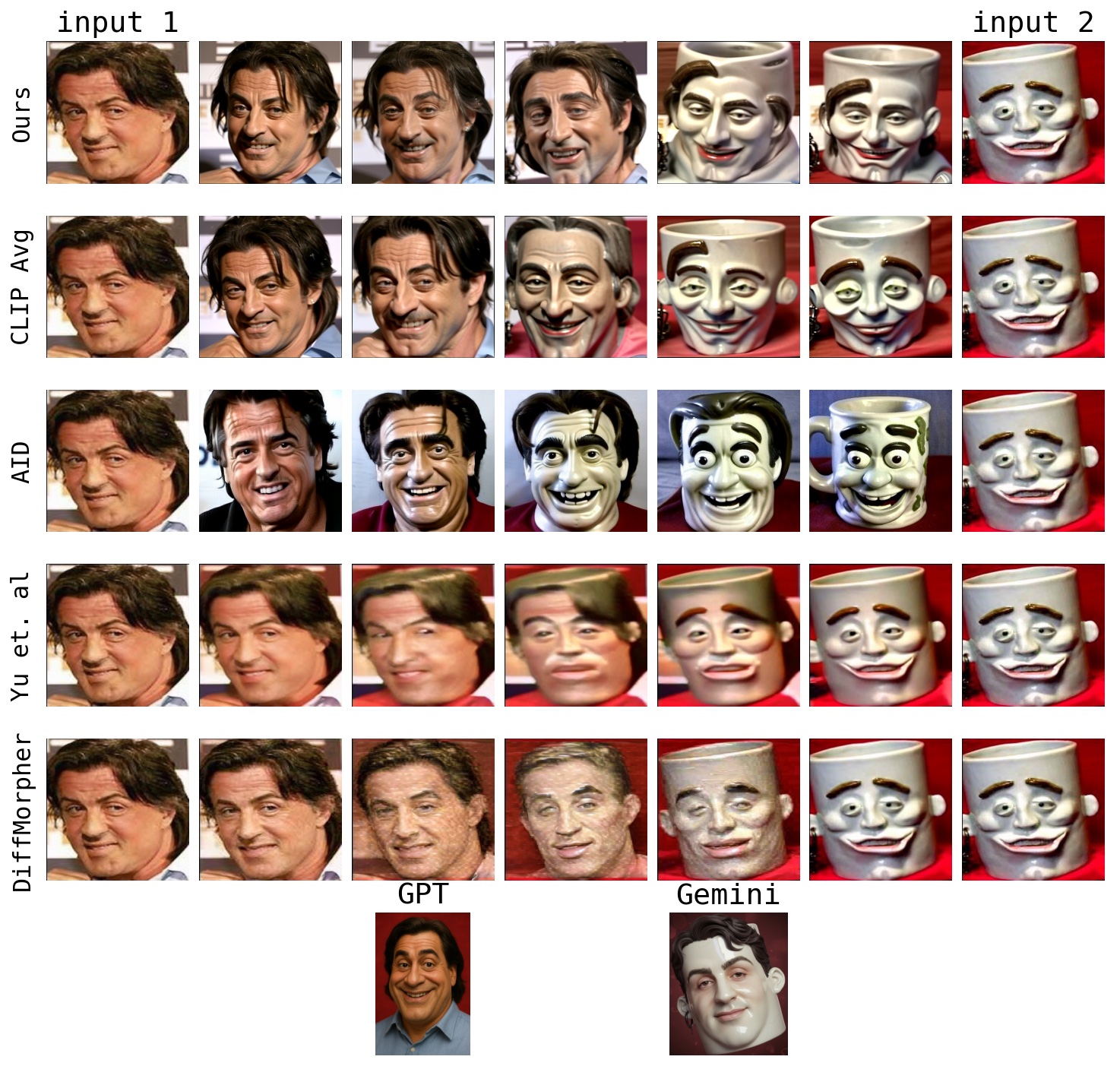}
        \caption{User rated blend difficulty is: Low. \\ User annotated main attribute: ``Grimace''}
    \end{subfigure}

    \vspace{1em}

    \begin{subfigure}{0.48\linewidth}
        \centering
        \includegraphics[width=\linewidth]{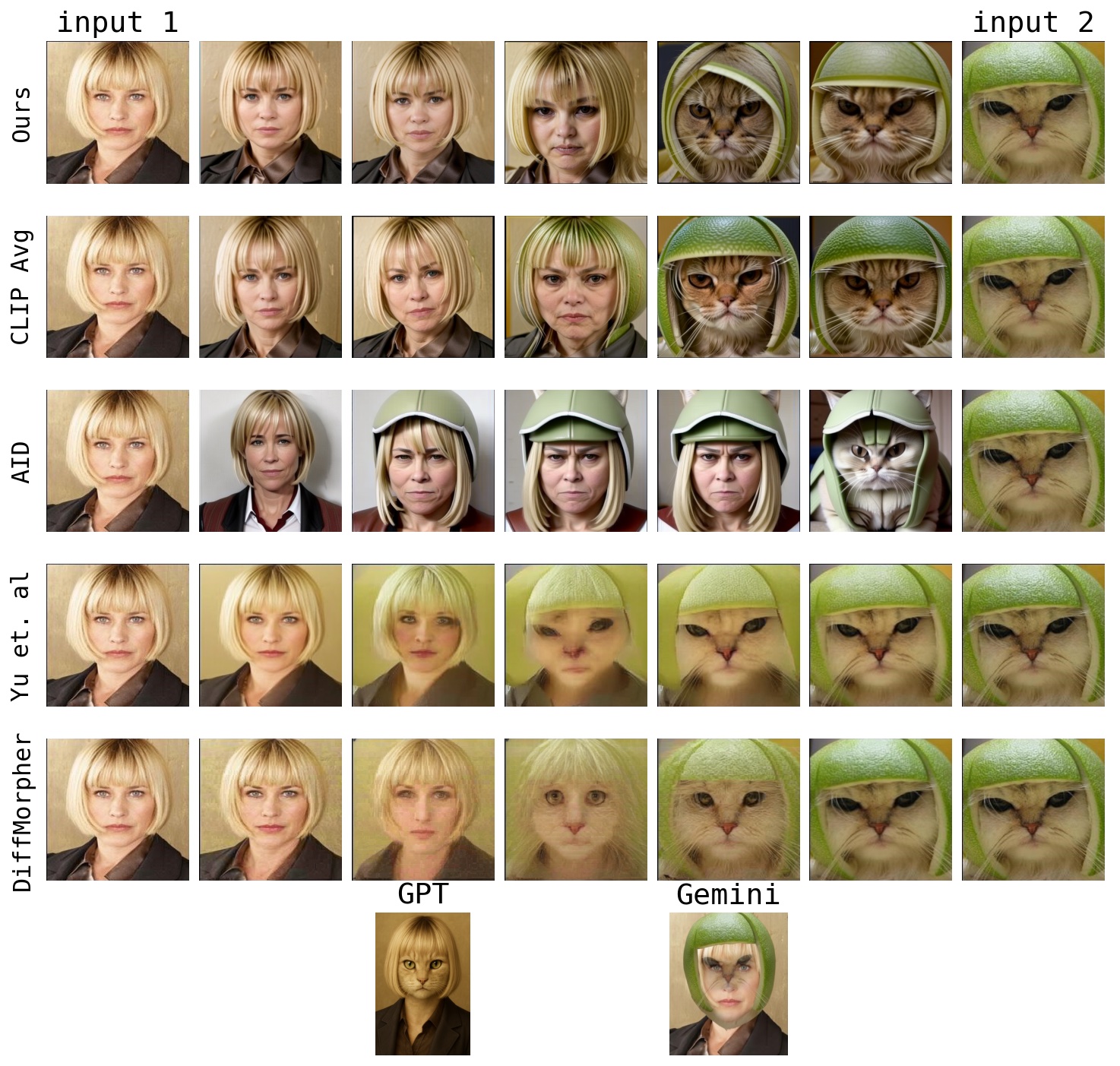}
        \caption{User rated blend difficulty is: Low. \\ User annotated main attribute: ``Hair Style''}
    \end{subfigure}
    \hfill
    \begin{subfigure}{0.48\linewidth}
        \centering
        \includegraphics[width=\linewidth]{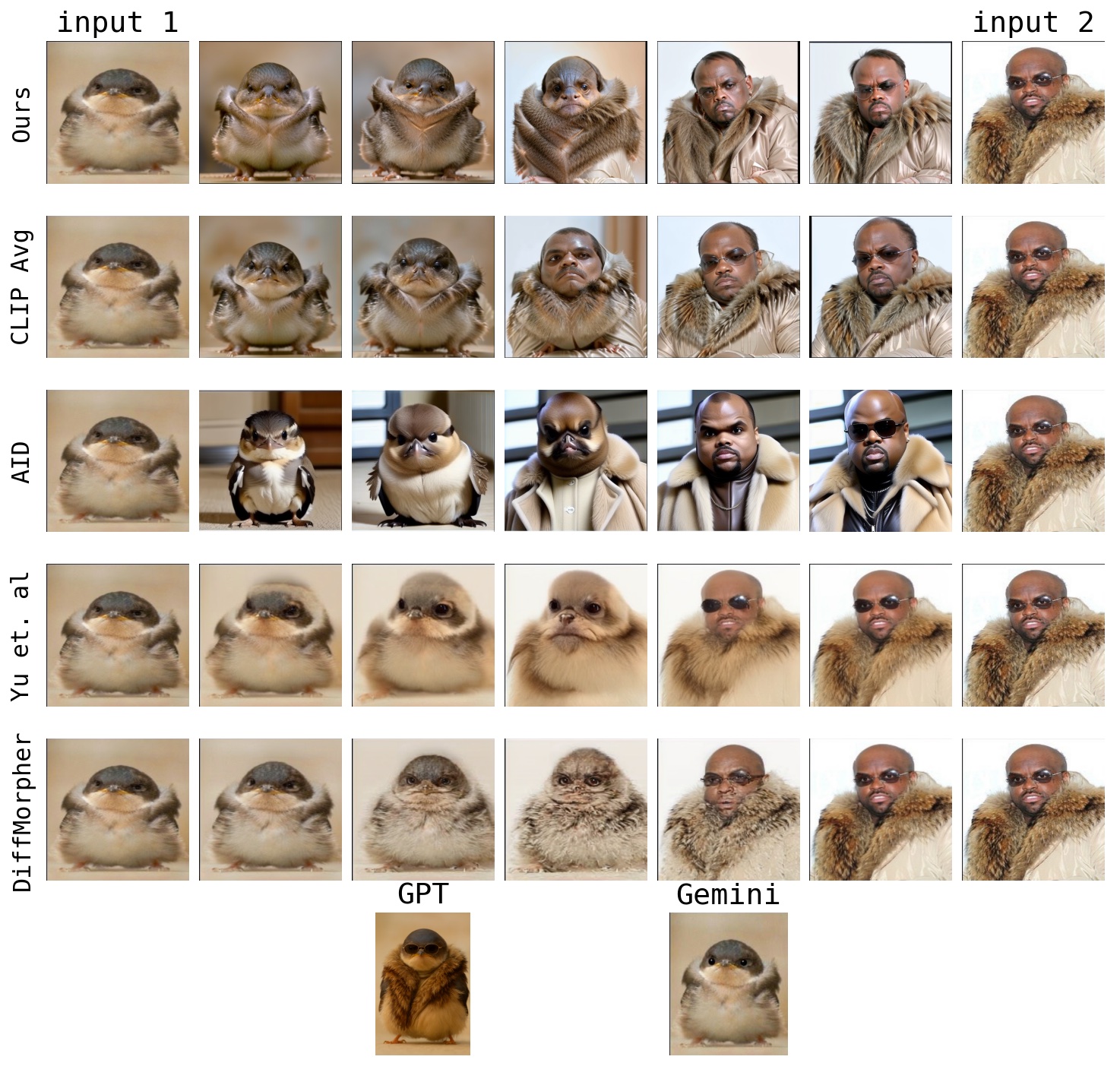}
        \caption{User rated blend difficulty is: Low. \\ User annotated main attribute: ``Clothing''}
    \end{subfigure}
    \caption{Comparison to current methods. In this set of examples, blend difficulty is ``Low''.  \textbf{Insight:} Our method captures the main attribute, AID \cite{he2024aid} failed to capture the main attribute and sometimes generate images unrelated to input 1 and input 2, Yu et. al \cite{yu2025probability} and DiffMorpher \cite{zhang2024diffmorpher} produce low-quality images. The visual difference between Ours vs CLIP Avg is small---both methods can generate coherent blend on the main attributes.}
    \label{appfig:compare-low}
\end{figure*}

As illustrated in \Cref{appfig:compare-low,appfig:compare-medium,appfig:compare-high}, pixel-level blending methods such as Yu et al.~\cite{yu2025probability} and DiffMorpher~\cite{zhang2024diffmorpher} often produce blurred or low-quality intermediate images. CLIP Avg and AID~\cite{he2024aid} generate high-quality images but frequently fail to capture the main shared attribute (e.g., \Cref{appfig:compare-medium}b: ``Hair Style''; \Cref{appfig:compare-high}b: ``Teeth and Eyes''). GPT and Gemini typically perform style transfer or part-level composition rather than attribute-level blending. In contrast, our \textbf{Vibe Space} consistently produces high-quality images while keeping the blending trajectory centered on the attribute identified by users.

\paragraph{Quantitative Comparison.}
We evaluate concept-level blending performance using our \emph{Attribute-Masked DreamSim} metric. For each image pair $(I_A, I_B)$, our user study provides a text description of the main shared attribute between the images, with examples in \Cref{appfig:compare-low}. We use this text to obtain a segmentation mask using an open-vocabulary segmentation model~\cite{lai2024lisa}. All image feature similarity computations are restricted to pixels inside the attribute mask.
Specifically, given an image $I$, let $x_{\mathrm{dreamsim}}(I)$ denote its dense DreamSim \cite{fu2023dreamsim} features (prior to global pooling). For a masked region $\text{Mask}$, we compute the attribute-masked DreamSim embedding via mean pooling:
\begin{equation}
v(I; \text{Mask})
=
\frac{1}{|\text{Mask}|}
\sum_{p \in \text{Mask}} x_{\mathrm{dreamsim}}(I)_p.
\end{equation}

DreamSim similarity between two images is defined as cosine similarity between the attribute-masked embeddings:
\begin{equation}
\mathrm{sim}=
\frac{
\langle v(I_A;\text{Mask}),\, v(I_B;\text{Mask}) \rangle
}{
\|v(I_A;\text{Mask})\|_2 \;
\|v(I_B;\text{Mask})\|_2
}.
\end{equation}

For each method, we measure how well the midpoint blend $I_{\mathrm{mid}}$ preserves the target attribute region by computing:
(1) $\mathrm{sim}(I_{\mathrm{mid}}, I_A)$,
(2) $\mathrm{sim}(I_{\mathrm{mid}}, I_B)$,
and (3) the average of these two scores.

As shown in \Cref{apptab:baselines}, \textbf{Vibe Space} achieves the strongest performance among concept-level blending methods, while our CLIP Avg baseline remains the best-performing pixel-level blending method. 

\begin{figure}[h]
    \centering
    \vspace{-10pt}
    \includegraphics[width=1\linewidth]{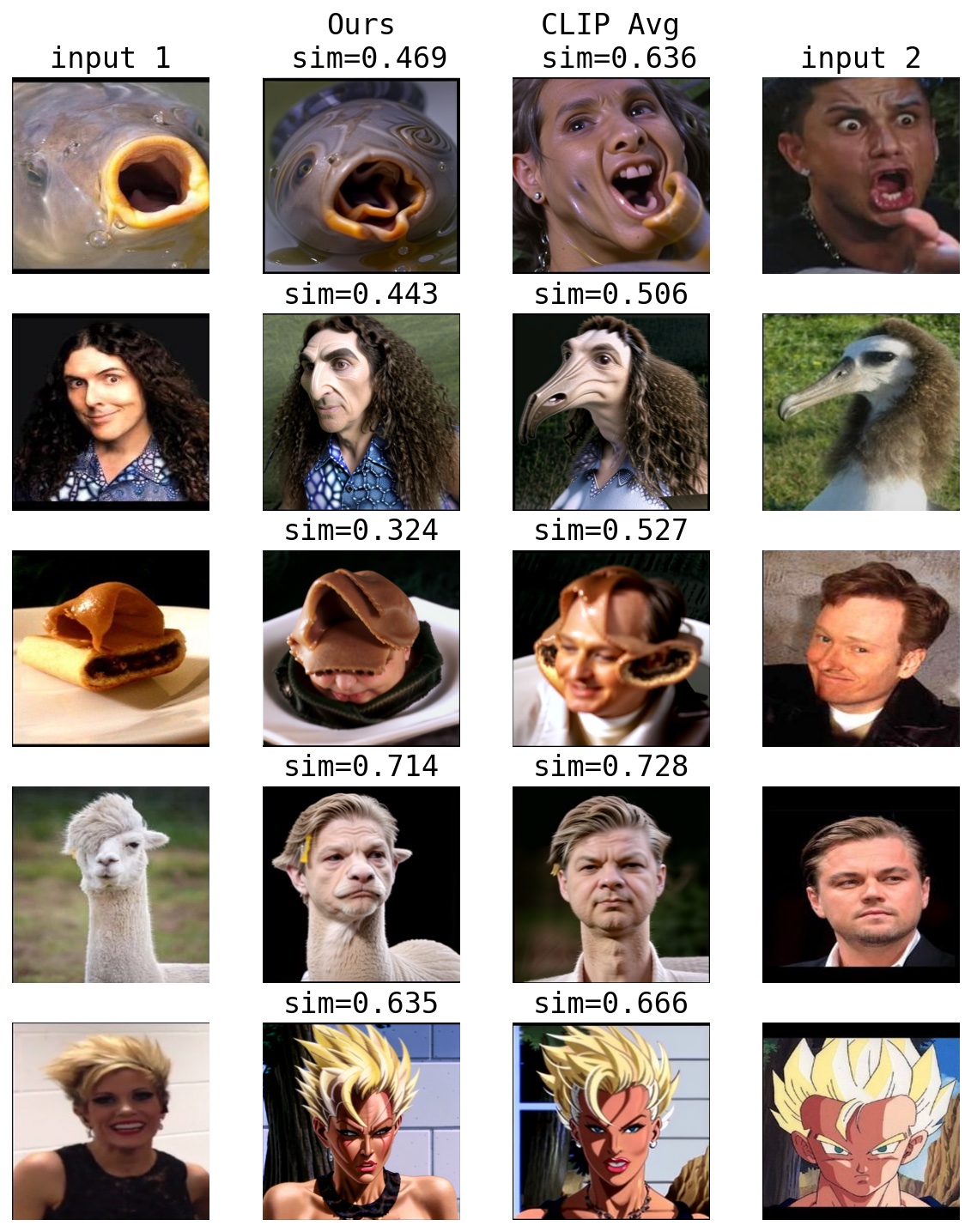}
    \vspace{-10pt}
    \caption{Limitation of quantitative evaluation. While CLIP Avg achieved a higher score on the first three pairs, it fails to capture the main attributes (mouth in the first row, hairstyle in the second row, hairstyle in the third row) and instead blends all possible attributes. Our Vibe Space, on the other hand, successfully blends the main attributes without blending distracting attributes.}
    \label{fig:metric-limitation}
\end{figure}

The \emph{Attribute-Masked DreamSim} metric does not fully capture the differences between the methods. While the scores for Vibe Space and the CLIP Avg baseline are numerically similar in \Cref{apptab:baselines}, \Cref{fig:metric-limitation} shows qualitative discrepancies. Specifically, CLIP Avg fails to accurately capture the primary attributes, such as the mouth in the first row, the hairstyle in the second row, and the hairstyle in the third row. Instead, CLIP Avg often blends all possible attributes simultaneously, whereas our Vibe Space approach effectively prioritizes and blends the main attributes before considering sub-attributes.

\clearpage

\clearpage

\clearpage

\begin{figure*}
    \centering
    \begin{subfigure}{0.48\linewidth}
        \centering
        \includegraphics[width=\linewidth]{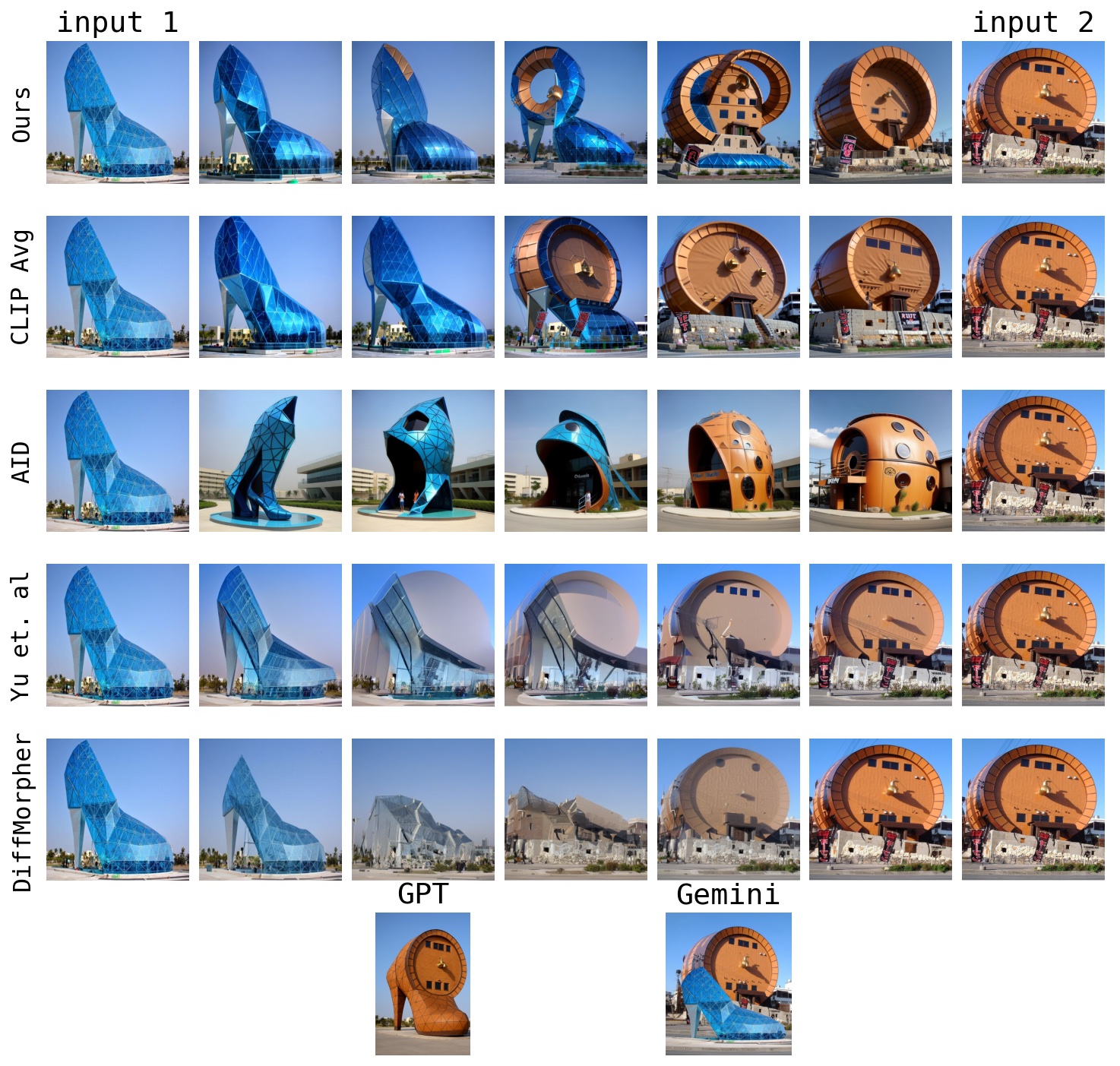}
        \caption{Architecture dataset.}
    \end{subfigure}
    \hfill
    \begin{subfigure}{0.48\linewidth}
        \centering
        \includegraphics[width=\linewidth]{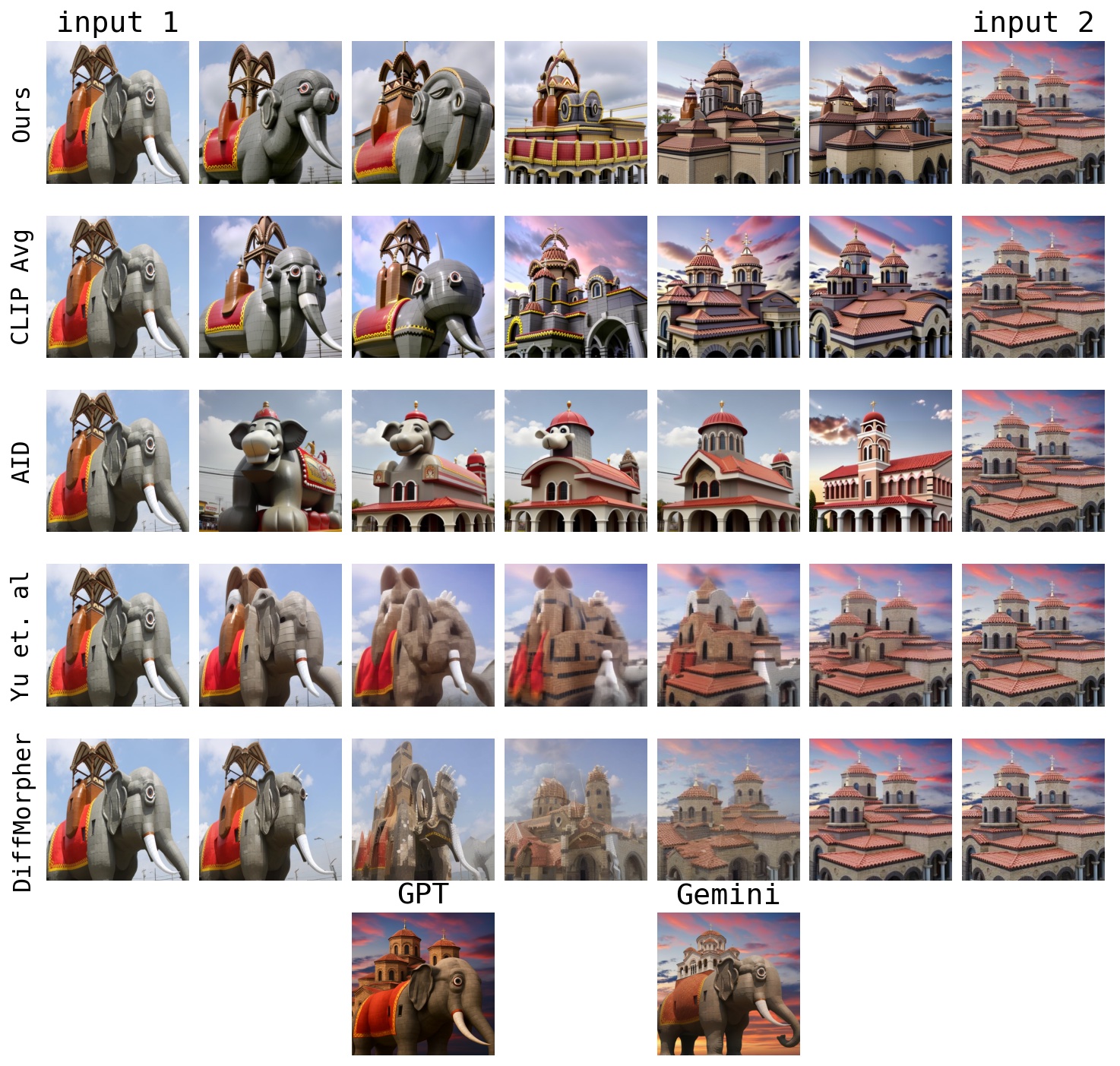}
        \caption{Architecture dataset.}
    \end{subfigure}

    \vspace{1em}

    \begin{subfigure}{0.48\linewidth}
        \centering
        \includegraphics[width=\linewidth]{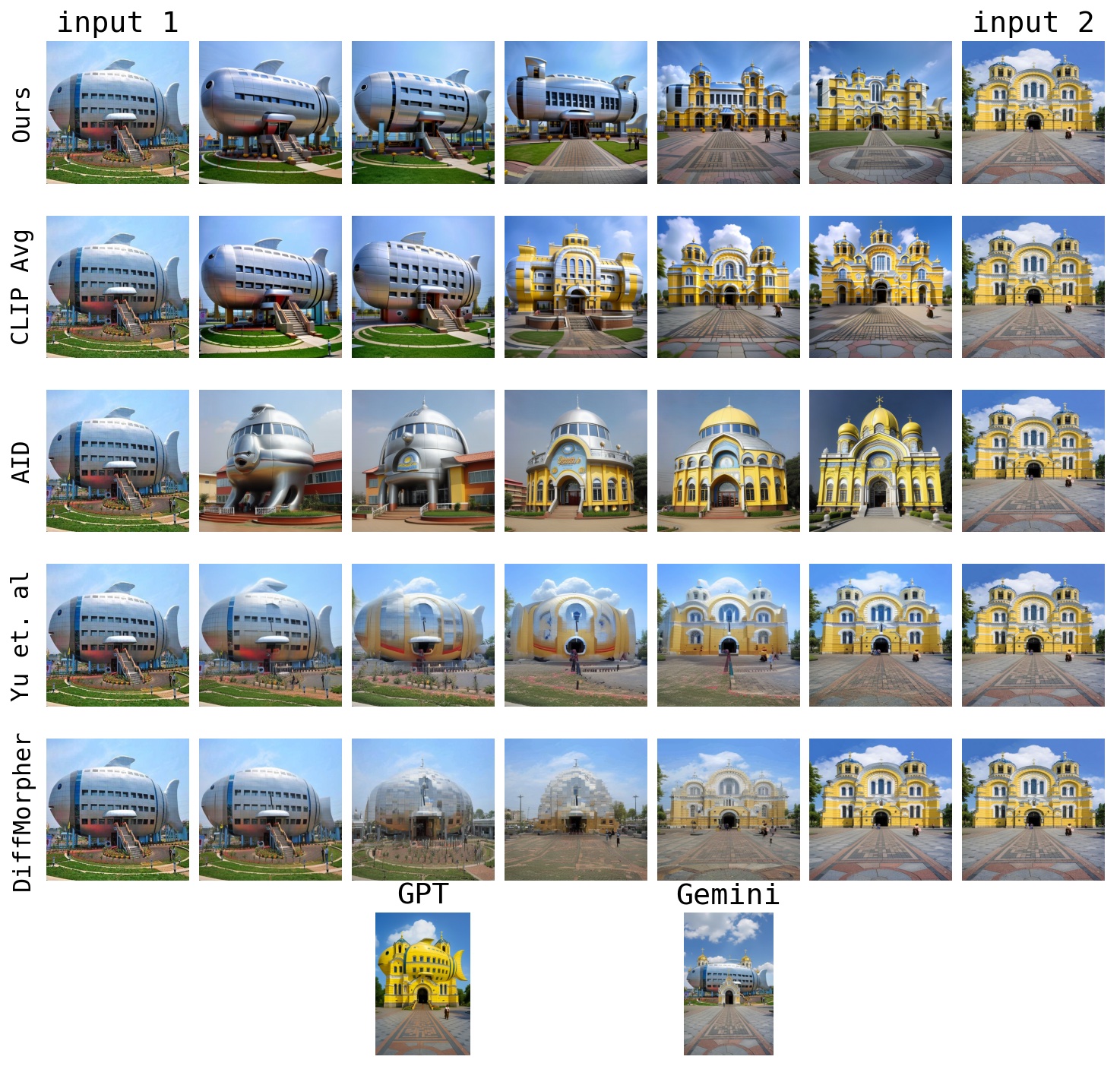}
        \caption{Architecture dataset.}
    \end{subfigure}
    \hfill
    \begin{subfigure}{0.48\linewidth}
        \centering
        \includegraphics[width=\linewidth]{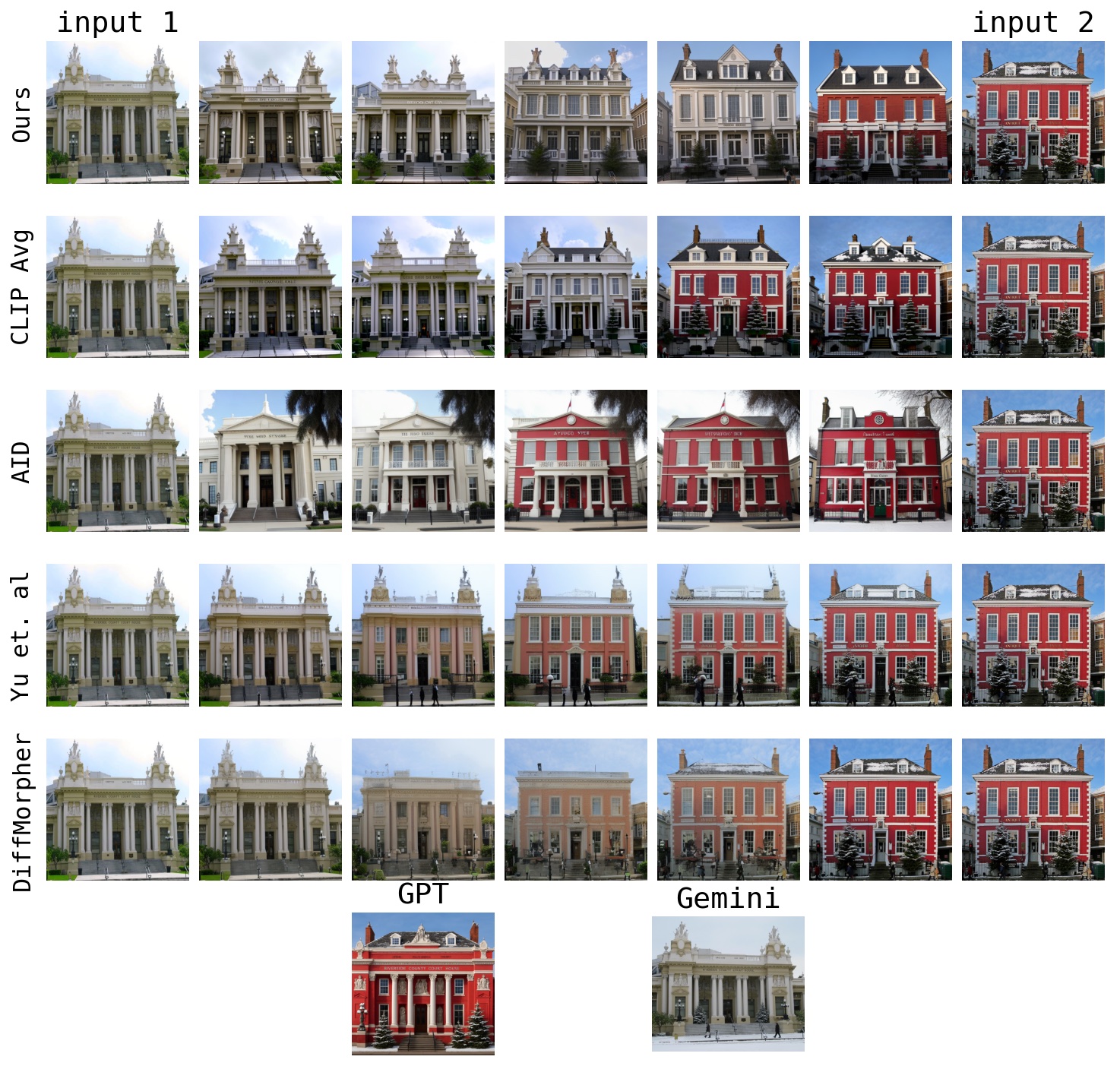}
        \caption{Architecture dataset.}
    \end{subfigure}
    \caption{Comparison to current methods. \textbf{Insight:} Our method captures the main attribute, AID \cite{he2024aid} failed to capture the main attribute and sometimes generate images unrelated to input 1 and input 2, Yu et. al \cite{yu2025probability} and DiffMorpher \cite{zhang2024diffmorpher} produce low-quality images.}
    \label{appfig:compare-architecture}
\end{figure*}

\clearpage

%% file: sec_appendix/ablations.tex
\begin{table*}[t]
\centering
\resizebox{\linewidth}{!}{
\begin{tabular}{lccccccccc}
\hline
& \multicolumn{9}{c}{Attribute Masked DreamSim} \\
\cmidrule(lr){2-10}
& \multicolumn{3}{c}{High Difficulty}
& \multicolumn{3}{c}{Medium Difficulty}
& \multicolumn{3}{c}{Low Difficulty} \\
\cmidrule(lr){2-4}
\cmidrule(lr){5-7}
\cmidrule(lr){8-10}
Method
& Input 1 & Input 2 & Mean
& Input 1 & Input 2 & Mean
& Input 1 & Input 2 & Mean \\ \hline

\begin{tabular}{l}
Multi-scale (coarse-to-fine) \\
\quad $\mathcal{M} = \{4, 8, \dots, 64\}$
\end{tabular}
& 0.626 & 0.420 & 0.523
& 0.669 & 0.490 & \textbf{0.580}
& 0.668 & 0.618 & 0.643 \\

\begin{tabular}{l}
Single-scale (fine) \\
\quad $\mathcal{M} = \{64\}$
\end{tabular}
& 0.650 & 0.459 & 0.554
& 0.621 & 0.490 & 0.555
& 0.731 & 0.573 & \textbf{0.652} \\

\begin{tabular}{l}
Single-scale (coarse) \\
\quad $\mathcal{M} = \{4\}$
\end{tabular}
& 0.625 & 0.525 & \textbf{0.575}
& 0.651 & 0.496 & 0.573
& 0.749 & 0.545 & 0.647 \\

\begin{tabular}{l}
No flag loss \\
\quad $\mathcal{M} = \varnothing$
\end{tabular}
& 0.450 & 0.579 & 0.514
& 0.568 & 0.545 & 0.556
& 0.658 & 0.546 & 0.602 \\

\hline

\end{tabular}}
\caption{
Ablation of our methods on the Totally-Looks-Like dataset. We report \emph{Attribute-Masked DreamSim} computed in three steps: 
(1) the main shared attribute for each image pair is obtained from our user study; 
(2) an open-vocabulary segmentation model~\cite{lai2024lisa} is used to generate a mask for this attribute; 
(3) DreamSim features are extracted and cosine similarity is computed only over the masked region. 
For each method, the blended midpoint image is compared to both input images: the ``Input 1'' and ``Input 2'' columns report DreamSim similarity between the midpoint and each input, respectively, and the ``Mean'' column reports their average.
Bold numbers denote the best-performing method. 
\\\textbf{Insights:} (1) With flag loss works better than no flag loss across all low-to-high difficulty cases. (2) Single-scale fine-grained-only works best on low difficulty cases; coarse-only works best on high difficulty cases. Multi-scale works well on all cases, and multi-scale is more robust than single-scale.
}
\label{apptab:ablation-scale}
\end{table*}

\begin{table*}[t]
\centering
\resizebox{\linewidth}{!}{
\begin{tabular}{lccccccccc}
\hline
& \multicolumn{9}{c}{Attribute Masked DreamSim} \\
\cmidrule(lr){2-10}
& \multicolumn{3}{c}{High Difficulty}
& \multicolumn{3}{c}{Medium Difficulty}
& \multicolumn{3}{c}{Low Difficulty} \\
\cmidrule(lr){2-4}
\cmidrule(lr){5-7}
\cmidrule(lr){8-10}
Method
& Input 1 & Input 2 & Mean
& Input 1 & Input 2 & Mean
& Input 1 & Input 2 & Mean \\ \hline

\begin{tabular}{l}
w/ correspondence \\
\quad $n_{\text{clusters}} = 10$
\end{tabular}
& 0.632 & 0.540 & \textbf{0.586}
& 0.592 & 0.572 & 0.582
& 0.714 & 0.557 & 0.636 \\

\begin{tabular}{l}
w/ correspondence \\
\quad $n_{\text{clusters}} = 20$
\end{tabular}
& 0.681 & 0.451 & 0.566
& 0.617 & 0.554 & 0.585
& 0.724 & 0.555 & 0.639 \\

\begin{tabular}{l}
w/ correspondence \\
\quad $n_{\text{clusters}} = 30$
\end{tabular}
& 0.658 & 0.466 & 0.562
& 0.642 & 0.562 & \textbf{0.602}
& 0.710 & 0.552 & 0.631 \\

\begin{tabular}{l}
w/ correspondence \\
\quad $n_{\text{clusters}} = 40$
\end{tabular}
& 0.581 & 0.545 & 0.563
& 0.625 & 0.517 & 0.571
& 0.708 & 0.596 & \textbf{0.652} \\

\begin{tabular}{l}
w/ correspondence \\
\quad $n_{\text{clusters}} = 50$
\end{tabular}
& 0.630 & 0.459 & 0.544
& 0.640 & 0.532 & 0.586
& 0.720 & 0.540 & 0.630 \\

\begin{tabular}{l}
w/o correspondence \\
\quad (pixel-level blending)
\end{tabular}
& 0.552 & 0.621 & \textbf{0.586}
& 0.600 & 0.544 & 0.572
& 0.660 & 0.591 & 0.626 \\ \hline

\end{tabular}}
\caption{
Ablation of our methods on the Totally-Looks-Like dataset. We report \emph{Attribute-Masked DreamSim} computed in three steps: 
(1) the main shared attribute for each image pair is obtained from our user study; 
(2) an open-vocabulary segmentation model~\cite{lai2024lisa} is used to generate a mask for this attribute; 
(3) DreamSim features are extracted and cosine similarity is computed only over the masked region. 
For each method, the blended midpoint image is compared to both input images: the ``Input 1'' and ``Input 2'' columns report DreamSim similarity between the midpoint and each input, respectively, and the ``Mean'' column reports their average.
Bold numbers denote the best-performing method. 
\\\textbf{Insights:} (1) When increasing blend difficulty from low to high, computing correspondence with less clusters works better, this is because correspondence is harder to compute for large number of clusters on high difficulty cases (see \Cref{sec:appendix_correspondence}) (2) Although w/o correspondence works great on high difficulty examples, it doesn't work well on medium and low difficulty cases.
}
\label{apptab:ablation-correspondence}
\end{table*}

\begin{table*}[t]
\centering
\resizebox{\linewidth}{!}{
\begin{tabular}{lccccccccc}
\hline
& \multicolumn{9}{c}{Attribute Masked DreamSim} \\
\cmidrule(lr){2-10}
& \multicolumn{3}{c}{High Difficulty}
& \multicolumn{3}{c}{Medium Difficulty}
& \multicolumn{3}{c}{Low Difficulty} \\
\cmidrule(lr){2-4}
\cmidrule(lr){5-7}
\cmidrule(lr){8-10}
Method
& Input 1 & Input 2 & Mean
& Input 1 & Input 2 & Mean
& Input 1 & Input 2 & Mean \\ \hline

\begin{tabular}{l}
$f:\text{DINO} \to \text{Vibe}$, \\
$g:\text{Vibe} \to \text{CLIP}$
\end{tabular}
& 0.626 & 0.420 & 0.523
& 0.669 & 0.490 & \textbf{0.580}
& 0.668 & 0.618 & \textbf{0.643} \\

\begin{tabular}{l}
$f:\text{CLIP} \to \text{Vibe}$, \\
$g:\text{Vibe} \to \text{CLIP}$
\end{tabular}
& 0.648 & 0.444 & \textbf{0.546}
& 0.731 & 0.422 & 0.577
& 0.665 & 0.492 & 0.579 \\ \hline

\end{tabular}}
\caption{
Ablation of our methods on the Totally-Looks-Like dataset. We report \emph{Attribute-Masked DreamSim} computed in three steps: 
(1) the main shared attribute for each image pair is obtained from our user study; 
(2) an open-vocabulary segmentation model~\cite{lai2024lisa} is used to generate a mask for this attribute; 
(3) DreamSim features are extracted and cosine similarity is computed only over the masked region. 
For each method, the blended midpoint image is compared to both input images: the ``Input 1'' and ``Input 2'' columns report DreamSim similarity between the midpoint and each input, respectively, and the ``Mean'' column reports their average.
Bold numbers denote the best-performing method. 
\\\textbf{Insights:} Mixing DINO and CLIP features works best on medium difficulty and low difficulty cases, however on high difficulty cases, using CLIP without mixing DINO works better.
}
\label{apptab:ablation-dinoclip}
\end{table*}

\section{Ablation Studies}
\label{appsec:ablation}

We conduct a series of ablations to understand the contribution of each component in Vibe Space. Experiments are performed on the Totally-Looks-Like dataset and evaluated using Attribute-Masked DreamSim (see \Cref{appsec:baseline}). Across all studies, we report performance separately for high-, medium-, and low-difficulty image pairs.

\paragraph{Flag Loss.}
The flag loss aligns the learned Vibe Space with the multiscale geometry of the underlying diffusion map by matching the Gram matrix of Vibe features to the flag-space kernel (see Equation (4)). This loss depends on a set of scales $\mathcal{M}$, where each $m \in \mathcal{M}$ corresponds to a prefix of low-frequency Laplacian eigenvectors.

We ablate the following setting:
\begin{itemize}
    \item \emph{multi-scale coarse-to-fine} ($\mathcal{M} = \{4, 8, \dots, 64\}$)
    \item \emph{single-scale fine-only} ($\mathcal{M}=\{64\}$)
    \item \emph{single-scale coarse-only} ($\mathcal{M}=\{4\}$)
    \item \emph{no flag loss} ($\mathcal{M}=\varnothing$)
\end{itemize}

\begin{figure}
    \centering
    \includegraphics[width=1\linewidth]{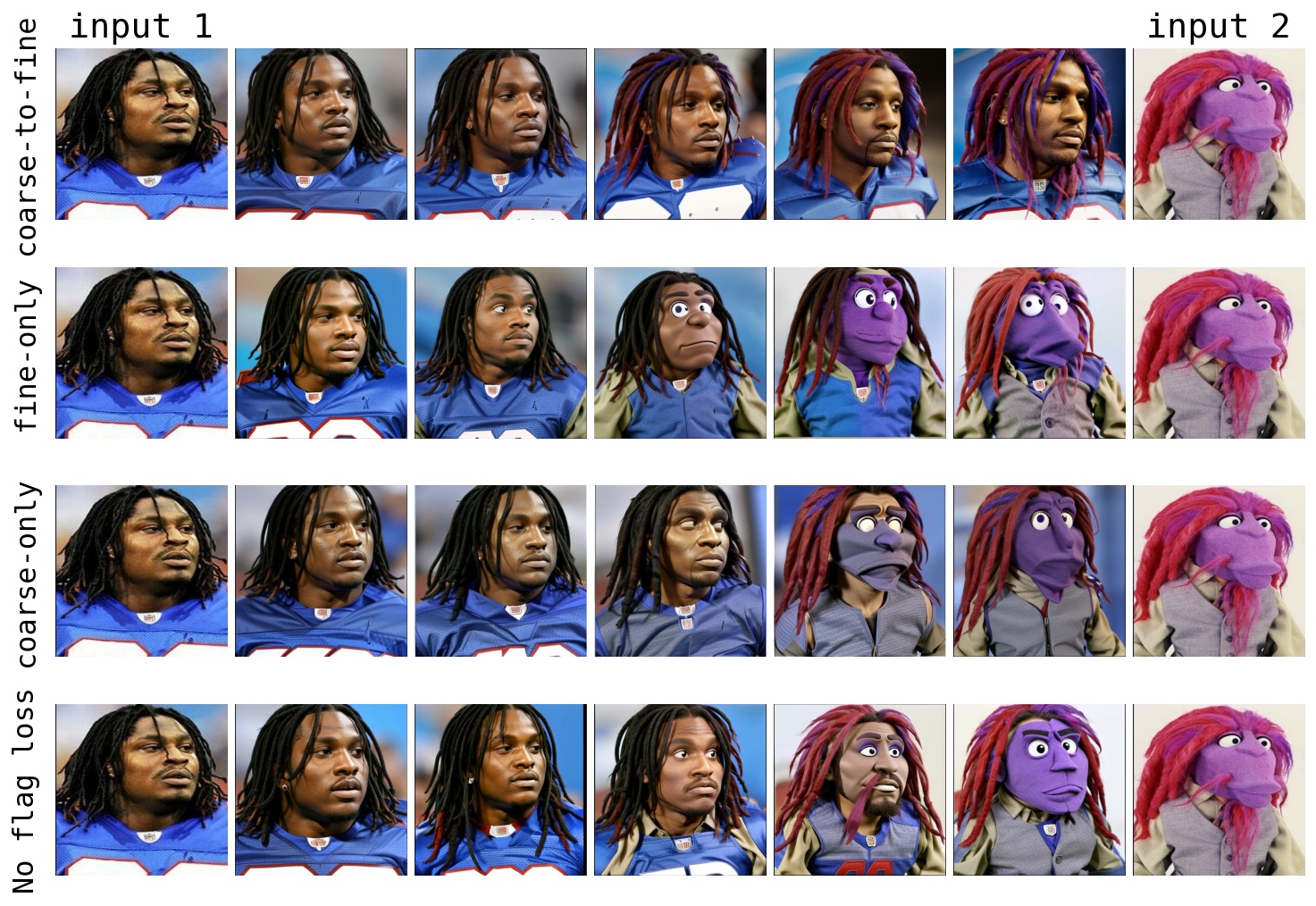}
    \vspace{5pt}
    \includegraphics[width=1\linewidth]{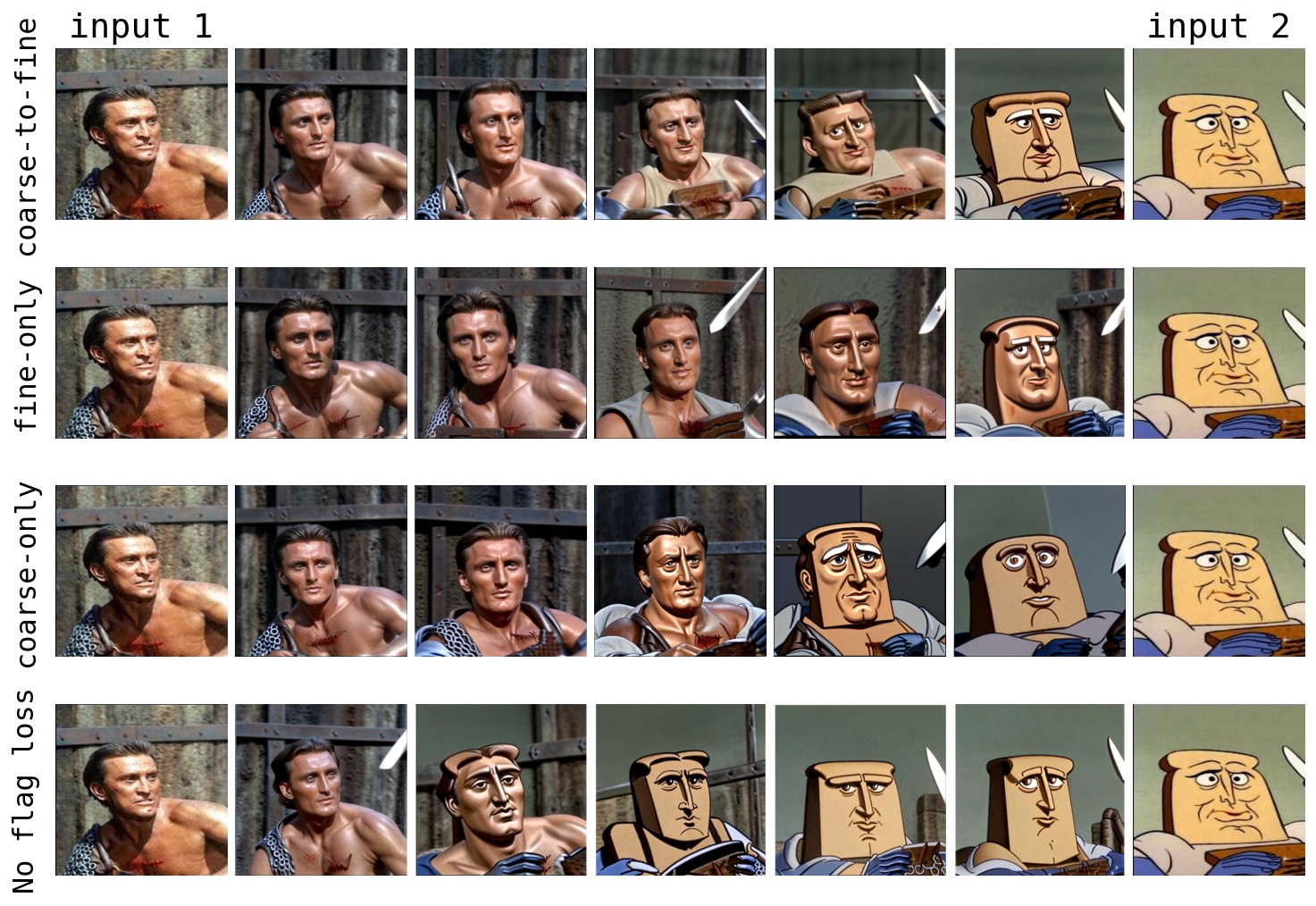}
    \vspace{5pt}
    \includegraphics[width=1\linewidth]{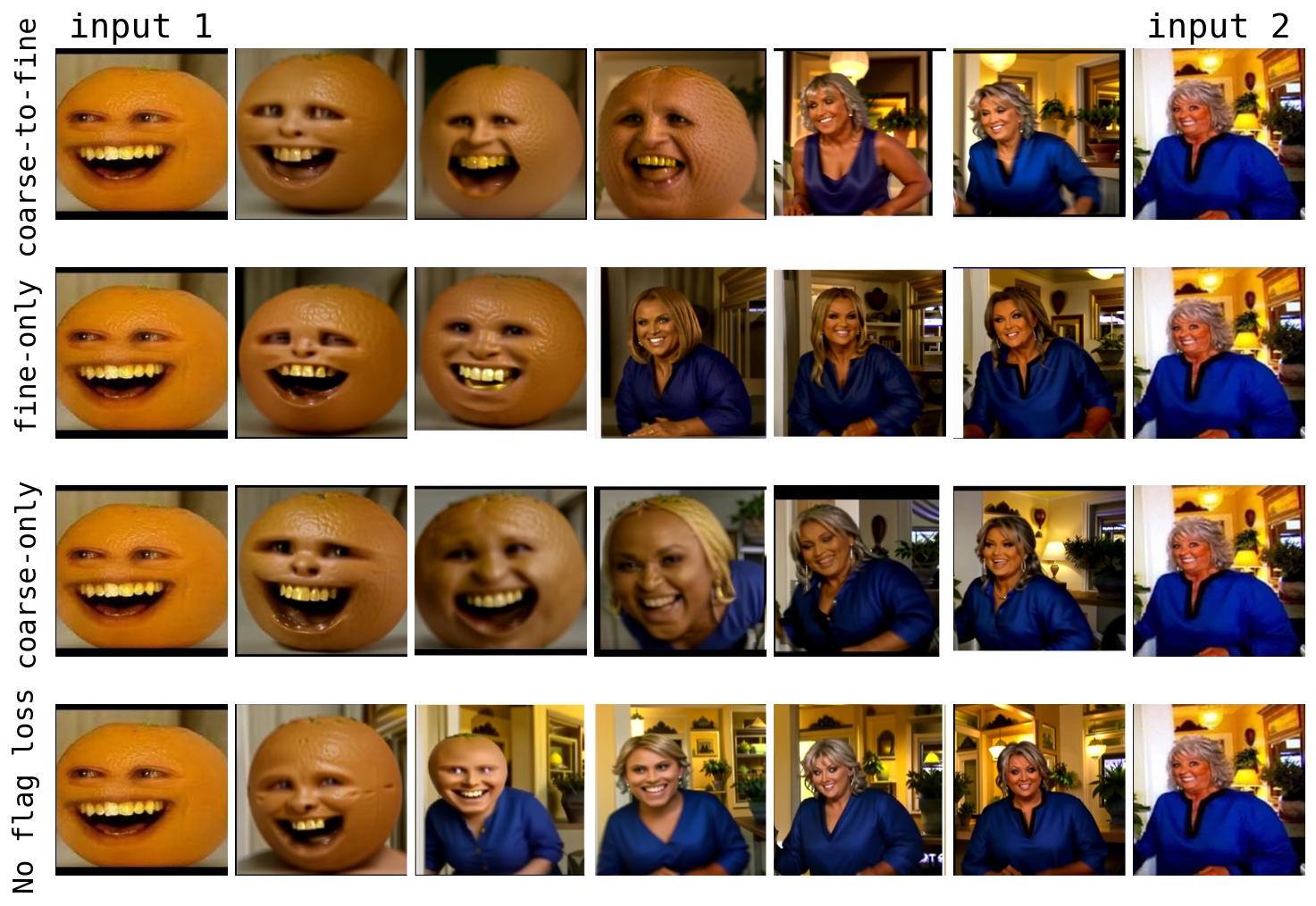}
    \vspace{-18pt}
    \caption{Ablation study on Flag Loss. \textbf{Insights:} without flag loss, the blending failed to capture the main attribute.}
    \vspace{-5pt}
    \label{fig:ablation_flag_loss} 
\end{figure}

Results are shown in \Cref{apptab:ablation-scale} and \Cref{fig:ablation_flag_loss}. Multi-scale supervision provides the most robust performance across all difficulty levels. Single-scale variants work well only in specific regimes—fine-grained-only excels on low-difficulty cases where attributes are small and localized; coarse-only works best on high-difficulty pairs where fine-grained attribute connections are hard to discover. Removing flag loss significantly degrades performance, confirming that multiscale geometric alignment is essential for coherent Vibe Blending.

\paragraph{Correspondence.}
Vibe Blending relies on region-level correspondence between DINO token clusters in $I_A$ and $I_B$ (see \Cref{sec:appendix_correspondence}). We ablate both the number of clusters and the complete removal of correspondence.

In \Cref{apptab:ablation-correspondence}, using correspondence consistently outperforms removing it on medium- and low-difficulty cases. When difficulty is high, using fewer clusters (e.g., $10$) is beneficial because high-difficulty pairs have less spatial alignment, making fine-grained clustering unstable. Without correspondence (i.e., pixel-level blending), performance is competitive only for high-difficulty pairs but fails on medium/low cases and produces incoherent blends when inputs are spatially misaligned (see qualitative examples in \Cref{sec:appendix_correspondence}). Overall, correspondence is crucial for concept-level blending.

\paragraph{DINO–CLIP Feature Mixing.}
Vibe Space is trained by mapping dense DINO features into Vibe Space and decoding back into CLIP space to leverage both fine-grained semantics (DINO) and generative compatibility (CLIP). We ablate this design by replacing DINO features with CLIP features as the encoder input.

As shown in \Cref{apptab:ablation-dinoclip}, mixing DINO and CLIP features achieves the best performance on medium- and low-difficulty pairs. For the most challenging pairs, CLIP-only encoding sometimes performs slightly better, likely because CLIP already captures high-level semantics that dominate these pairs. However, DINO–CLIP mixing remains the more stable choice across all difficulty levels.

\clearpage
\newpage

%% file: sec_appendix/path_finding.tex
\clearpage
\section{Path Finding Details}
\label{appsec:path_finding}

\subsection{Proof of Vibe Space Analytical Minimum}

This section provides the mathematical justification for the claim made in the main text:  
\emph{linear interpolation in Vibe Space decodes to an approximate geodesic in the original feature manifold.}  
We show this by starting from the formulation of the multiscale inverse diffusion map, rewriting it in terms of the flag-space kernel, and then substituting the geometric alignment constraint learned during Vibe Space training.  
This results in a latent-space surrogate objective whose unique minimizer is exactly the linear interpolation between endpoints.

\paragraph{Notation Recap.}
\begin{itemize}[leftmargin=1.5em]
\item $\mathbf{x} \in \mathbf{R}^{(HW)\times D}$ : DINO feature tokens.
\item $f$: encoder mapping tokens to Vibe Space.
\item $\mathbf{z} \in \mathbf{R}^{(HW)\times d}$: Vibe Space embeddings, $\mathbf{z}=f(\mathbf{x})$.
\item $g$: decoder mapping Vibe Space to CLIP feature space.
\item $\mathbf{\Psi}^{1:m_k}(\mathbf{x})  \in \mathbf{R}^{(HW)\times m_k}$: first $m_k$ diffusion-map eigenvectors.
\item $\mathcal{M} = \{m_1, \dots, m_k\}$: set of multiscale truncation levels.
\item $S(\mathbf{\Psi}(\mathbf{x}))$: flag-space kernel
\[
S_{ij}(\mathbf{\Psi}(\mathbf{x}))
=
\frac{1}{|\mathcal{M}|}
\sum_{m_k\in\mathcal{M}}
\mathbf{\Psi}^{1:m_k}(\mathbf{x}_i)\,
\mathbf{\Psi}^{1:m_k}(\mathbf{x}_j)^\top.
\]
\item $\mathbf{z}_A, \mathbf{z}_B$ : latent endpoints.
\item $\alpha \in [0, 1]$: interpolation weights.
\item $\gamma(\alpha)$: path in original manifold.
\item $\mathbf{z}_\alpha$: latent interpolation:
\(
\mathbf{z}_{\alpha,i}
=
(1-\alpha)\mathbf{z}_{A}
+
\alpha\,\mathbf{z}_{B}.
\)
\end{itemize}

\paragraph{Inverse Diffusion Map in Flag-Space Kernel Form}
Given interpolated diffusion map coordinates
\[
\mathbf{\Psi}_t(\mathbf{x}_\alpha)
=
(1-\alpha)\mathbf{\Psi}_t(\mathbf{x}_A)
+
\alpha\,\mathbf{\Psi}_t(\mathbf{x}_B),
\]
the multiscale inverse diffusion map seeks
\begin{equation}
\gamma(\alpha)
=
\arg\min_{\mathbf{x}^*}
\frac{1}{|\mathcal{M}|}
\sum_{m_k\in\mathcal{M}}
\big\|
\mathbf{\Psi}^{1:m_k}_t(\mathbf{x}^*)
-
\mathbf{\Psi}^{1:m_k}_t(\mathbf{x}_\alpha)
\big\|_2^2.
\tag{A.1}
\end{equation}

Expand the squared norm
\begin{align}
\big\|
\mathbf{\Psi}^{1:m_k}_t(\mathbf{x}^*)
-
\mathbf{\Psi}^{1:m_k}_t(\mathbf{x}_\alpha)
\big\|_2^2
&=
\big\|\mathbf{\Psi}^{1:m_k}_t(\mathbf{x}^*)\big\|_2^2 \\
&+
\big\|\mathbf{\Psi}^{1:m_k}_t(\mathbf{x}_\alpha)\big\|_2^2
\notag\\
&
-2\,\Big\langle
\mathbf{\Psi}^{1:m_k}_t(\mathbf{x}^*),
\mathbf{\Psi}^{1:m_k}_t(\mathbf{x}_\alpha)
\Big\rangle.
\tag{A.2}
\end{align}

Summing (A.2) over all $m_k\in\mathcal{M}$ produces three corresponding sums:

\begin{itemize}[leftmargin=1.5em]
\item 
\textbf{Term 1:}  
$\frac{1}{|\mathcal{M}|}\sum_{m_k}\|\mathbf{\Psi}^{1:m_k}_t(\mathbf{x}^*)\|_2^2$  
collects all inner products  
$\langle \mathbf{\Psi}^{1:m_k}_t(\mathbf{x}^*_i),\, \mathbf{\Psi}^{1:m_k}_t(\mathbf{x}^*_i)\rangle$  
across scales.  
Across tokens, these are exactly the \emph{diagonal entries} of the flag-space kernel  
$S(\mathbf{\Psi}(\mathbf{x}^*))$.

\item
\textbf{Term 2:}  
$\frac{1}{|\mathcal{M}|}\sum_{m_k}\|\mathbf{\Psi}^{1:m_k}_t(\mathbf{x}_\alpha)\|_2^2$  
does not depend on $\mathbf{x}^*$ and is absorbed into a constant.

\item
\textbf{Term 3 (cross term):}  
\[
\frac{2}{|\mathcal{M}|}\sum_{m_k}
\big\langle
\mathbf{\Psi}^{1:m_k}_t(\mathbf{x}^*),
\mathbf{\Psi}^{1:m_k}_t(\mathbf{x}_\alpha)
\big\rangle.
\]
Element-wise, each pair of tokens $(i,j)$ contributes  
\[
\frac{1}{|\mathcal{M}|}
\sum_{m_k}
\mathbf{\Psi}^{1:m_k}(\mathbf{x}^*_i)\,
\mathbf{\Psi}^{1:m_k}(\mathbf{x}_{\alpha,j})^\top,
\]
which is precisely the $(i,j)$ entry of the kernel  
$S(\mathbf{\Psi}(\mathbf{x}^*))$ matched to  
$S(\mathbf{\Psi}(\mathbf{x}_\alpha))$.
\end{itemize}

Putting the three terms together, the entire objective in (A.1) is the Frobenius norm between the two corresponding flag-space kernel matrices:
\begin{equation}
\gamma(\alpha)
=
\arg\min_{\mathbf{x}^*}
\big\|
S(\mathbf{\Psi}(\mathbf{x}^*)) - S(\mathbf{\Psi}(\mathbf{x}_\alpha))
\big\|_F^2.
\tag{A.3}
\end{equation}

Thus the inverse diffusion map reduces to matching multiscale flag-space kernels.

\paragraph{Vibe Space Approximation.}
Vibe Space training enforces the alignment between Gram matrix $\mathbf{z}\mathbf{z}^\top$ and flag-space kernel $S(\mathbf{\Psi}(\mathbf{x}))$
\begin{equation}
\mathbf{z}\mathbf{z}^\top \approx S(\mathbf{\Psi}(\mathbf{x})),
\qquad
\mathbf{z} = f(\mathbf{x}).
\tag{A.4}
\end{equation}

Let $\mathbf{z}^* = f(\mathbf{x}^*)$ and $\mathbf{z}_\alpha = (1-\alpha)\,\mathbf{z}_{A}+\alpha\,\mathbf{z}_{B}$.
Substituting (A.4) into (A.3) gives the latent surrogate:
\begin{equation}
\boxed{
\mathbf{z}^*
=
\arg\min_{\mathbf{z}}
\big\|
\mathbf{z}\mathbf{z}^\top - \mathbf{z}_\alpha\mathbf{z}_\alpha^\top
\big\|_F^2
}.
\tag{A.5}
\end{equation}

Because the Gram matrix $\mathbf{z}\mathbf{z}^\top$ matches
$\mathbf{z}_\alpha\mathbf{z}_\alpha^\top$ uniquely when $\mathbf{z}=\mathbf{z}_\alpha$, the unique minimal is
\begin{equation}
\boxed{
\mathbf{z}^* = \mathbf{z}_\alpha
}.
\tag{A.6}
\end{equation}

Thus $\,\gamma(\alpha)\approx g(\mathbf{z}_\alpha)$, i.e.,
\emph{linear interpolation in Vibe Space provides a closed-form approximation
to the multiscale inverse-diffusion geodesic.}

\subsection{Graph Laplacian and Diffusion Map}
\label{app:graph_laplacian}

\paragraph{Affinity graph and Laplacian.}
Let $\{\mathbf{x}_i\}_{i=1}^{N}$ denote the DINO feature tokens used as graph nodes.
We construct a weighted affinity graph with
\begin{equation}
\mathbf{W}_{ij}
=
\exp\!\left(
-\frac{\|\mathbf{x}_i - \mathbf{x}_j\|_2^2}{\sigma^2}
\right),
\qquad
\mathbf{D}_{ii}
=
\sum_{j=1}^N \mathbf{W}_{ij},
\tag{A.7}
\end{equation}
where $\sigma > 0$ is a bandwidth parameter.
We set 
\(
\sigma^2 = \sum_{d} \operatorname{Var}(x_i^{(d)})
\)
to match the global feature variance, ensuring that the affinity kernel adapts to the scale of the DINO feature distribution.

The graph Laplacian is
\begin{equation}
\mathbf{L} = \mathbf{D} - \mathbf{W}.
\tag{A.8}
\end{equation}
The corresponding random-walk diffusion operator is the row-stochastic matrix
\begin{equation}
\mathbf{P} = \mathbf{D}^{-1}\mathbf{W},
\tag{A.9}
\end{equation}
where $\mathbf{P}_{ij}$ gives the one-step transition probability from node $i$ to node $j$.

\paragraph{Nyström approximation.}
Constructing the full affinity matrix $\mathbf{W}\in\mathbb{R}^{N\times N}$ is $\mathcal{O}(N^{2})$ in memory and compute, which is prohibitive for tens of thousands of DINO tokens per image.
We therefore employ the Nyström approximation \cite{ncut-pytorch}:

\begin{itemize}[leftmargin=1.3em]
\item Sample a subset of $S \ll N$ anchor tokens.
\item Form the subgraph affinities $\mathbf{W}_{SS}$ exactly.
\item Compute cross-affinities $\mathbf{W}_{NS}$ between all nodes and anchors.
\item Approximate the full kernel by
\begin{equation}
\widetilde{\mathbf{W}}
=
\mathbf{W}_{NS}\,
\mathbf{W}_{SS}^{-1}\,
\mathbf{W}_{NS}^\top .
\tag{A.10}
\end{equation}
\end{itemize}

This reduces the eigen-decomposition cost from $\mathcal{O}(N^3)$ to $\mathcal{O}(S^3)$ and the kernel construction cost from $\mathcal{O}(N^2)$ to $\mathcal{O}(NS)$.
In practice, we use $S=500$; the graph eigenvectors can be computed in milliseconds.

\paragraph{Generalized eigenproblem.}
The diffusion-map coordinates are obtained by solving
\begin{equation}
\mathbf{L}\,\boldsymbol{\Psi}
=
\lambda\,\mathbf{D}\,\mathbf{\Psi},
\tag{A.11}
\end{equation}
which is equivalent to diagonalizing the diffusion operator:
\begin{equation}
\mathbf{P}\,\mathbf{\Psi}
=
(1-\lambda)\,\mathbf{\Psi}
.
\tag{A.12}
\end{equation}

%% file: figures/more_analogies.tex
\begin{figure}[!ht]
    \centering
    \includegraphics[width=\linewidth]{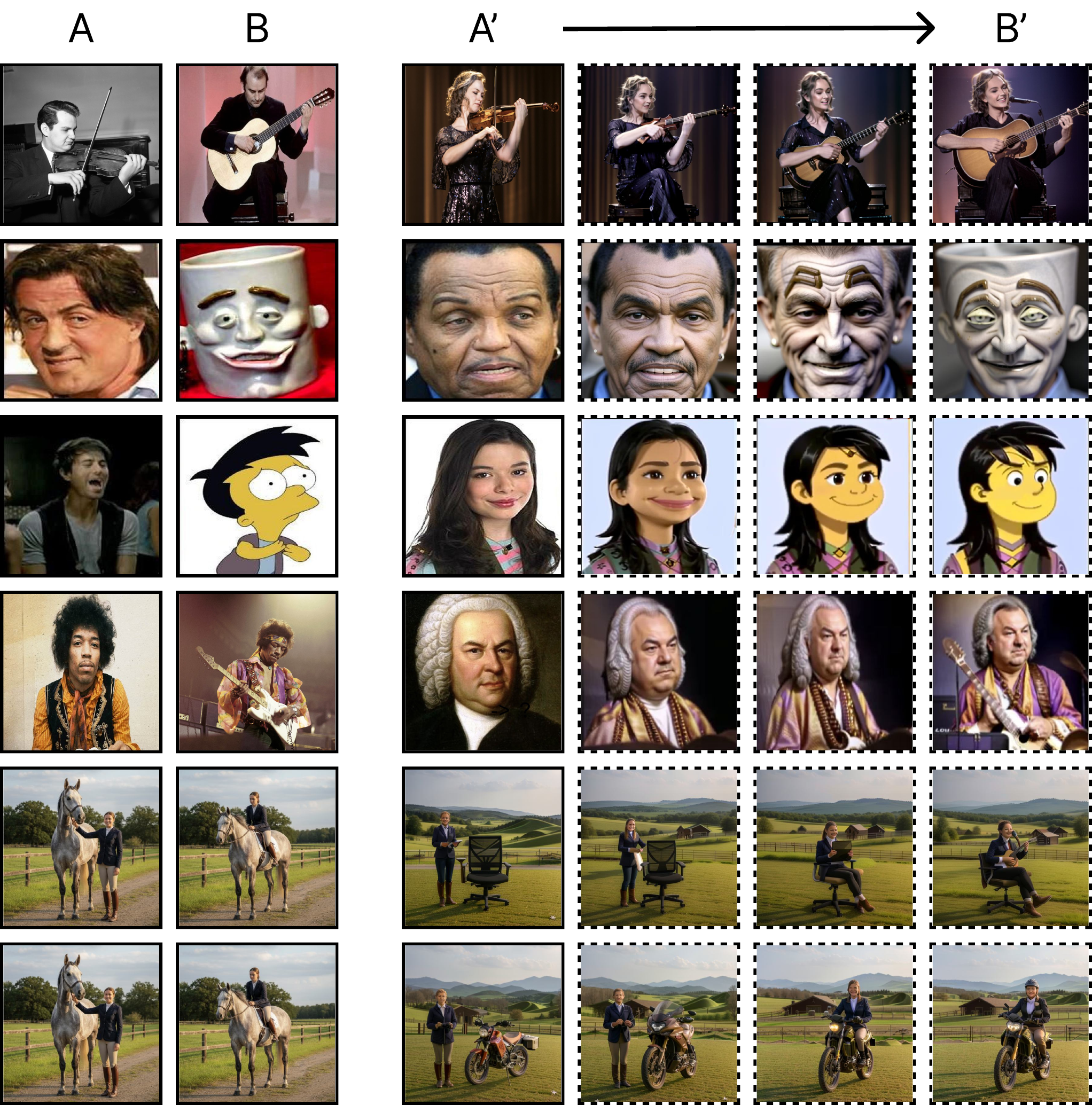}
    \vspace{-18pt}
    \caption{Vibe Analogy examples. Vibe Analogy can transfer the vibe of images $A \to B$ to another non-trivial but related image $A'$ to generate $B'$. Examples of transferred vibes are human-object interaction and art styles between similar facial expressions.}
    \label{fig:more_analogies}
\end{figure}

%% file: sec_appendix/algo_analogy.tex
\begin{algorithm}[h]
\caption{Vibe Analogy}
\label{alg:vibe-analogy}
\small{
\begin{algorithmic}[1]
\Require Images $(I_A, I_B, I_{A'})$, image features $(\mathbf{x}^{\text{dino}}, \mathbf{x}^{\text{clip}})$
\Ensure Generated intermediate images $\{I_\alpha\}_{\alpha\in[0,1]}$

\State $\mathbf{W}_{ij} = \text{exp}{(-\frac{\|\mathbf{x}_i^{\text{dino}} - \mathbf{x}_j^{\text{dino}}\|^2}{\sigma^2})}$, $\mathbf{D}_{ii} = \sum_j \mathbf{W}_{ij}$ \Comment{Graph}

\State $(\mathbf{D} - \mathbf{W})\mathbf{\Psi}(\mathbf{x}^{\text{dino}}) = \lambda \mathbf{D}\mathbf{\Psi}(\mathbf{x}^{\text{dino}})$ \Comment{Graph Diffusion Map}

\State $f, g \gets \text{Train}(\mathbf{x}^{\text{clip}}, \mathbf{x}^{\text{dino}}, \mathbf{\Psi}(\mathbf{x}^{\text{dino}}))$  \Comment{Train Vibe Space}

\State $\mathbf{z}_A = f(\mathbf{x}_A^{\text{dino}})$; $\mathbf{z}_B = f(\mathbf{x}_B^{\text{dino}})$; \\$\mathbf{z}_{A'} = f(\mathbf{x}_{A'}^{\text{dino}})$\Comment{Encode vibe}

\State $\pi_{B \leftrightarrow A} \gets \text{Match}(\mathbf{x}_A^{\text{dino}}, \mathbf{x}_B^{\text{dino}})$ \\
$\pi_{A \leftrightarrow A'} \gets \text{Match}(\mathbf{x}_A^{\text{dino}}, \mathbf{x}_{A'}^{\text{dino}})$
\Comment{Cluster correspondence}

\State $\boldsymbol{\Delta}_{A\to B} = \pi_{B \leftrightarrow A}(\mathbf{z}_B) - \mathbf{z}_A$ \\
$\boldsymbol{\Delta}_{A'\to B'} = \pi_{A \leftrightarrow A'}(\boldsymbol{\Delta}_{A\to B})$
\Comment{Path direction}

\For{$\alpha \in [0,1]$}
    \State $\mathbf{z}_\alpha = 
           \mathbf{z}_{A'} + \alpha\,\boldsymbol{\Delta}_{A'\to B'}$ \Comment{Path interpolation}
    \State $\mathbf{x}_\alpha^{\text{clip}} = g(\mathbf{z}_\alpha)$ \Comment{Decode vibe}
    \State $I_\alpha \gets 
           \text{IPAdapter}(\mathbf{x}_\alpha^{\text{clip}})$ \Comment{Generate image}
\EndFor

\end{algorithmic}
}
\end{algorithm}

%% file: figures/neg_vibes_extended.tex
\begin{figure*}[!ht]
    \centering
    \includegraphics[trim=0in 0in 0in 0in, clip,width=\textwidth]{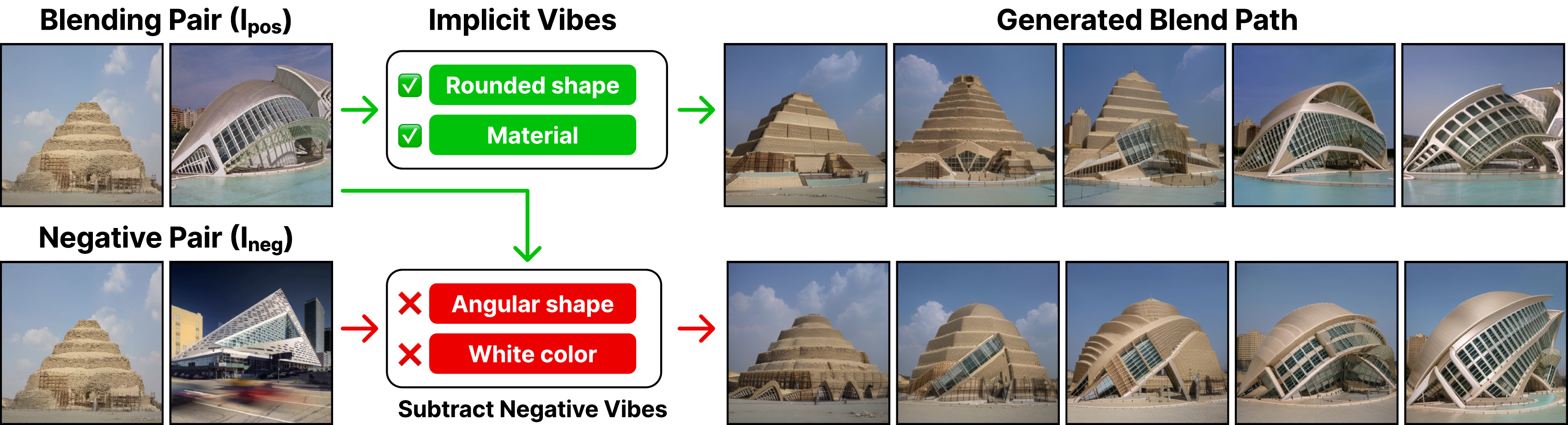}
    \vspace{-18 pt}
    \caption{Negative vibe control. Vibe attributes are implicitly extracted by Vibe Space. The blending pair defines desired vibes (rounded shape and material). The negative pair defines vibes to suppress (angular shape and white color). Blending without negative examples may transfer more attributes than desired. Subtracting the negative vibes, we better preserve the rounded shape and tan color of the building.}
    \label{fig:neg_vibes_extended}
\end{figure*}

%% file: sec_appendix/algo_negative_vibe.tex
\begin{algorithm}[h]
\caption{Vibe Blending with Negative Vibe Control}
\label{alg:negative-vibe}
\small{
\begin{algorithmic}[1]
\Require Pos images $(I_A, I_B)$ with pos image features $(\mathbf{x}_{\text{pos}}^{\text{dino}}, \mathbf{x}_{\text{pos}}^{\text{clip}})$, Neg images $(I_A, I_C)$ with neg image features $(\mathbf{x}_{\text{neg}}^{\text{dino}})$
\Ensure Generated intermediate images $\{I_\alpha\}_{\alpha\in[0,1]}$

\State $\mathbf{W}_{\text{pos}}, \mathbf{D}_{\text{pos}} \gets \text{ComputeAffinity}(\mathbf{x}_{\text{pos}}^{\text{dino}})$ \Comment{Pos graph}
\State $\mathbf{W}_{\text{neg}}, \mathbf{D}_{\text{neg}} \gets \text{ComputeAffinity}(\mathbf{x}_{\text{neg}}^{\text{dino}})$ \Comment{Neg graph}

\State $\mathbf{\Psi}_{\text{pos}} \gets \text{GraphDiffusionMap}(\mathbf{W}_{\text{pos}}, \mathbf{D}_{\text{pos}})$ \Comment{Pos eigvecs}
\State $\mathbf{\Psi}_{\text{neg}} \gets \text{GraphDiffusionMap}(\mathbf{W}_{\text{neg}}, \mathbf{D}_{\text{neg}})$ \Comment{Neg eigvecs}

\State $\mathbf{\Psi}_{\text{filtered}} \gets \mathbf{\Psi}_{\text{pos}} - \beta \cdot \mathbf{\Psi}_{\text{neg}} ( \mathbf{\Psi}_{\text{neg}}^\top \mathbf{\Psi}_{\text{pos}} )$ \Comment{Orthogonalize}

\State $f, g \gets \text{Train}(\mathbf{x}_{\text{pos}}^{\text{clip}}, \mathbf{x}_{\text{pos}}^{\text{dino}}, \mathbf{\Psi}_{\text{filtered}})$  \Comment{Train Vibe Space}

\State $\mathbf{z}_A \gets f(\mathbf{x}_A^{\text{dino}})$; $\mathbf{z}_B \gets f(\mathbf{x}_B^{\text{dino}})$ \Comment{Encode pos vibes}

\State $\pi \gets \text{Match}(\mathbf{x}_A^{\text{dino}}, \mathbf{x}_B^{\text{dino}})$ \Comment{Cluster correspondence}
\State $\Delta_{A \to B} \gets \pi(\mathbf{z}_B) - \mathbf{z}_A$ \Comment{Path direction in filtered space}

\For{$\alpha \in [0, 1]$}
    \State $\mathbf{z}_\alpha \gets \mathbf{z}_A + \alpha \cdot \Delta_{A \to B}$ \Comment{Path interpolation}
    \State $\mathbf{x}_\alpha^{\text{clip}} \gets g(\mathbf{z}_\alpha)$ \Comment{Decode vibe}
    \State $I_\alpha \gets \text{IPAdapter}(\mathbf{x}_\alpha^{\text{clip}})$ \Comment{Generate image}
\EndFor
\end{algorithmic}
}
\end{algorithm}

%% file: figures/grad_channels.tex
\begin{figure*}[!th]
    \vspace{-18 pt}
    \centering
    \includegraphics[width=\linewidth]{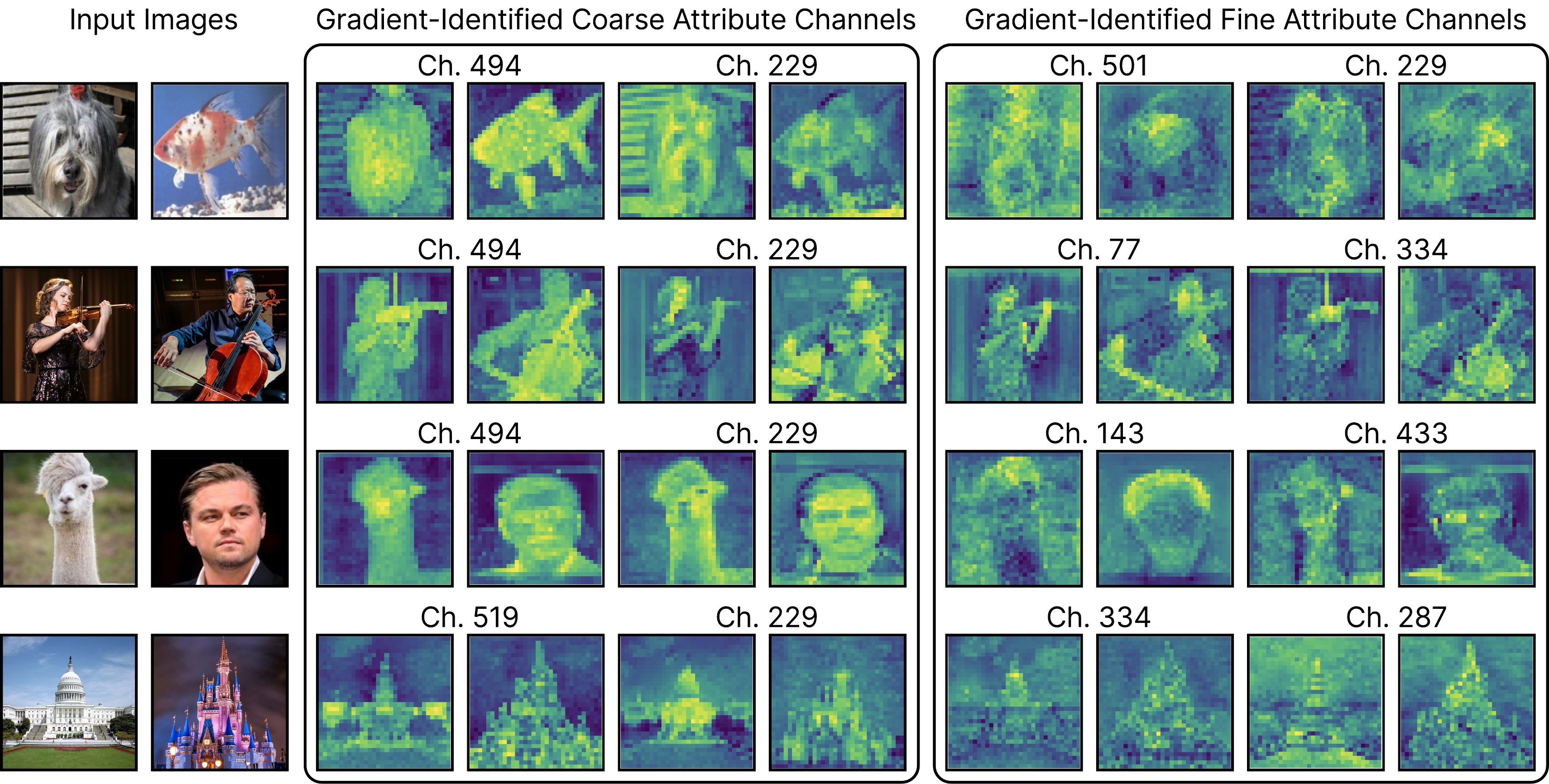}
    \vspace{-18 pt}
    \caption{Heatmaps illustrating the top-ranked DINO channels corresponding to the coarse and fine attributes shared by the input images. For example, in the second row, the coarse attributes may include the body pose and object identity, while the fine-grained attributes include the hand-object interaction.}
    \vspace{-10 pt}
    \label{fig:grad_channels}
\end{figure*}

%% file: figures/clip_dip_plot.tex
\begin{figure}[!ht]
    \centering
    \includegraphics[width=\linewidth]{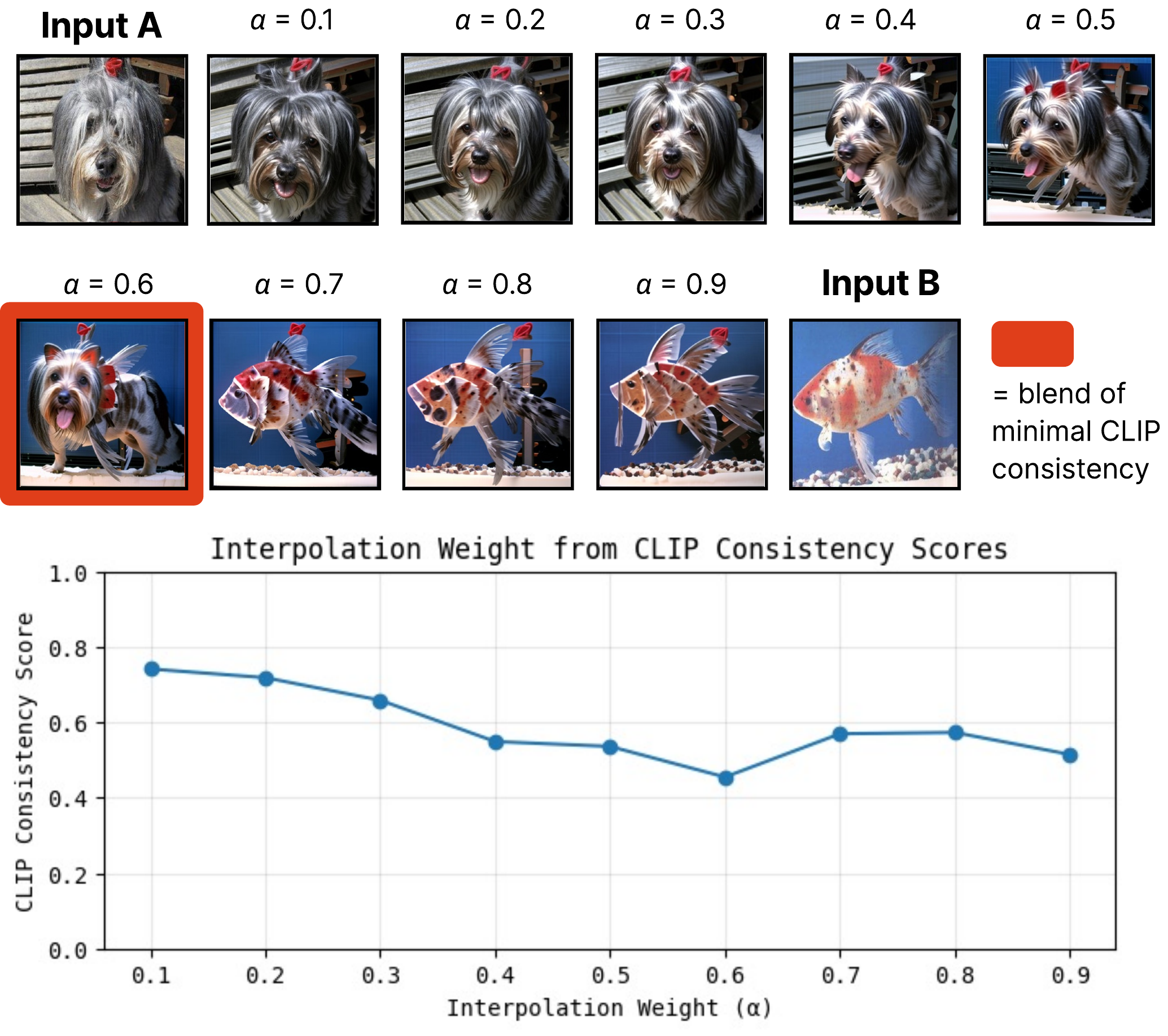}
    \vspace{-20pt}
    \caption{Our method generates a continuous path of blends for any interpolation weight $\alpha \in [0, 1]$. We propose an automated procedure to select the best $\alpha$ by identifying the point of greatest conceptual transition along the blend path. Specifically, we compute a \emph{consistency score} to measure the difference between an ``ideal'' path decoded from Vibe Space and a ``realized'' path obtained by re-encoding the generated images. In this example, the optimal weight $\alpha^* = 0.6$ as determined by the ``dip'' in the score. Qualitatively, we observe that the $\alpha^*$ obtained via this algorithm achieves the best creative blend of the inputs, and we use this approach for our evaluation.}
    \vspace{-14pt}
    \label{fig:clip_dip_plot}
\end{figure}

%% file: figures/clip_dip.tex
\begin{figure*}[!th]
    \vspace{-18 pt}
    \centering
    \includegraphics[trim=0in 0in 0in 0in, clip,width=\textwidth]{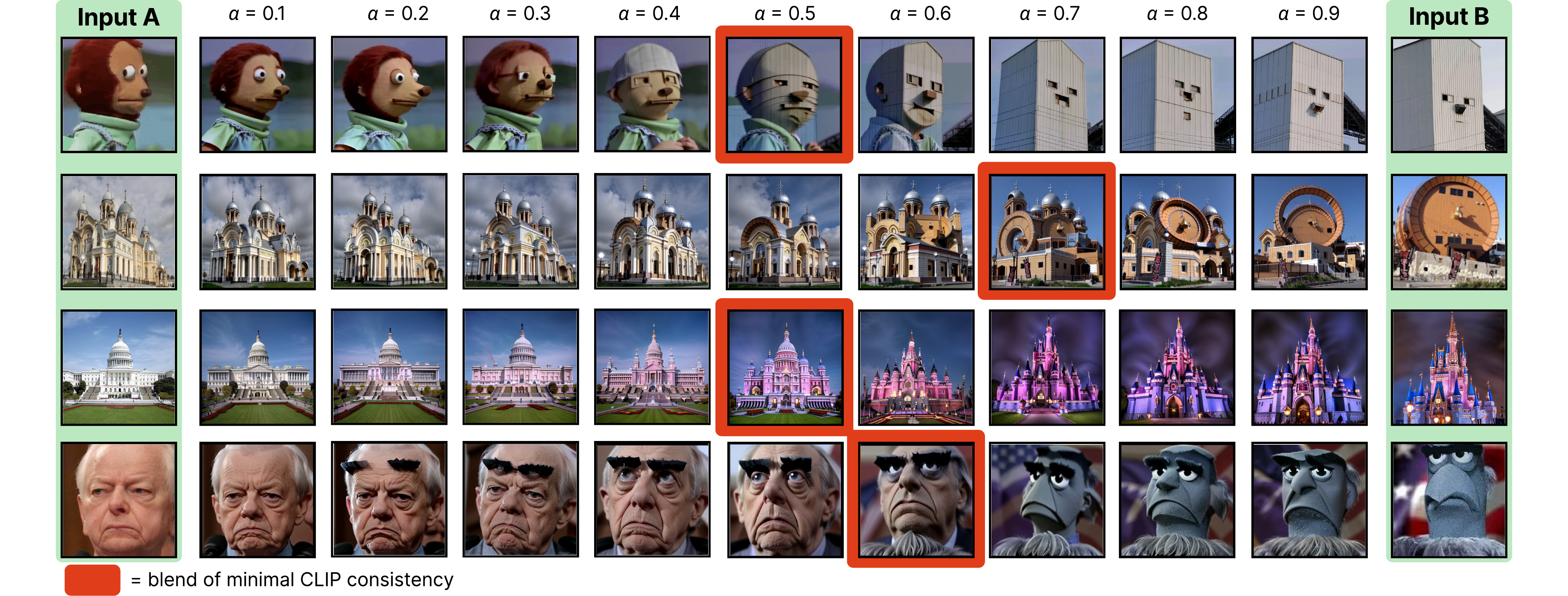}
    \vspace{-18 pt}
    \caption{Examples of blended images selected via the optimal blend weight $\alpha^*$ determined by our ``dip'' in CLIP consistency score, described in \Cref{fig:clip_dip_plot}. We observe that the ``dip'' in consistency score generally results in the best blend of the relevant attributes in both input images.}
    \vspace{-10 pt}
    \label{fig:clip_dip}
\end{figure*}

%% file: figures/extrapolate.tex
\begin{figure}[!ht]
    \centering
    \includegraphics[width=\linewidth]{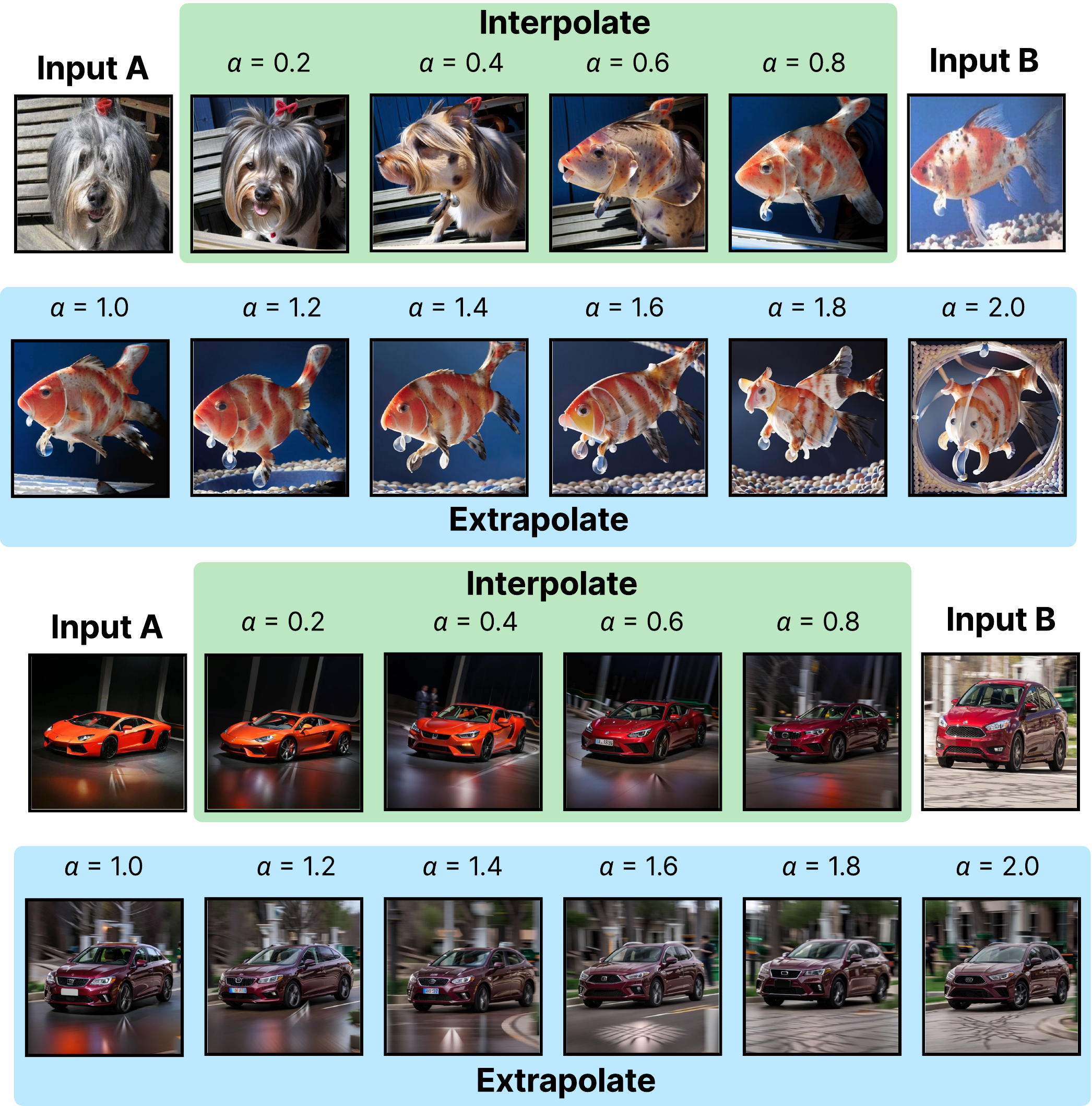}
    \vspace{-20pt}
    \caption{
    Paths computed by Vibe Blending can be extrapolated to exaggerate attributes. \textbf{Top:} Extrapolating the path from a dog to a fish emphasizes the body shape of the fish and some elements of the background. However, extrapolation behavior is not well controlled at higher weights. \textbf{Bottom:} Extrapolating from an orange sports car to a red sedan continues the color shift, resulting in further darkening.
    }
    \label{fig:extrapolate}
\end{figure}

%% file: figures/dog_ram_horse.tex
\begin{figure}[!ht]
    \centering
    \includegraphics[width=\linewidth]{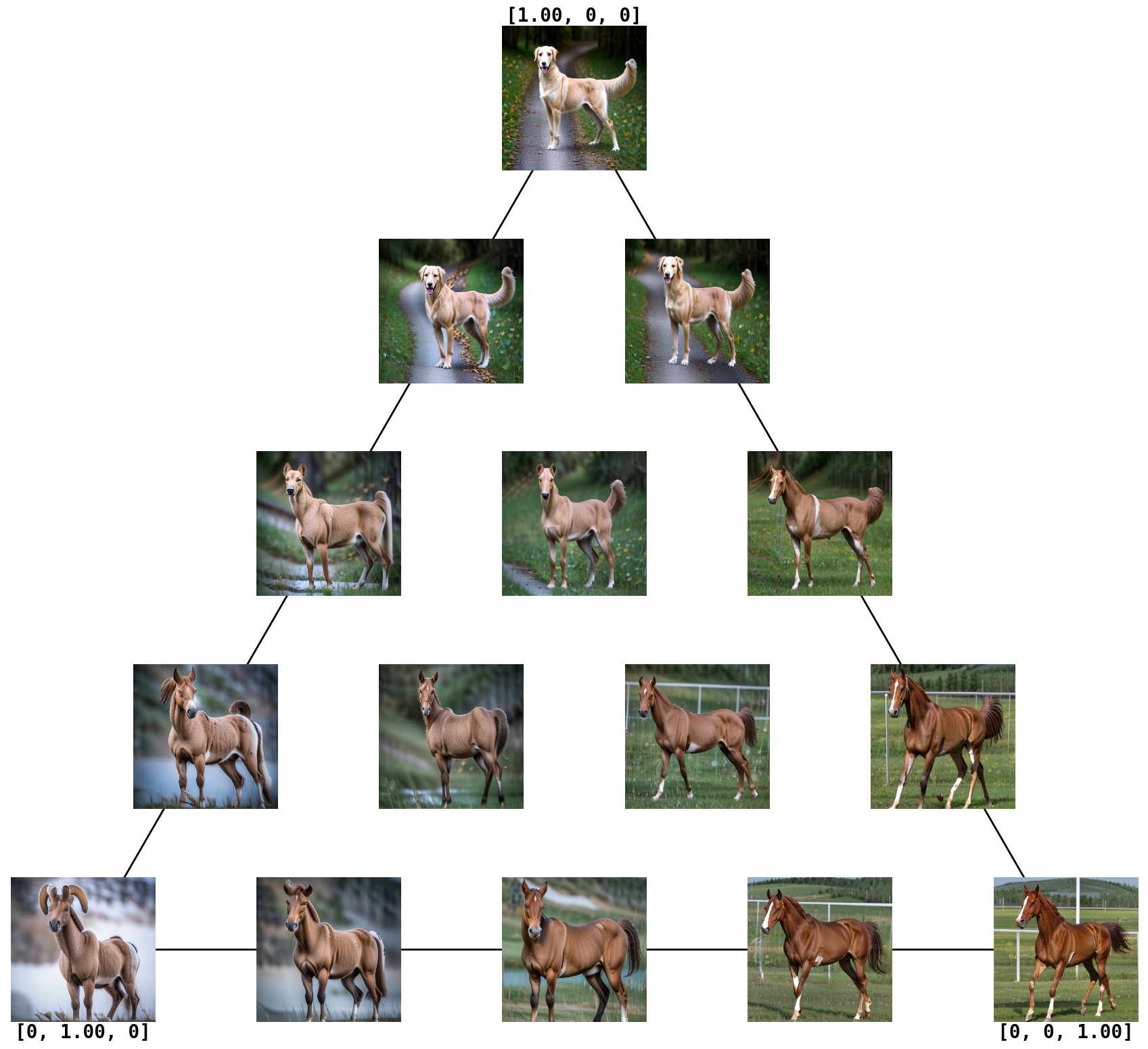}
    \vspace{0pt}
    \caption{
    N-Image Blending extends Vibe Blending to an arbitrary number of images. In this visualization, we show 3-Image Blending between a dog, ram, and horse. The top image (dog) is used as the base image to anchor correspondence matching and blending.
    }
    \vspace{0pt}
    \label{fig:n_image_triangle}
\end{figure}

%% file: sec_appendix/implementation_details.tex
\section{Implementation Details}
\label{sec:implementation_details}

\subsection{Training Procedure}
\label{sec:training_procedure}
We train the encoder--decoder $(f, g)$ for each image pair independently using a lightweight MLP architecture.

\begin{itemize}
    \item \textbf{Optimizer:} Adam (learning rate = 0.001).
    \item \textbf{Batch size:} 2 images.
    \item \textbf{Total iterations:} 1000 steps.
    \item \textbf{MLP layers:} 4.
    \item \textbf{MLP hidden dimension:} 256.
    \item \textbf{Number of Parameters:} 0.72M.
\end{itemize}

Vibe Blending running time is under 30 seconds on a RTX4090 GPU. In \Cref{alg:vibe-extra-image} lines 1-2, solving the graph diffusion map for 2-5 images (512 input image resolution, 1024 DINO tokens each image) only takes milliseconds with the help of Nyström approximation \cite{ncut-pytorch}. In \Cref{alg:vibe-extra-image} line 3, training encoder-decoder MLPs for 1000 steps takes 15 seconds. In \Cref{alg:vibe-extra-image} line 5, computing correspondence matching takes milliseconds. In \Cref{alg:vibe-extra-image} line 10, image generation with Stable Diffusion takes 2 seconds per image. 

Memory usage is under 1GB for graph diffusion map and training the Vibe Space. Memory usage is 12GB after loading the Stable Diffusion model.

\subsection{Loss Balancing}
\label{sec:loss_balancing}

Our full training objective is:
\begin{equation}
\begin{split}
\mathcal{L} &=
\lambda_{\text{flag\_enc}} \mathcal{L}_{\text{flag\_enc}}
+ \lambda_{\text{flag\_dec}} \mathcal{L}_{\text{flag\_dec}} \\
&+ \lambda_{\text{sample}} \mathcal{L}_{\text{sample}}
+ \lambda_{\text{recon}} \mathcal{L}_{\text{recon}}
\end{split}
\end{equation}

\noindent The loss terms include:
\begin{itemize}
    \item $\mathcal{L}_{\text{flag\_enc}}$, $\mathcal{L}_{\text{flag\_dec}}$: multiscale flag-space geometry preservation for the encoder $f$ and decoder $g$ respectively.
\[
\begin{aligned}
\mathcal{L}_{\text{flag\_enc}}(f)
&= \big\|\mathbf{z}\mathbf{z}^\top - \mathbf{S}(\mathbf{\Psi}(\mathbf{x}))\big\|_2^2, \\
\mathcal{L}_{\text{flag\_dec}}(f, g)
&= \big\|\mathbf{z}\mathbf{z}^\top - \mathbf{S}(\mathbf{\Psi}(g(\mathbf{z})))\big\|_2^2,
\end{aligned}
\qquad
\raisebox{0.5ex}{%
\(\begin{aligned}
\mathbf{z} &= f(\mathbf{x}),
\end{aligned}\)
}
\]
    \item $\mathcal{L}_{\text{sample}}$: flag-space geometry consistency in extrapolated space, for the decoder.
\[
\mathcal{L}_{\text{sample}}(g)
= \big\|\mathbf{z}_{\text{sample}}\mathbf{z}_{\text{sample}}^\top
- \mathbf{S}\!\left(\mathbf{\Psi}\!\big(g(\mathbf{z}_{\text{sample}})\big)\right)\big\|_2^2.
\]
    \item $\mathcal{L}_{\text{recon}}$: Reconstruct CLIP features.
\[
\mathcal{L}_{\text{recon}}(f,g)=\|\mathbf{x}^{\text{clip}}-g(f(\mathbf{x}^{\text{dino}}))\|_2^2.
\]
\end{itemize}

\noindent Loss weight balancing:
\[
\lambda_{\text{flag\_enc}} = 1, \quad
\lambda_{\text{flag\_dec}} = 0.01, \quad
\lambda_{\text{sample}} = 0.01, \quad
\lambda_{\text{recon}} = 1, \quad
\]

We observe that $\mathcal{L}_{\text{flag\_dec}}$ and $\mathcal{L}_{\text{sample}}$ can overwhelm the CLIP feature reconstruction loss $\mathcal{L}_{\text{recon}}$, this leads to poor image generation quality. Thus, we down-weighted $\mathcal{L}_{\text{flag\_dec}}$ and $\mathcal{L}_{\text{sample}}$ by a factor of 0.01.
Additionally, $\mathcal{L}_{\text{sample}}$ can cause training instability (NaN) during the warm-up periods, thus we only apply $\mathcal{L}_{\text{sample}}$ after first 500 steps of training iterations.

%% file: prompts/prompt_gemini_blend.tex
\begin{tcolorbox}[mypromptbox, title=Gemini 2.5 Flash Image Blending Prompt]
($I_A \to \text{T}$) Identify the object in this first input image.

\vspace{6pt}
\hrule
\vspace{6pt}

($I_B \to \text{T}$) Identify the object in this second input image.

\vspace{6pt}
\hrule
\vspace{6pt}

($\varnothing \to \text{T}$) I want to blend the objects in both images so their main visual attributes merge into a single, coherent hybrid. First, identify the main visual attributes that are shared by the input images, such as facial expression, hair style, or object shape. Additionally, interpret what the blend of objects might be intended to convey.

\vspace{6pt}
\hrule
\vspace{6pt}

($\varnothing \to I_{blend}$) Based on all previous turns, generate an image that blends the main attributes in the input images. The output should not be a collage, overlay, or style transfer. You may modify any image region.
\end{tcolorbox}

%% file: prompts/prompt_gpt_blend.tex
\begin{tcolorbox}[mypromptbox, title=GPT Image 1 Blending Prompt]
($I_A, I_B \to I_{blend}$) Generate an image that blends the objects in the input images so their main visual attributes merge into a single, coherent hybrid. First, identify the main visual attributes that are shared by the input images, such as facial expression, hair style, or object shape. Second, blend the main attributes in the images. The output should not be a collage, overlay, or style transfer. You may modify any image region.
\end{tcolorbox}

%% file: figures/gemini_blend.tex
\begin{figure*}[!th]
    \vspace{-10pt}
    \centering
    \includegraphics[trim=0in 0in 0in 0in, clip,width=\textwidth]{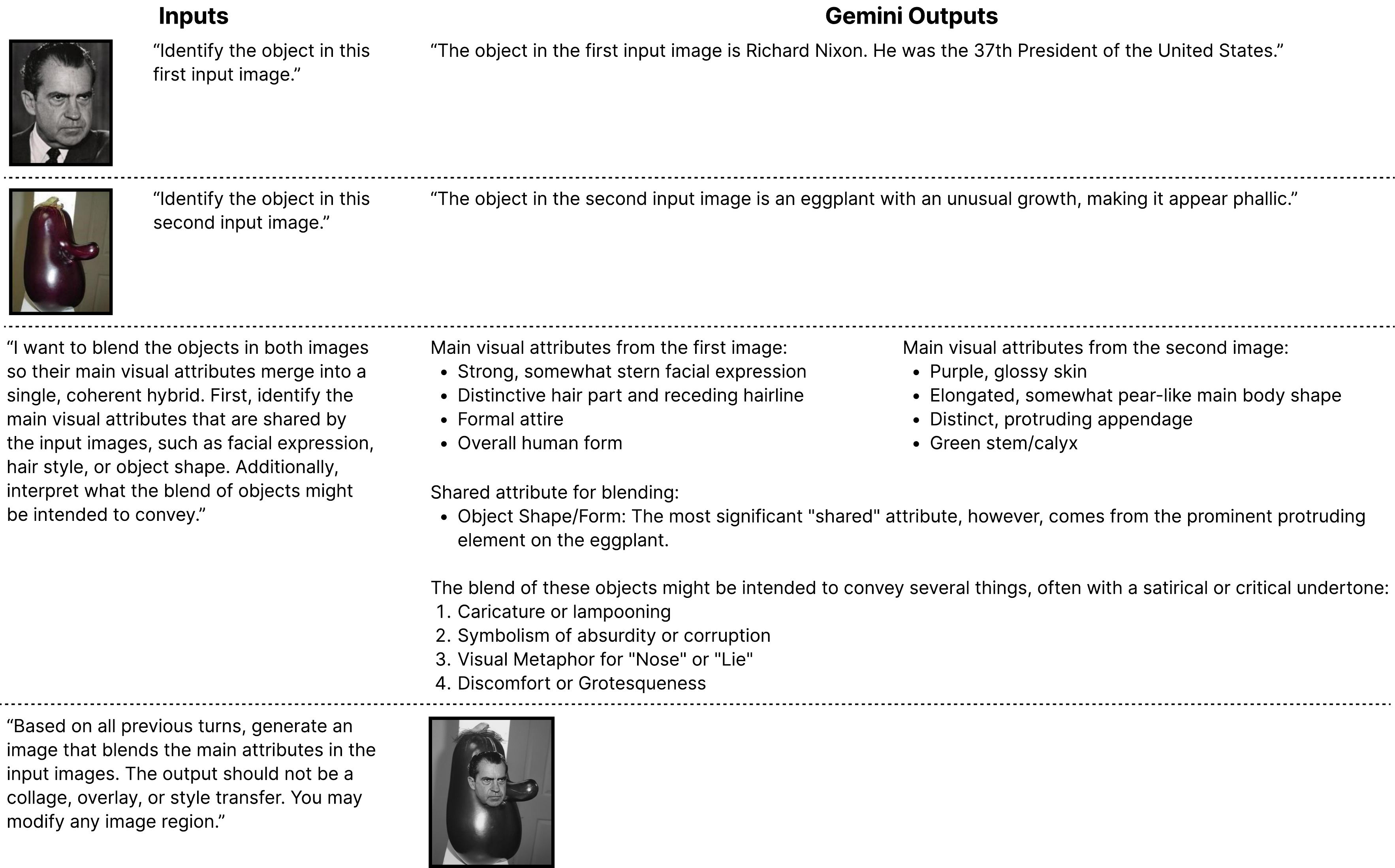}
    \vspace{-18pt}
    \caption{Example chat with Gemini \cite{comanici2025gemini} to generate a creative blend.}
    \vspace{-10pt}
    \label{fig:gemini_blend}
\end{figure*}

%% file: prompts/prompt_llm_judge.tex
\begin{tcolorbox}[
    mypromptbox,
    title=LLM Judge Evaluation Prompt,
    breakable,
    title after break={LLM Judge Evaluation Prompt (continued)}
]

($I_A \to \text{T}$) Identify the object in this first input image.

\vspace{6pt}
\hrule
\vspace{6pt}

($I_B \to \text{T}$) Identify the object in this second input image.

\vspace{6pt}
\hrule
\vspace{6pt}

($\varnothing \to \text{T}$) I want to blend the objects in both images so their visual attributes merge into a single, coherent hybrid concept. First, identify the main visual attributes that are similar between the inputs, such as facial expression, hair style, or object shape. Additionally, interpret what the blend of objects might be intended to convey.

\vspace{6pt}
\hrule
\vspace{6pt}

($\varnothing \to \text{T}$) Assess how well this first output image blends the main attributes in both input images.

\vspace{6pt}
\hrule
\vspace{6pt}

($\varnothing \to \text{T}$) Assess how well this second output image blends the main attributes in both input images.

\vspace{6pt}
\hrule
\vspace{6pt}

($\varnothing \to \text{T}$) Assess how well this third output image blends the main attributes in both input images.

\vspace{6pt}
\hrule
\vspace{6pt}

($\varnothing \to \text{T}$) Assess how well this fourth output image blends the main attributes in both input images.

\vspace{6pt}
\hrule
\vspace{6pt}

($\varnothing \to \text{T}$) Based on all previous turns, which output image (1, 2, 3, or 4) best blends the main attributes in both input images? Do not pick any output that does not perform blending, such as a collage, overlay, style transfer, or juxtaposition of the inputs. Do not pick any output that preserves the exact structure or identity of the inputs. Do not pick any output that is a part-level composition, where parts of one input are simply attached to parts of the other input. Provide your reasoning.
\end{tcolorbox}

%% file: figures/llm_eval.tex
\begin{figure*}[!th]
    \vspace{-10pt}
    \centering
    \includegraphics[trim=0in 0in 0in 0in, clip,width=\textwidth]{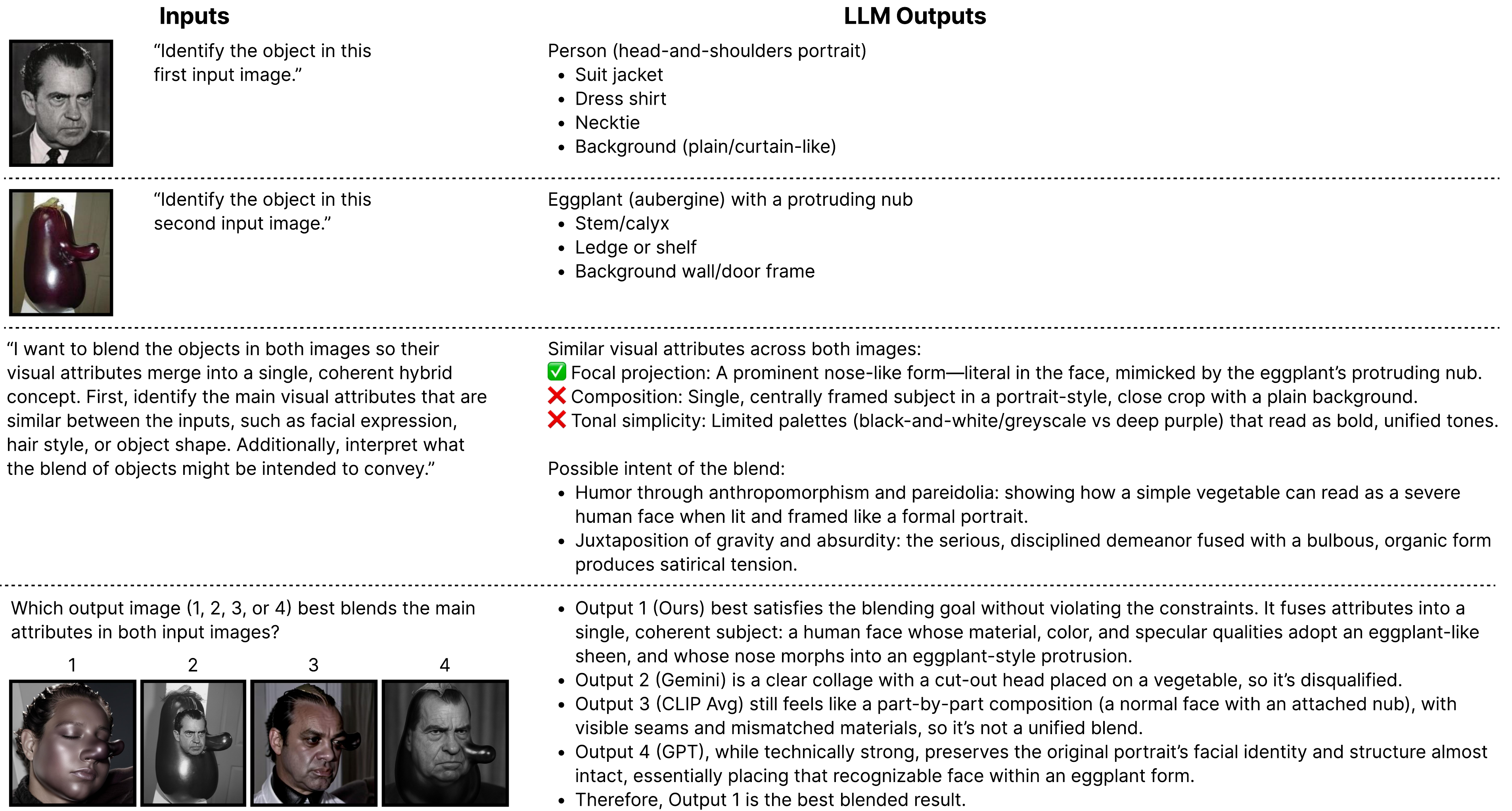}
    \vspace{-18pt}
    \caption{Example chat with the LLM judge \cite{gpt-5} to select the best blended output.}
    \label{fig:llm_eval}
\end{figure*}

%% file: figures/human_llm_agreement.tex
\begin{figure*}[!th]
    \centering
    \includegraphics[trim=0in 0in 0in 0in, clip,width=\textwidth]{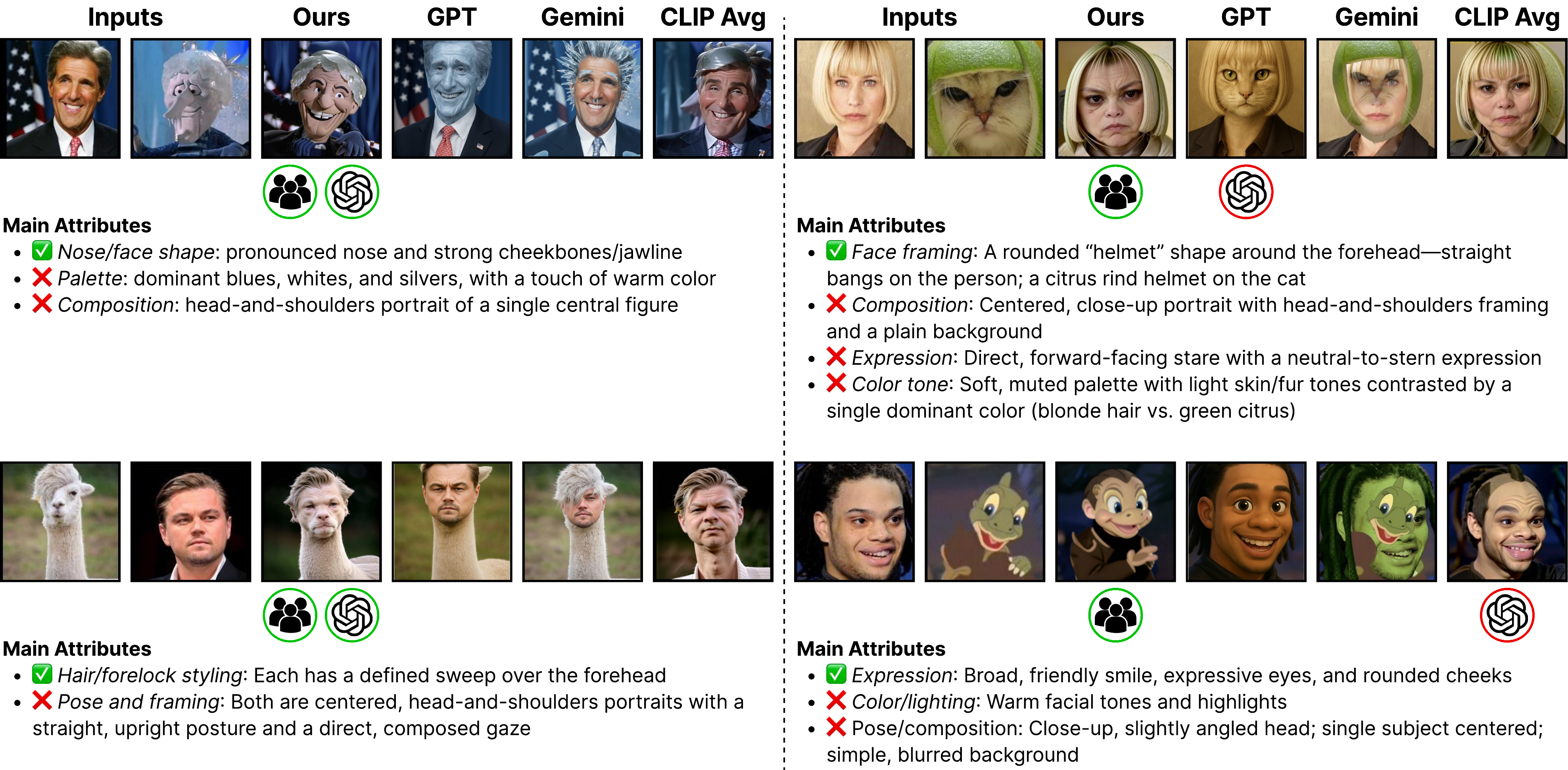}
    \vspace{-18pt}
    \caption{Examples where human raters and the LLM judge agree (\textbf{left}) and disagree (\textbf{right}) on the best blended output.}
    \label{fig:human_llm_agreement}
\end{figure*}

%% file: figures/full_human_creativity.tex
\begin{figure*}[!th]
    \vspace{-10pt}
    \centering
    \includegraphics[trim=0in 0in 0in 0in, clip,width=\textwidth]{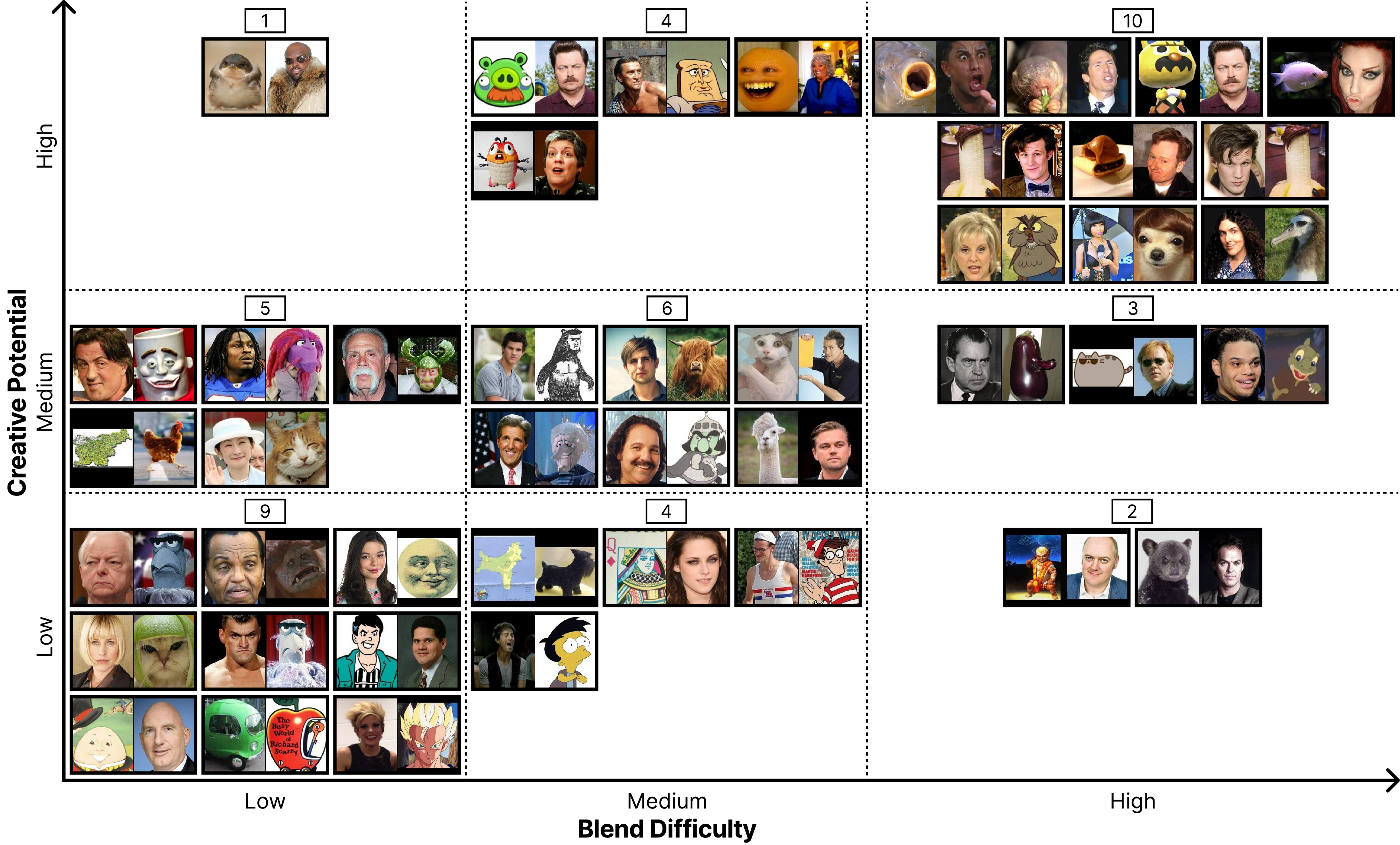}
    \vspace{-18pt}
    \caption{Extension of \Cref{fig:human_creativity}. To gauge human perceptions of creativity, we ask raters to compare image pairs along two axes: \emph{Creative Potential} refers to how interesting a blend might be, and \emph{Blend Difficulty} indicates how challenging it is to form a coherent blend. Image pairs with higher Blend Difficulty tend to have higher Creative Potential and are often more conceptually different. Numbers in each cell indicate the number of examples.}
    \vspace{-10pt}
    \label{fig:full_human_creativity}
\end{figure*}

%% file: figures/user_study_form.tex
\begin{figure*}[!th]
    \vspace{-10pt}
    \centering
    \includegraphics[trim=0in 0in 0in 0in, clip,width=\textwidth]{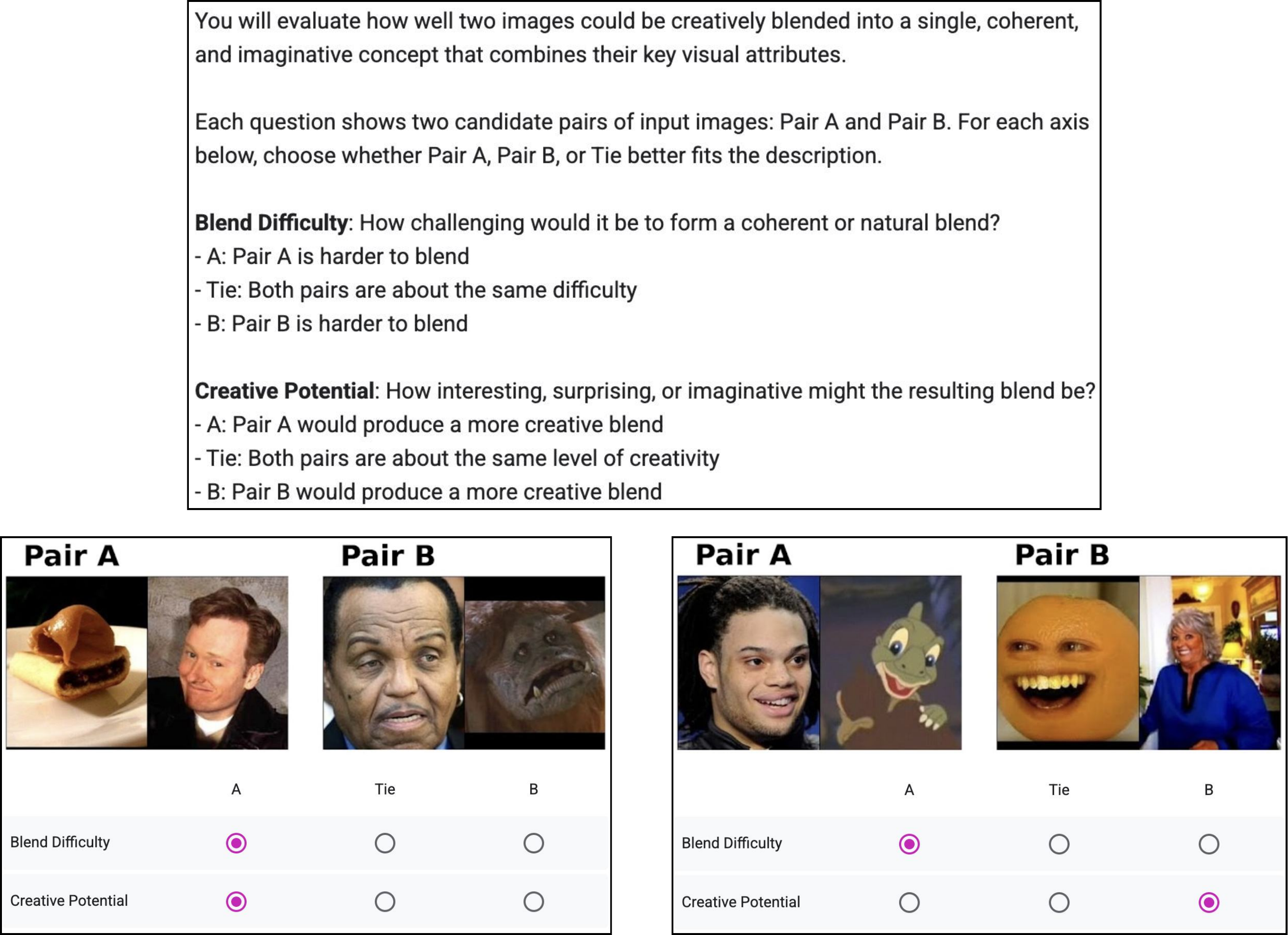}
    \vspace{-18pt}
    \caption{User study for evaluating the Creative Potential and Blend Difficulty of image pairs. \textbf{Top:} The prompt shown to users. \textbf{Bottom:} Two examples with high rater agreement. Participants select which image pair has higher Creative Potential and higher Blend Difficulty.}
    \vspace{-10pt}
    \label{fig:user_study_form}
\end{figure*}

%% file: figures/user_study_ttl.tex
\begin{figure*}[!th]
    \vspace{-10pt}
    \centering
    \includegraphics[trim=0in 0in 0in 0in, clip,width=\textwidth]{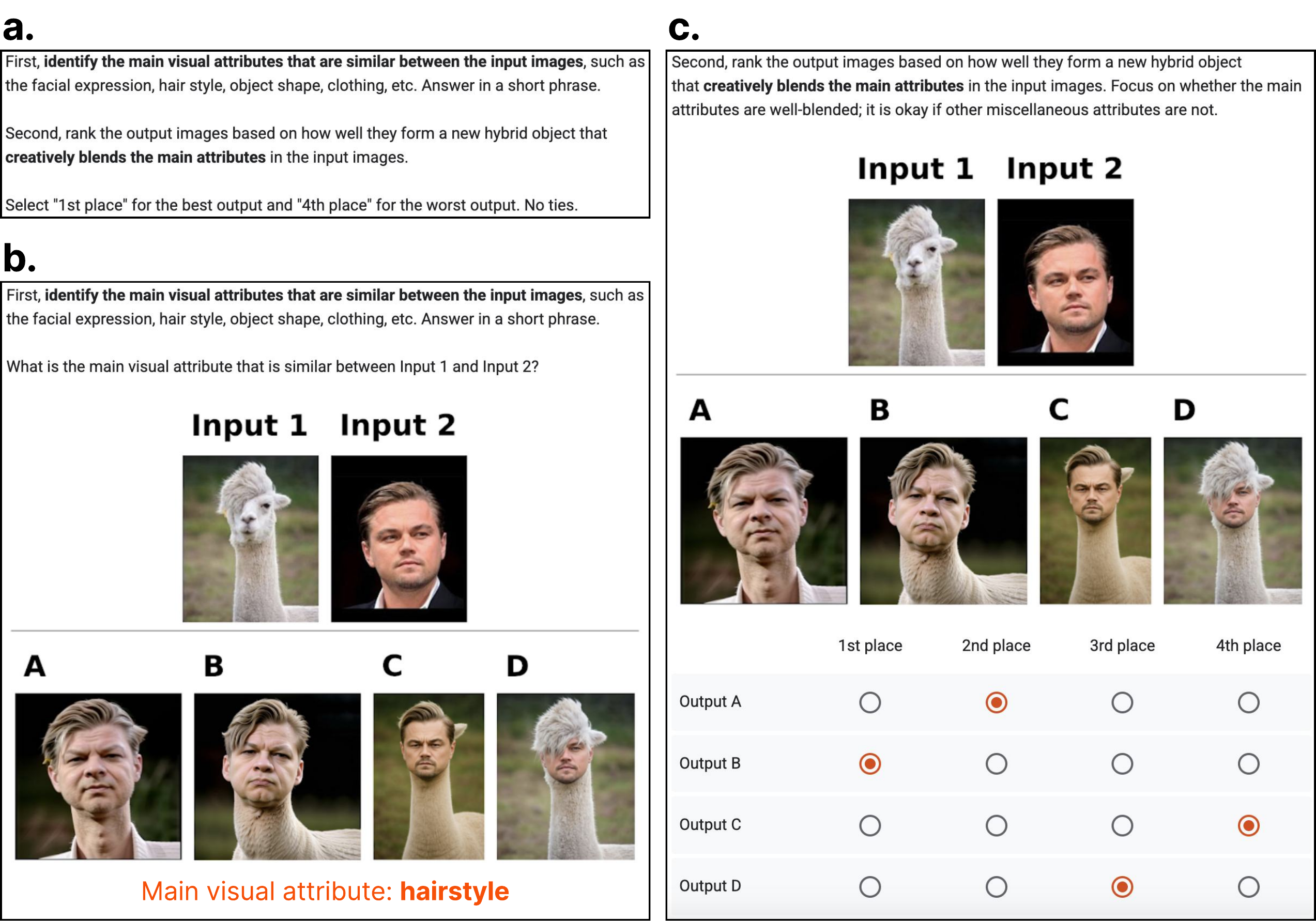}
    \vspace{-18pt}
    \caption{User study for assessing human preference of creative blends. \textbf{(a.)} The prompt shown to users. \textbf{(b.)} First, we ask participants to describe the main visual attributes shared by the two input images. \textbf{(c.)} Second, raters rank the blended output images from different methods based on how well they blend the main attributes.}
    \vspace{-10pt}
    \label{fig:user_study_ttl}
\end{figure*}

%% file: prompts/prompt_enhance.tex
\begin{tcolorbox}[mypromptbox, title=Enhancement Prompt]
Enhance this image to a high-resolution, professional quality while preserving all details and textures. Improve sharpness, smooth noise, and remove text and image artifacts for a clean, realistic, and well-balanced look.
\end{tcolorbox}

%% file: tables/supp_blending_quantitative.tex
\begin{table}[!th]
\centering
\resizebox{0.9\columnwidth}{!}{%
\begin{tabular}{lcccc}
\toprule

& \multicolumn{2}{c}{Totally Looks Like} & \multicolumn{2}{c}{Architecture} \\
\cmidrule(lr){2-3} \cmidrule(lr){4-5} \addlinespace[2pt]

Method & CLIP & DreamSim & CLIP & DreamSim \\
\midrule

CLIP Avg & 0.079 & 0.096 & 0.067 & 0.112 \\ 
Gemini \cite{comanici2025gemini} & 0.189 & 0.305 & 0.129 & 0.257 \\ 
GPT \cite{gpt-image-1} & 0.121 & 0.193 & 0.088 & 0.177 \\ 
Ours & \textbf{0.223} & \textbf{0.339} &\textbf{0.150} & \textbf{0.291} \\ 

\bottomrule
\end{tabular}
}%
\vspace{-5pt}
\caption{We generate multiple blends for each input image pair and measure the mean pairwise diversity of the output images using CLIP \cite{radford2021learning} and DreamSim \cite{fu2023dreamsim}. Higher diversity is better. Our method produces more diverse blends across multiple trials, in addition to more creative blends as shown in \Cref{tab:blending_quantitative}.}
\label{tab:supp_blending_quantitative}
\end{table}

%% file: sec_appendix/failure_cases.tex
\section{Failure Cases}
\label{sec:failure_case}

\input{figures/neg_vibe_fail}

\paragraph{Entangled Vibes} 
A limitation of negative vibe blending arises when desired and undesired attributes are entangled within the feature space. The orthogonalization process assumes that the positive and negative vibes can be separated into distinct, albeit non-orthogonal, subspaces. \Cref{fig:neg_vibe_fail} illustrates a failure example where the vibe of ``style change" is entangled with ``color change", which prevents the entangled vibe from being filtered out.

\input{figures/extrapolate_fail}
\paragraph{Extrapolation Uncertainty}
While effective in certain cases, extrapolating Vibe Blending does not always produce images with a meaningful exaggeration of relevant attributes (\Cref{fig:extrapolate_fail}).

\begin{figure*}[ht]
    \centering
    \includegraphics[width=\linewidth]{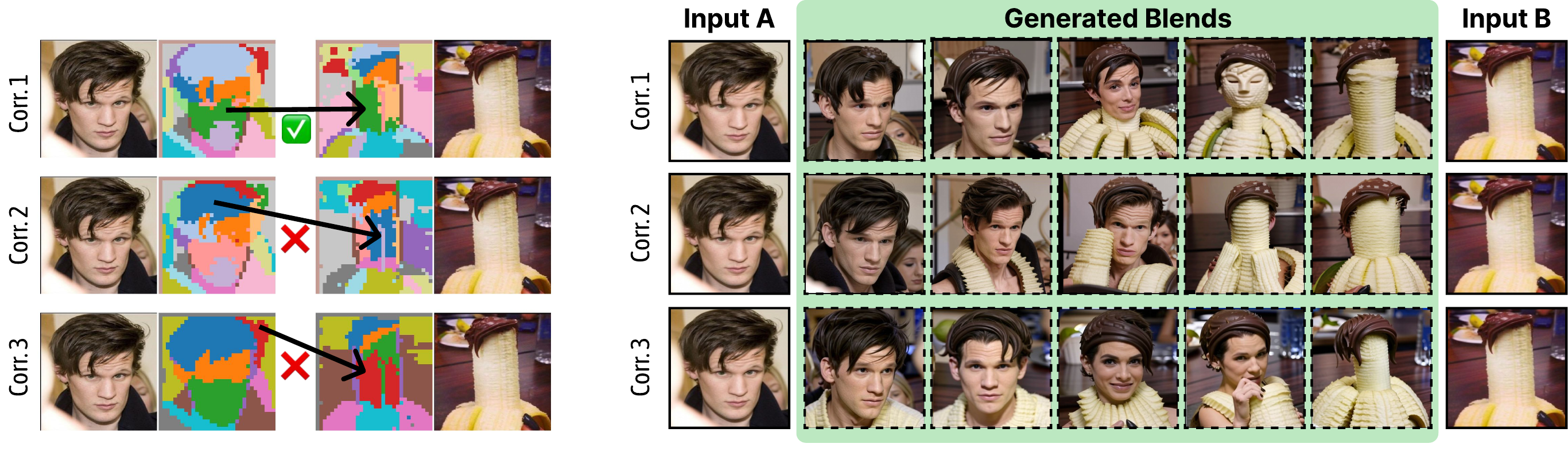}
    \caption{
    Failure cases of unsupervised correspondence. Our method depends on correspondence matching to identify which semantic regions should be blended. Different random seeds lead to different clusterings and therefore different correspondence maps. Good correspondence (top row) yields clean and semantically meaningful blends, whereas poor correspondence (middle and bottom rows) leads to low-quality blends with incorrect part-level mixing.
    }
    \label{fig:fail-correspondence}
\end{figure*}

\paragraph{Correspondence Failure.}
Vibe Space relies on unsupervised region-level correspondence matching between DINO token clusters (see \Cref{sec:appendix_correspondence}). This correspondence is essential for concept-level blending, as it determines which semantic regions in $I_A$ and $I_B$ should be merged. However, because the matching is unsupervised, it is not always reliable. When correspondence is incorrect (vary by random seeds), the resulting blends can degrade significantly, often producing incoherent or mismatched object-part combinations (\Cref{fig:fail-correspondence}).

\begin{figure*}[th]
    \centering
    \includegraphics[width=0.7\linewidth]{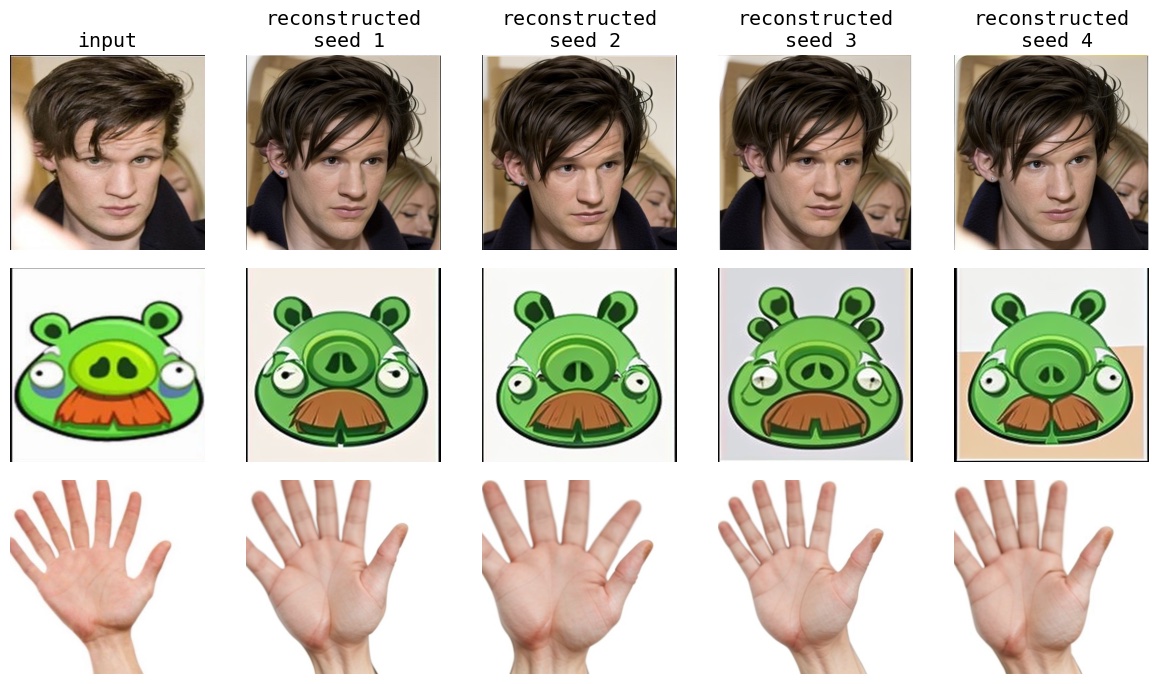}
    \caption{
    Failure cases of IP-Adapter reconstruction. Left column is input images; right four columns are reconstructed images from 4 random seeds. IP-Adapter reliably reconstructs images that lie within the Stable Diffusion training distribution (top row), producing consistent outputs across random seeds. In contrast, for out-of-distribution inputs (middle and bottom rows), IP-Adapter consistently fails to reproduce the original image, illustrating a fundamental limitation of SD-based decoders when applied to concept-level blending tasks.
    }
    \label{fig:fail-ipa}
\end{figure*}

\paragraph{Reconstruction Failure.}
Our method depends on IP-Adapter \cite{ye2023ip} to generate images conditioned on dense CLIP features. However, IP-Adapter is not always reliable outside the training distribution of Stable Diffusion. To isolate this limitation, we perform the following diagnostic:  
(1) extract dense CLIP features from an input image;  
(2) perform no blending and do not train Vibe Space;  
(3) feed the extracted features directly into IP-Adapter to reconstruct the input image.  
As shown in \Cref{fig:fail-ipa}, IP-Adapter can consistently reconstruct in-distribution images (e.g., human faces) across random seeds, but fails on out-of-distribution (OOD) inputs, producing unstable and inconsistent reconstructions.

%% file: figures/neg_vibe_fail.tex
\begin{figure}[!th]
    \vspace{-2pt}
    \centering
    \includegraphics[width=1.0\columnwidth]{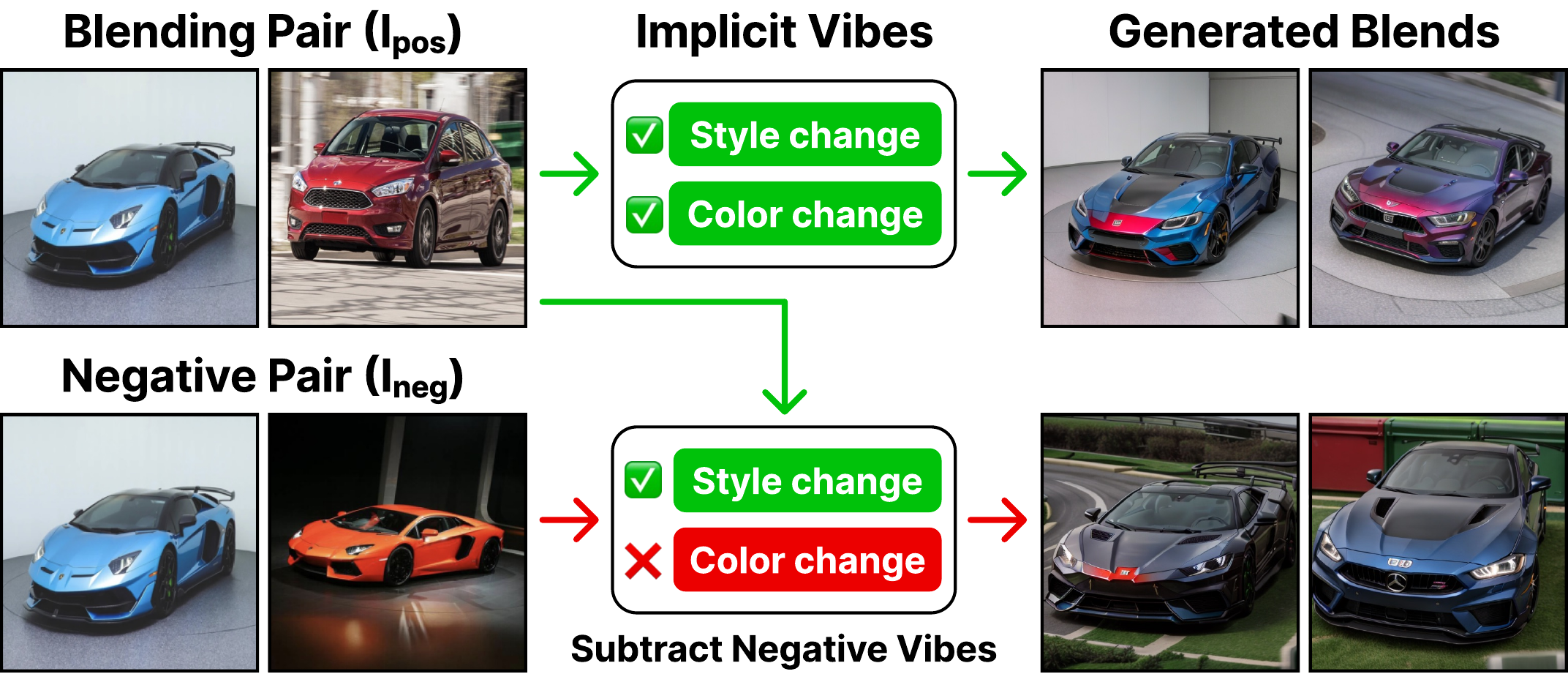}
    \vspace{-18pt}
    \caption{Negative vibe failure case. The positive inputs capture both a style change based on the types of car and a color change. The negative inputs intend to capture only color change. However, the attributes of style and color are entangled, making them difficult to separate with negative examples.}
    \vspace{-10pt}
    \label{fig:neg_vibe_fail}
\end{figure}

%% file: figures/extrapolate_fail.tex
\begin{figure}[!ht]
    \centering
    \includegraphics[width=\linewidth]{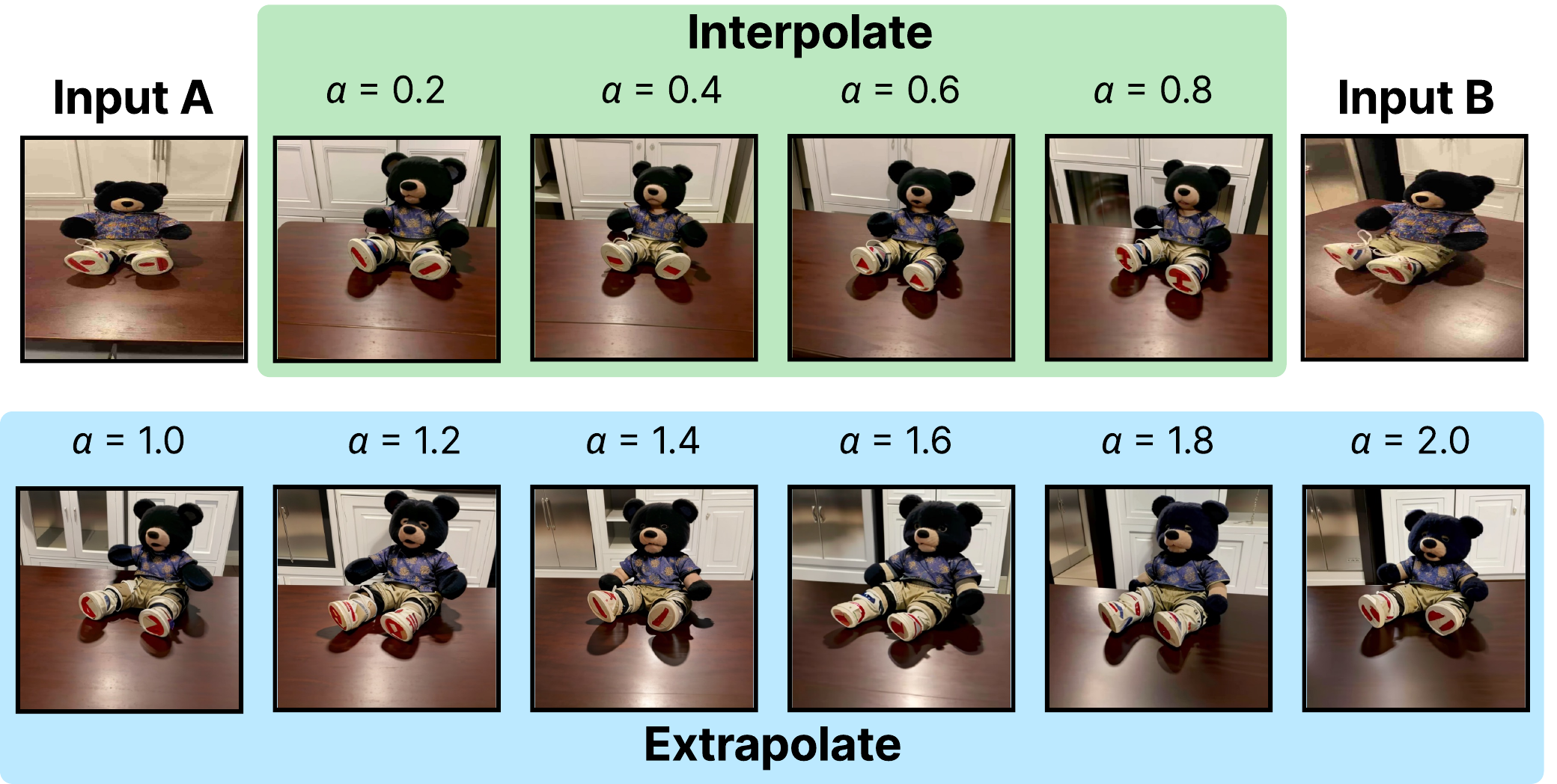}
    \vspace{-20pt}
    \caption{
    Failure case of Vibe Blending extrapolation. Extrapolating beyond $\alpha > 1$ does not produce further rotation of the object in the input images.
    }
    \label{fig:extrapolate_fail}
\end{figure}